\documentclass[]{interact}

\usepackage{epstopdf}
\usepackage{subfigure}

\usepackage{natbib}
\bibpunct[, ]{(}{)}{,}{a}{}{,}
\renewcommand\bibfont{\fontsize{10}{12}\selectfont}

\theoremstyle{plain}

\theoremstyle{definition}

\theoremstyle{remark}

\usepackage{lscape}
\usepackage{soul}
\usepackage{multirow}
\usepackage[table]{xcolor}
\PassOptionsToPackage{hyphens}{url}\usepackage[backref = page]{hyperref}
\hypersetup{
  colorlinks   = true, 
  urlcolor     = blue,
  linkcolor    = blue, 
  citecolor   = blue,
}

\usepackage{makecell}
\usepackage{comment}

\usepackage{afterpage}
\usepackage{capt-of}
\usepackage{amsmath}
\usepackage{siunitx}
\usepackage{booktabs, threeparttable} 

\usepackage{array}
\usepackage[toc,page]{appendix}
\usepackage{mfirstuc}
\usepackage{multicol}

\usepackage{graphicx}
\usepackage{algorithm} 
\usepackage{algpseudocode}
\usepackage{color}
\usepackage{slashbox}
\usepackage{makecell}
\usepackage[normalem]{ulem}
\usepackage{amsmath}

\usepackage{tabu}

\newcommand\setrow[1]{\gdef\rowmac{#1}#1\ignorespaces}
\newcommand\clearrow{\global\let\rowmac\relax}

\begin{document}


\title{Location retrieval using qualitative place signatures of visible landmarks}

\author{
\name{Lijun Wei\textsuperscript{a}\thanks{CONTACT Lijun Wei. Author. Email: l.j.wei@leeds.ac.uk},
Val\'erie Gouet-Brunet\textsuperscript{b}, 
and Anthony Cohn\textsuperscript{c, a, d, e
}
}
\affil{\textsuperscript{a}School of Computing, University of Leeds, UK
	\\\textsuperscript{b}LaSTIG, IGN-ENSG, University Gustave Eiffel, France
 	\\\textsuperscript{c}Department of Computer Science and Technology, Tongji University, Shanghai, China\\
 	\textsuperscript{d}School of Civil Engineering, Shandong University, Jinan, China\\
	\textsuperscript{e}School of Mechanical and Electrical Engineering, Qingdao University of Science \& Technology, Qingdao, China
	}
}

\maketitle
\begin{abstract} 
Location retrieval based on visual information is to retrieve the location of an agent (e.g. human, robot) or the area they see by comparing the observations with a certain form of representation of the environment. Existing methods generally require precise measurement and storage of the observed environment features, which may not always be robust due to the change of season, viewpoint, occlusion, etc. They are also challenging to scale up and may not be applicable for humans due to the lack of measuring/imaging devices. Considering that humans often use less precise but easily produced qualitative spatial language and high-level semantic landmarks when describing an environment, a qualitative location retrieval method is proposed in this work by describing locations/places using qualitative place signatures (QPS), defined as the perceived spatial relations between ordered pairs of co-visible landmarks from viewers' perspective. After dividing the space into place cells each with individual signatures attached, a coarse-to-fine location retrieval method is proposed to efficiently identify the possible location(s) of viewers based on their qualitative observations. The usability and effectiveness of the proposed method were evaluated using openly available landmark datasets, together with simulated observations by considering the possible perception error.   
\end{abstract}



\section{Introduction}
\label{sec:intro}


\textcolor{black}{
Depending on the context and scale of applications, a \emph{location}\footnote{\emph{Location} is used interchangeably with \emph{Place} in the rest of this paper.} can either be defined as a geographical name (e.g. Place Dauphine, Paris) with a varied size, and a crisp or rough boundary~\citep{Bittner2000}, an area with homogeneous observations~\citep{Kuipers2000}, a 2D/3D linestring (e.g. a road section), a zero-dimensional 2D/3D point, 
or a point  with the agent’s pose (i.e. roll, pitch and yaw angles) attached~\citep{Kendall2015,Irschara2009,Sattler2012}.
Location retrieval is then to identify the corresponding location and (or) pose of an agent in an environment based on certain types of sensing modalities and any existing knowledge of the environment.}
\textcolor{black}{
When satellite systems (e.g. GPS) cannot be used to locate, such as indoors and urban canyons where there are not enough satellites in view, other sensing modalities such as visual information has been explored in visual localization~\citep{Korrapati2012,Zamir2016,Piasco2018,Pion2020} and place recognition~\citep{Lowry2016}.
} 
\textcolor{black}{
In these methods, agents are located by comparing their observations with a pre-generated representation of the environment, such as a collection of geo-tagged images~\citep{Pion2020, piascoIJCV21}, a virtual city model~\citep{Cappelle2012}, a topological map of 2D/3D image features~\citep{Durrant-Whyte2016}, or high-definition maps with semantic information of objects~\citep{zang2017, Hery2021}.}
\textcolor{black}{
However, existing methods generally use low-level visual features~\citep{Jegou2010, Korrapati2012, Chen2017, Zhang2021} or high-level features with their explicit configurations to describe the environment~\citep{Zamir2014,li14geolocation, Lamon2001,Lamon2003,Schlichting2014,panphattarasap2016visual}. 
These descriptions may not be robust to change of season, viewpoint, occlusion, even there are recent contributions focusing on addressing these challenging conditions~\citep{piascoIJCV21}.
Moreover, these methods may not be applicable to agents with human-like visual sensing capabilities when accurate measuring and estimation of distance/angles is not available.}
\textcolor{black}{
In fact, humans often rely on high-level semantic features and less precise but easily produced and understood spatial language in description of environment~\citep{Tversky1993,chen2013}, such as \emph{`(next to) a tree'}, \emph{`(between) a tree and a road sign'}, \emph{`(under) the Eiffel Tower'}. 
In these descriptions, those semantic features (or place names) with known or relatively better known locations are selected as the \emph{landmarks}~\citep{Sadalla1980} or \emph{anchor points}~\citep{Couclelis1987} to define the locations of adjacent points. Compared to low-level features, semantic features are easier to capture, communicate, and more robust to changes of season, viewpoint, occlusion, etc.
}

\textcolor{black}{
In this work, locations are described using \emph{place signatures}, defined as the perceived qualitative spatial relations between ordered pairs of co-visible semantic landmarks from that location, including the ordered (e.g. clock-wise) sequences of landmark \emph{types}, the \emph{relative orientations} and the \emph{qualitative angles} between each ordered pair of landmarks. Based on this definition, a location retrieval method is proposed by first dividing the navigable space into distinct locations (i.e. place cells), pre-calculating their signatures, then applying a coarse-to-fine retrieval method to efficiently identify the viewers' possible location(s) based on their observations using approximate Hashing.}
\vspace{0.3cm}

\begin{figure}[t] 
	\centering
 	\includegraphics[width = 0.8\linewidth]{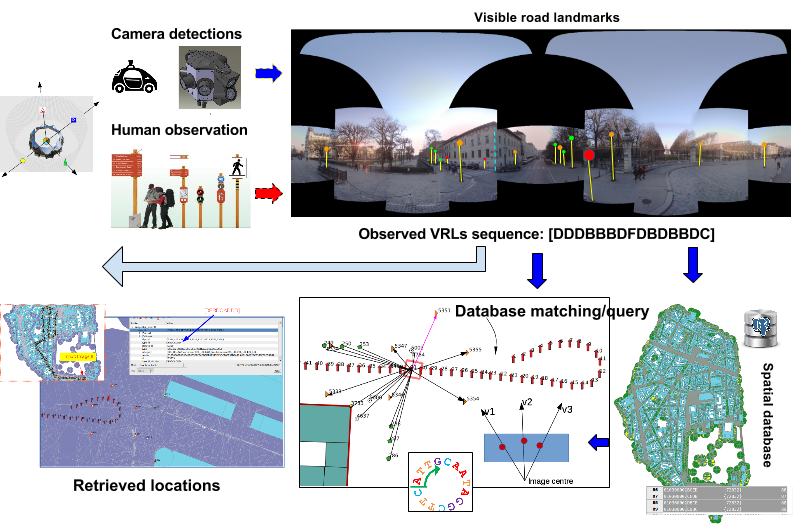}
 	\caption{\textcolor{black}{Demonstration of the proposed location retrieval method: given the perceived ordered sequences of visible road landmark (VRL) \emph{types} and other qualitative spatial relations by users (or images), the location retrieval problem is to find the reference place cells with the most similar place  signature to the observed one.}}
    \label{fig:procedure}
\end{figure}

\noindent\textbf{Related Work}.
\textcolor{black}{A location can either be described using the spatial relations between a semantic landmark and a reference frame from a fixed perspective, such as the `East/West-North/South' frame when reading a map}; or using the perceived spatial relations between landmarks from the point of view of a moving agent.
These two modes form the foundation of human spatial mental models~\citep{Tversky1993}, and have both been studied for representing locations qualitatively for applications such as navigation~\citep{Wang2005,Fogliaroni2009}, and spatial information queries~\citep{Yao2006}. More specifically,
\begin{itemize}
    \item For methods with a known reference frame, disjoint points~\citep{Clementini1997}, polygons, regions~\citep{Bittner2000} or the combination of points, lines and polygons~\citep{du2015} were used to divide space into non-overlapping areas, enabling spatial inference using relevant direction, distance or topological relations~\citep{cohn2014,Freksa2018}. For example, \citet{Wang2005} proposed to describe the qualitative position of target objects using their cardinal direction relations~\citep{Frank1991,Egenhofer1999} with respect to the corresponding landmarks determined by a Voronoi model. 
    \item For methods without a reference frame, \citet{LEVITT1990305} proposed to divide the environment into regions using the lines connecting pairs of point landmarks such that the same order of landmarks can be perceived by agents from any points within each region. This method was improved on by \citet{Schlieder1993} to differentiate between adjacent regions by augmenting the order of landmarks with their complementary directions. \citet{Fogliaroni2009} proposed a similar approach while the decomposition of space was based on the visibility and occlusion of extended convex landmarks instead of points. Places were also represented using qualitative distances to landmarks such as [\emph{very close, close, medium, far, very far}]~\citep{Panorama2004}, or nearness relations identified through data mining~\citep{Duckham2001}.
\end{itemize}
Both of these methods generally assume landmarks can be correctly, completely, and
uniquely identified by agents, the locations (or \emph{ID}) of the landmarks are known, and the initial positions of agents are given when used in navigation applications. However, visual perceptions are prone to errors due to environmental or internal \textcolor{black}{factors}, and the initial position of agents may be unknown.
Moreover, although various theoretical models were proposed to identify qualitative locations, the scales of existing experiments are generally small with limited number of landmarks. The scalability as well as the time complexity of \textcolor{black}{these frameworks} were rarely investigated. Little work has been done in this area from the perspective of information retrieval.
\textcolor{black}{In our approach (Figure~\ref{fig:procedure}), instead of using the explicit location/ID of landmarks~\citep{Wang2005}, only the semantic information of landmarks, e.g. \emph{`type'} (e.g. street light), is used as it is generally easy to capture by humans compared to other more accurate measurement.
}

\textcolor{black}{Among the large range of existing location retrieval techniques, the concept proposed by \cite{Weng_2020} is close to our approach. However, 1) instead of sampling locations using $10$x$10$m regular grids, in our approach the navigable space is divided into distinct locations (i.e. place cells) following the definition of individual qualitative spatial relations, ensuring that consistent spatial relations being observed by agents from anywhere inside each place cell; 2) instead of computing the direction of landmarks with respect to the \emph{True North} like in \citep{Weng_2020} which is not always feasible by agents like humans to judge, the relative angles between the lines of sight of order landmark pairs are considered; 3) the possible occlusion of landmarks by other objects is considered in our work when creating place signatures; and 4) instead of using distance measures under an exhaustive searching strategy, which is time-expensive and impractical\footnote{Detailed discussion on time complexity will be given in Sections~\ref{sec:edit_distance}, and \ref{sec:compare_distance}.} for real-time applications, a coarse-to-fine location retrieval method is proposed in this work with an approximate hashing technique to improve the retrieval efficiency.}

In the remainder of this paper, the proposed qualitative place signature is presented in Section~\ref{sec:signatureDefinition} and the location retrieval method in Section~\ref{sec:locationRetrieval}; experimental results are given in Section~\ref{sec:experiments}, followed by conclusions in Section~\ref{sec:conclusion}.

\begin{figure}[t] 
	\centering
    \subfigure[Different types of visible road landmarks (in Paris)]{\label{fig:background}
    \includegraphics[width = 0.48\linewidth]{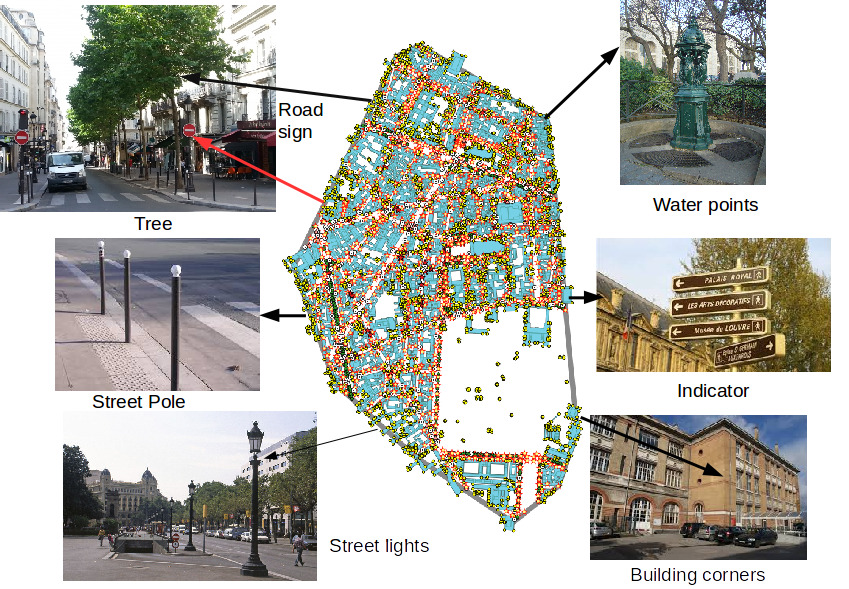}} 
    \subfigure[The visible zones of road landmarks.]{\label{fig:mapQGIS}
 	\includegraphics[width = 0.48\linewidth]{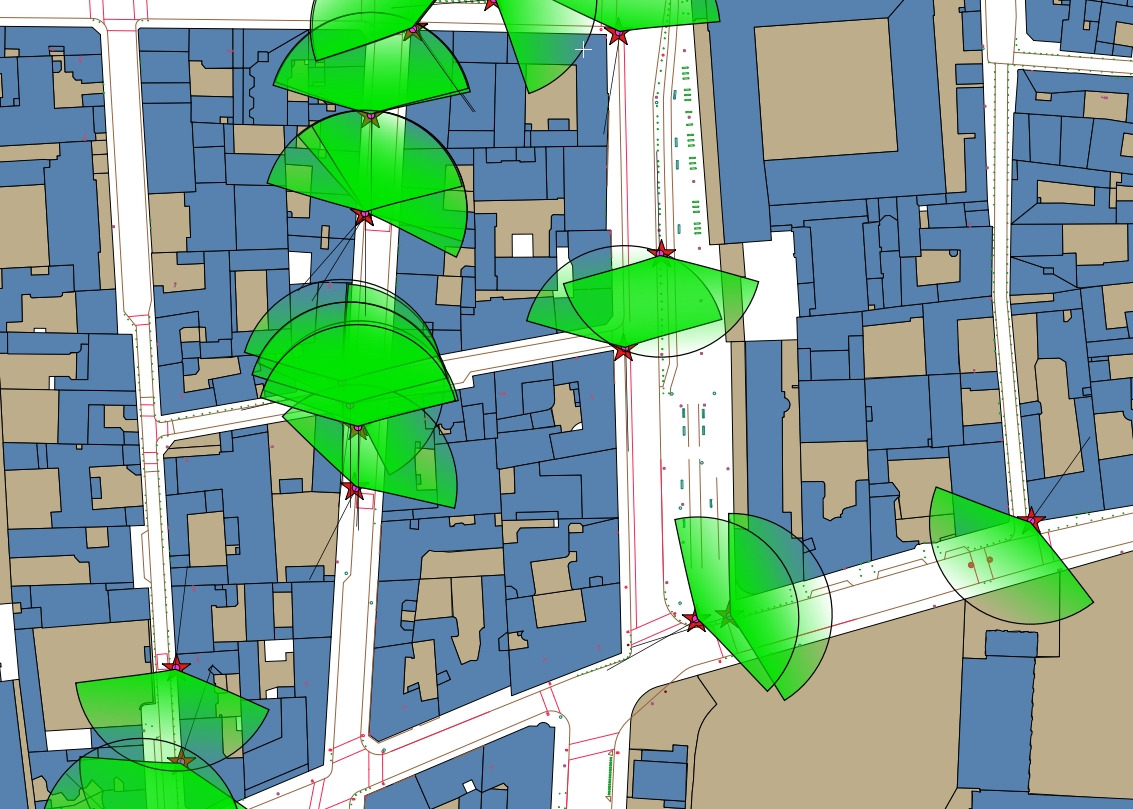} }
 	\caption{Examples of visible landmarks in a city and the visible zones of landmarks which are not always full circles since some of them may have an inner front. Pictures are from Google images by searching ``Paris''.} 
\end{figure}
 
\section{Landmarks}
\label{sec:landmarks}

In this work, landmarks are defined as distinctive, stable and recognisable objects in the environment, such as road signs~\citep{Soheilian2013}, trees, street lights. Some examples are shown in Figure~\ref{fig:background}.
Though the appearance of certain landmarks may change in a regular or an irregular way, for example, most trees change colour and forms from spring to winter, street lights are switched off and on during day and night, their locations are mostly static and they can be easily identified by humans\textcolor{black}{, whose descriptions of them often carry a strong semantic information, which is informative and robust to visual changes}. The \textcolor{black}{location information} of these landmarks \textcolor{black}{is often readily available}, either being automatically reconstructed from optic sensor data captured by mobile platforms, manually annotated by human surveyors,
or sourced from various open data services, such as \emph{Ordnance Survey Roadside Asset Data Services}\footnote{All website links and last accessed dates are listed in Appendix~\ref{sec:links}.}
and \emph{Find Open Data} (the open data portal of UK central government, local authorities and public bodies) 
in the UK, \emph{OpenDataParis} 
in France, and crowd-sourced maps like  \emph{OpenStreetMap}
~\citep{Rousell2015}.
\textcolor{black}{In terms of volume, semantic maps also represent the environment in a much more compact and refined way than other kinds of more commonly used  maps, such as images or LiDAR. Thanks to the fact that these initiatives are now widespread and describing spatial regions on a large scale, it would be interesting to exploit their usage in applications like large-scale location retrieval. } 

In this work, each landmark $S_i$ has an attached  set of attributes, \textcolor{black}{written} as:
\begin{equation}
S_i =(id, centre(x, y), contour_{2d},  visible\_zone, type, type\_id)
\end{equation}
where \emph{id} is the unique index of a landmark in a database but it is usually unknown to agents; \emph{type} is the general category of this landmark, such as road sign, tree; and \emph{type\_id} is a character encoding such category, such as \emph{`B'} for \emph{bin}. \textcolor{black}{More detailed visual, semantic or spatial landmark attributes can be considered to further improve the discriminating ability of landmarks 
if these information cannot be uniquely inferred from the general `type'. For example,
\begin{itemize}
    \item for semantic information, the granularity of the category of landmarks can be gradually increased depending on the application and context, such as including the sub-category of a tree (e.g. oak, horse chestnut), or representing a road sign as `Road sign'$\rightarrow$`Speed limit sign'$\rightarrow$`Max speed sign'$\rightarrow$`20/30/40mph' sign, etc.
    \item for visual information, the shape (e.g. round, triangle, square), colour (e.g. red, blue, brown), text content, material (e.g. metal, concrete), direction (e.g. front, back) and other visual attributes can be added;
    \item for spatial extent, the granularity of the spatial extent of landmarks (e.g. point, 2D outline, 2D contour, 3D bounding box, explicit 3D model) and their projected extent on observers' view (e.g. point, horizontal line segment, horizontal \& vertical line segments, projected bounding box/contour) can also vary.
\end{itemize}}
\noindent \textcolor{black}{For other attributes, \emph{centre(x,y)} is the location of the landmark centre, $contour_{2d}$ is the polygon marking the 2D extent of the landmark, and
\emph{visible\_zone} is a (multi)-polygon marking the area from where this landmark can be perceived in the same projected coordinate system, such as \emph{EPSG:27700} in the UK}. \textcolor{black}{The \emph{visible\_zone} of each landmark can be affected by multiple factors, such as }the location and intrinsic direction of the landmark (if there is any), the maximum range from where this landmark can be observed based on its size, height, and visual salience\footnote{Visual salience is the perceptual quality which makes some items stand out from their neighbours.}, its occlusion caused by other objects, as well as observers' eyesight, height, and the weather and lighting conditions (e.g. day/night).

In this work, the \emph{default} visibility area of a landmark is defined as a circle with a fixed radius for different types of landmarks, such as a $30m$ buffer as illustrated in Figure~\ref{fig:mapQGIS}. Then, this default area is reshaped by considering the intrinsic direction and occlusion of landmarks by other objects (Section~\ref{sec:database}).
For a landmark $T$ with \textcolor{black}{an} intrinsic front/back, such as some traffic signs, the circular visibility area can be separated into front and back half-circles and treated as \textcolor{black}{belonging to} two different landmarks noted as $T_F$ and $T_B$. The orientation of such landmarks can either be collected by on-site survey, reconstructed from sensing data, or inferred from the landmark type, direction of nearby road networks, and existing standards on infrastructure installation. 
For example, traffic signs normally face oncoming traffic except that those indicating on-street parking controls are parallel to the edge of the carriageway, and some flag-type direction signs are pointing approximately in the direction to be taken.



\section{Place Signatures based on the Qualitative Spatial Relations between Visible Landmarks}  
\label{sec:signatureDefinition}

\textcolor{black}{
In this section, qualitative place signatures (\emph{QPS}) are introduced to describe the configuration of visible road landmarks from viewers' point of view, including the order of appearance, relative orientations, and qualitative angles between the lines of sight of ordered pairs of landmarks. 
Based on this definition, a study space can be divided into distinct reference place cells such that the same \emph{QPS} being observed by agents from anywhere inside each cell.
Given a viewer’s new observation, their location can be retrieved by finding the place cell(s) with the best matched reference signatures}.

\textcolor{black}{In this section, the three types of qualitative spatial relations are introduced in Section~\ref{sec:ioc_order} to~\ref{sec:augment}, followed by practical steps for creating and maintaining a reference database and discussions on the impact of landmarks uncertainty in Section~\ref{sec:database}.
}

\subsection{The viewing order of visible landmarks on a panorama}
\label{sec:ioc_order}

\begin{figure*}[htp!] 
	\centering 
    \includegraphics[width = 0.45\linewidth]{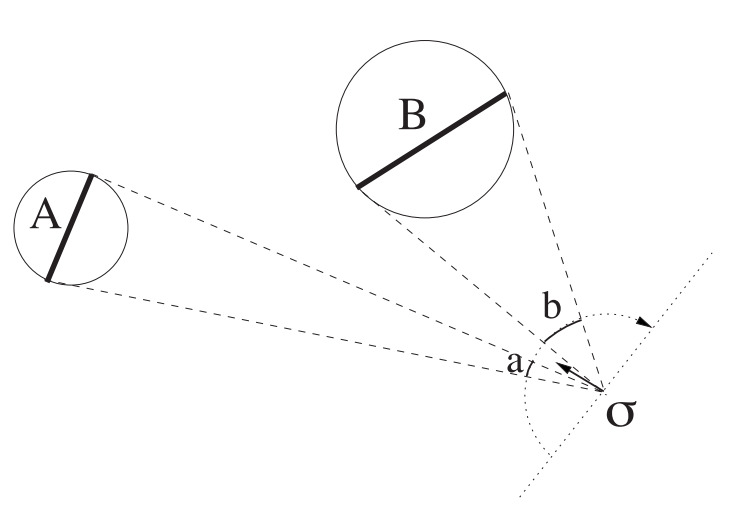} 
    \caption{The projections of two landmarks A and B \textcolor{black}{on the image plane of viewers from} a viewpoint $\sigma$. The dashed lines represent the lines of sight (\citet{Ligozat2015}).}     \label{fig:landmarkOrder1}
\end{figure*}

As landmarks seen from a particular viewpoint appear as if they are overlaying on the surface of a sphere centred on the viewer's eyes (Figure~\ref{fig:landmarkOrder1}), or the image plane of a panoramic camera \citep{Galton1994}, the first component of our proposed place signature is the ordered sequence of the types of visible landmarks \textcolor{black}{seen} from a location. 

If we represent a viewer as $\sigma_i =(P_i, \nu_i)$, where $P_i$ is the 2D position of the viewer's centre (of eyes) \textcolor{black}{in a coordinate system}, and $\nu_i$ is an unit vector representing the viewer's viewing direction. The viewer's field of view (FOV) can be defined as the fan-shaped area centred at their position and their direction $\nu_i$. 
Then, the \textcolor{black}{projected} interval of a visible landmark is defined by the extreme points of its projections on a selected horizontal line of the viewer's image plane, written as $I=(x_1, x_2)$ where $x_1$ and $x_2$ are real numbers and $x_1 \leq x_2$. \textcolor{black}{When multiple landmarks are present,} a viewer will be able to describe their environment using the \textcolor{black}{ordering} relations of the intervals \textcolor{black}{of these landmarks}. Additional information like the proximity/distance $d(I)$ of a landmark to the viewer and the height $h(I)$ of the landmark on the viewer's image plane can also be captured. 

\noindent \textbf{The viewing order of two landmarks \textcolor{black}{A, B}} from a viewpoint can be captured by assuming the viewer is facing towards the landmarks and looking from left-to-right by turning clockwise (or alternatively right-to-left in the anti-clockwise direction). Then, the viewing order of these two landmarks are decided using the starting  \textcolor{black}{and ending} points of the two projected intervals $I_A(x_1^A,x_2^A)$ and $I_B(x_1^B,x_2^B)$, as:
\begin{enumerate}
    \item If $x_1^A<x_1^B$, then $A\rightarrow B$; otherwise, $B\rightarrow A$: if the leftmost extremes \textcolor{black}{of the two intervals} are different, the one with a smaller \textcolor{black}{starting point} is considered appearing first.
    \item If $((x_1^A=x_1^B) \land (|x_2^A-x_1^A|>|x_2^B-x_1^B|))$, then $A\rightarrow B$; otherwise, $B\rightarrow A$: if the two leftmost extremes are the same and one interval is longer than the other, the longer/wider landmark is considered appearing first.
    \item If $(x_1^A=x_1^B)\land(|x_2^A-x_1^A|=|x_2^B-x_1^B|) \land (d(A)<d(B))$, then $A\rightarrow B$; otherwise, $B\rightarrow A$: : if the two leftmost extremes and the lengths \textcolor{black}{of two intervals} are the same, the front landmark is considered as appearing first (followed by the landmark behind if it is visible).
\end{enumerate}
For point-like landmarks such as those infrastructure assets attached to the ground with a single pole, e.g. traffic lights, street lamps, trees, the x-location of the pole on the horizontal line can be used to decide the order of landmarks.

\noindent \textbf{The relative positions of two landmarks on the panorama}. \textcolor{black}{In addition to the ordering}, as spatial occlusion occurs when a landmark appears before another with respect to a viewpoint, the \emph{Interval Occlusion Calculus} (IOC) \citep{Ligozat2015} \textcolor{black}{is adapted in this work to further} describe the relative positions of the 
\textcolor{black}{projected intervals of ordered} 
pairs of landmarks. 

\begin{table}[ht!]
\small
\begin{center} 
\caption{The 13 Allen's Interval Calculus relations between an interval A and B~\citep{allen1983} are shown in the top row of each cell, and the corresponding adapted Interval Occlusion Calculus relations~\citep{Ligozat2015} are shown in the second row. The symbol $+$, $-$ encode the relative closeness of $A$ and $B$ to the viewer, i.e. $+$ for in front of and $-$ for behind. } 
\label{tab:ioc}
  \begin{tabular}{ | c  | c|c|c||  } \hline
  \textbf{Relation} & \textbf{Relation}&\textbf{Relation} & \textbf{Relation}\\  \hline\hline
  \includegraphics[scale=0.8]{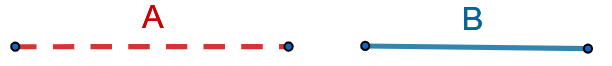} &
  \includegraphics[scale=0.8]{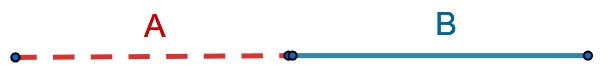}&\multicolumn{2}{|c|}{\includegraphics[scale=0.8]{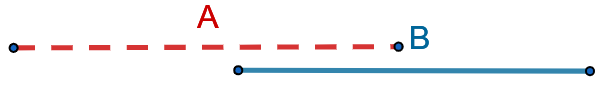}} \\ \hline
  {\includegraphics[width=3cm]{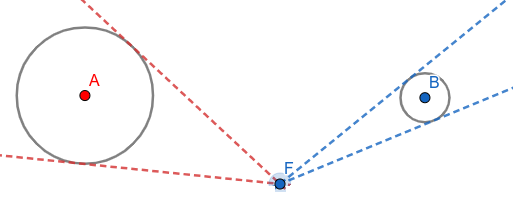}} &
  { \includegraphics[width=3cm]{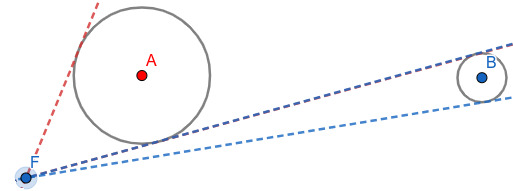}}&
  { \includegraphics[width=3cm]{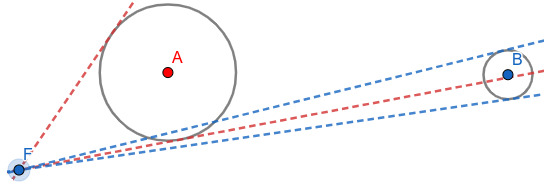}} & {
  \includegraphics[width=3cm]{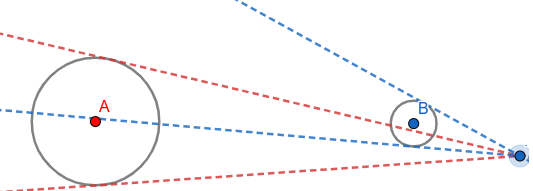}}   \\  
  precedes : $A\;p\;B$  &meets : $A\;m\;B$&\makecell[l]{overlaps\\ \& in front}: $A\;o^+\;B$&\makecell[l]{overlaps\\ \& behind}: $A\;o^-\;B$\\ \hline\hline
   
 \multicolumn{2}{|c|}{\includegraphics[scale=0.8]{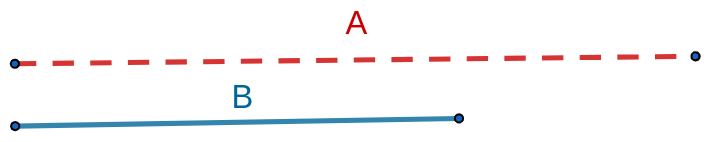}}&\multicolumn{2}{|c|}{\includegraphics[scale=0.8]{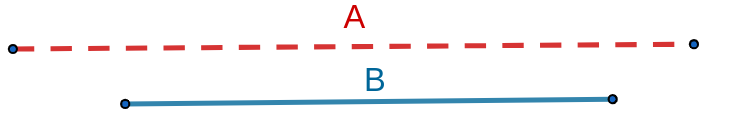}} \\ \hline
 {\includegraphics[width=3cm]{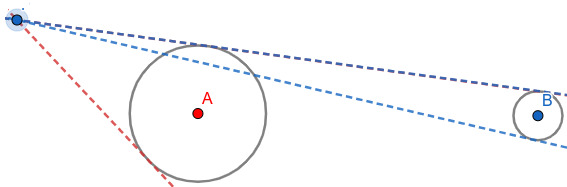}}&
 \includegraphics[width=3cm]{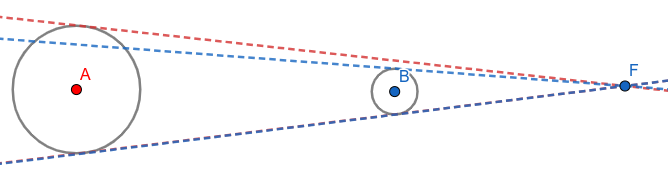} &{\includegraphics[width=3cm]{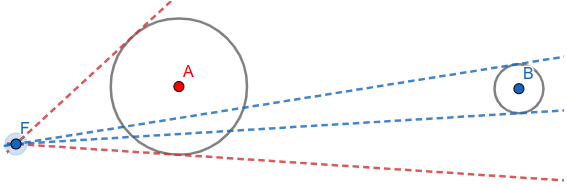}} &
 {\includegraphics[width=3cm]{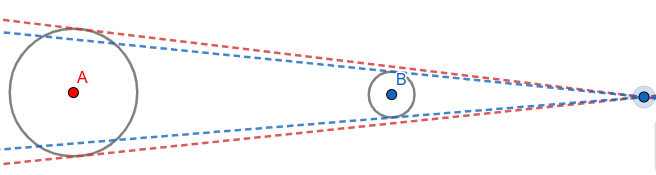}}\\  
 \makecell[l]{started by \\\& in front}: $A\;si^+_*\;B$ &\makecell[l]{started by \\\& behind}: ${A\;si^-\;B}$ &
 \makecell[l]{\textcolor{black}{contains}\\\& in front}: $A\;di^+_*\;B$ &
 \makecell[l]{\textcolor{black}{contains} \\\& behind}: $A\;di^-\;B$\\ \hline\hline
  
 \multicolumn{2}{|c|}{\includegraphics[scale=0.8]{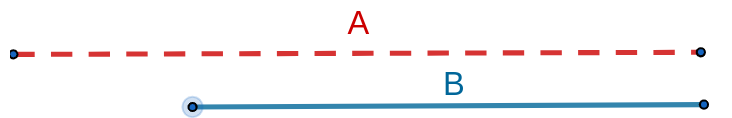}}&\multicolumn{2}{|c|}{\includegraphics[scale=0.8]{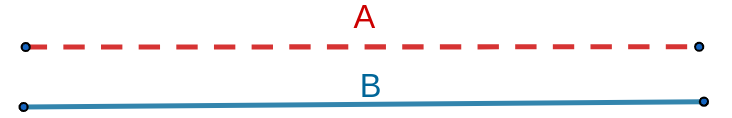}} \\ \hline 
 {\includegraphics[width=3cm]{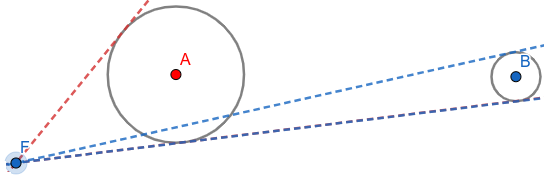}}&
  {\includegraphics[width=3cm]{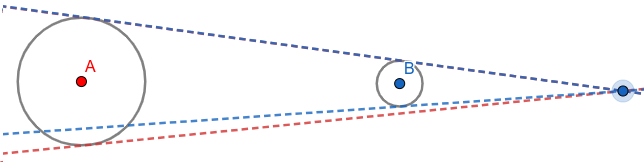}}&\includegraphics[width=3cm]{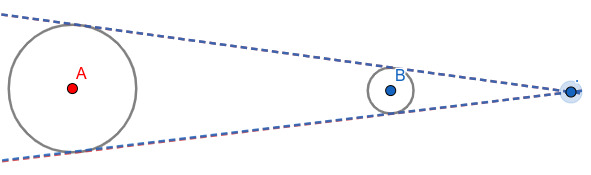}& \includegraphics[width=3cm]{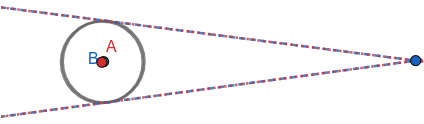}\\ 
 \makecell[l]{finished by \\\& in front}: $A\;fi^+_*\;B$&  \makecell[l]{finished by \\\& behind}: $A\;fi^-\;B$&\makecell[l]{coincides with\\ \& in front}: $B\;c^+_*\;A$&
 $A\;eq\;B$ (not used)
  \\  \hline \hline
\end{tabular}  
\end{center} 
\end{table}

\begin{figure*}[ht!] 
	\centering 
    \includegraphics[width = \linewidth]{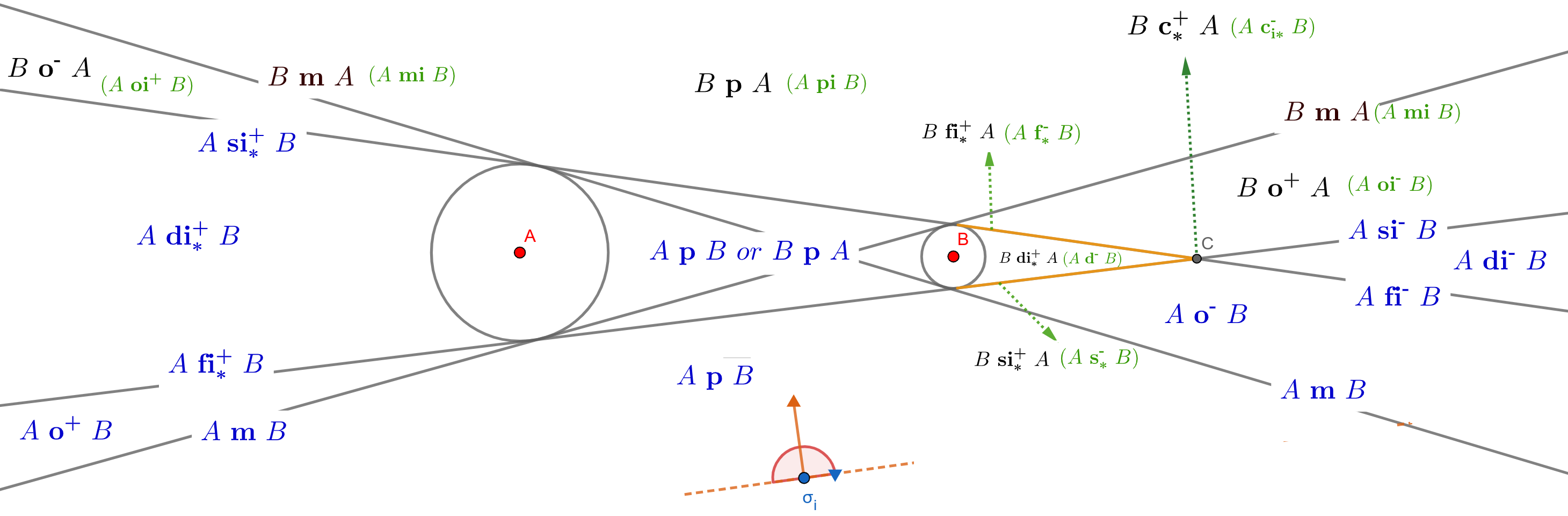}
    \caption{The lines of sight between two landmarks A and B and the corresponding \emph{IOC} (modified from~\citep{Ligozat2015}) relations. The two elements of a relation are given based on the left-to-right order from a viewer's perspective. \textcolor{black}{Note: the relations shown in green in the brackets are only for illustration purpose. They are not directly used in this work as only relations retaining a left-to-right order are considered. }}        \label{fig:landmarkOrder2}
\end{figure*}
\textcolor{black}{In terms of their ordering on a $0-360^o$ coordinate system\footnote{When a best viewing direction(s) can be found such that all landmarks are located in a viewer's field of view (FOV), this coordinate system can start from the leftmost landmark; otherwise, the zero-direction of this coordinate system will depend on the starting direction of the viewer.},}
landmarks with smaller left extremes are always considered first in this work. 
Then, their relative positions on the panorama are represented using a reduced set of the 11 \emph{Interval Occlusion Calculus (IOC)} relations as shown in Table~\ref{tab:ioc}, \textcolor{black}{resembling}\emph{Allen's Interval Algebra}~\citep{allen1983, du2016, QSRlib2016} except that the intervals are the projections on a panorama and the occlusion information 
is
considered to encode the relative closeness of landmarks to viewers. \textcolor{black}{In more detail:}
\begin{itemize}
    \item when there is no occlusion between two landmarks \textcolor{black}{from a viewpoint}, the first seen landmark could precede ($\boldsymbol{A\;p\;B}$) or meet the \textcolor{black}{following} one ($\boldsymbol{A\;m\;B}$). 
    \item when there is partial occlusion between them, the first seen landmark could overlap and be in front of ($\boldsymbol{A\;o^+\;B}$) or behind the other one ($\boldsymbol{A\;o^-\;B}$).
    
    \item when the starting points of the two intervals are the same and the interval of one landmark, e.g. $A$, is longer than the other, e.g. $B$, $A$ could be started by and in front of ({$\boldsymbol{A\;si^+_*\;B}$} or behind $B$ ($\boldsymbol{A\;si^-\;B}$).
    \item when the starting points are different, the landmark with a longer interval, e.g. $A$, could \textcolor{black}{contain},
    and be in front of ($\boldsymbol{A\;di^+_*\;B}$) or behind  ($\boldsymbol{A\;di^-\;B}$) the landmark $B$; or be finished by, and in front of ($\boldsymbol{A\;fi^+_*\;B}$) or behind $B$ ($\boldsymbol{A\;fi^-\;B}$).
    
    \item when the two intervals coincide, one landmark could coincide with and be in front of another, e.g. $\boldsymbol{A\;c^+_*\;B}$\footnote{Note since the front landmark is always considered first, the relation `A coincides with B and B is in front of A' is not used in this work.}.
    
    \item When two landmarks are equal 
    to each other \textcolor{black}{or the front one completely occludes the one behind}, only one of them will be observed/considered in this work. Therefore, the equivalent relation \emph{eq} is not used. 
\end{itemize}
{Note that the $*$ in $A\{si^+_*$, $di^+_*$, $fi^+_*$, $c^+_*\}B$ means that the landmark behind, i.e. B, can only be seen by the viewer when it is tall enough although it’s bottom half is occluded by a shorter landmark in front of it; otherwise, it will be completely occluded and only the front landmark A will be observed. This constraint can be expressed with the projection $h_A, h_B$ of the two landmarks on the vertical axis using Allen's interval calculus (\textcolor{black}{by adding a subscript $_a$ for each relation}) as ($\neg[(h_A\;si_a\;h_B) \lor (h_A\;di_a\;h_B) \lor (h_A\;fi_a\;h_B) \lor (h_A\;eq_a\;h_B)]$.})

In summary, five of the original \emph{IOC} relations~\citep{Ligozat2015} and six of the inverse \emph{IOC} relations are used in this work. \textcolor{black}{Moreover,} constraints are added to the projected intervals on the y-axis for four relations ($si^+_*,~di^+_*,~fi^+_*,~c^+_*$) using Allen's interval relations, similar to the Rectangle Algebra~\citep{rectangleAlgebra1999}. 


Using the above definition of viewing order and the 11 modified \emph{IOC} relations, there could be 18 types of relations between two co-visible landmarks \textcolor{black}{depending on the viewpoint and location of viewers}. As shown in Figure~\ref{fig:landmarkOrder2}, \textcolor{black}{the ten relations starting from \emph{A} are marked in blue ($A\{si^+_*,di^+_*,fi^+_*,o^+,m,p,o^-,fi^-,di^-,si^-\}B$), and the eight relations starting from \emph{B} are marked in black ($B\{o^-,m,p,fi^+_*,di^+_*,si^+_*,c^+_*,o^+\}A$). $ApB$, $AmB$, $BpA$, and $BmA$ are each appearing twice in different place cells.}
\textcolor{black}{
An exemplar location $o_1$ (with viewing direction marked in pink) is given to illustrate how the relation $\langle A\;p\;B\rangle$ can be observed in the bottom place cell.} 
If a viewer can provide \textcolor{black}{their} observed relation between the two landmarks, we can then roughly identify their located areas. Note that when the viewer is between the two landmarks, the landmarks can only be seen when the viewer turns around. Therefore, the viewed relation can either be $\langle A\;p\;B\rangle$ or $\langle B\;p\;A\rangle$ depending on the viewer's initial direction.

\begin{figure*}[ht!] 
	\centering 
    \includegraphics[width = 0.7\linewidth]{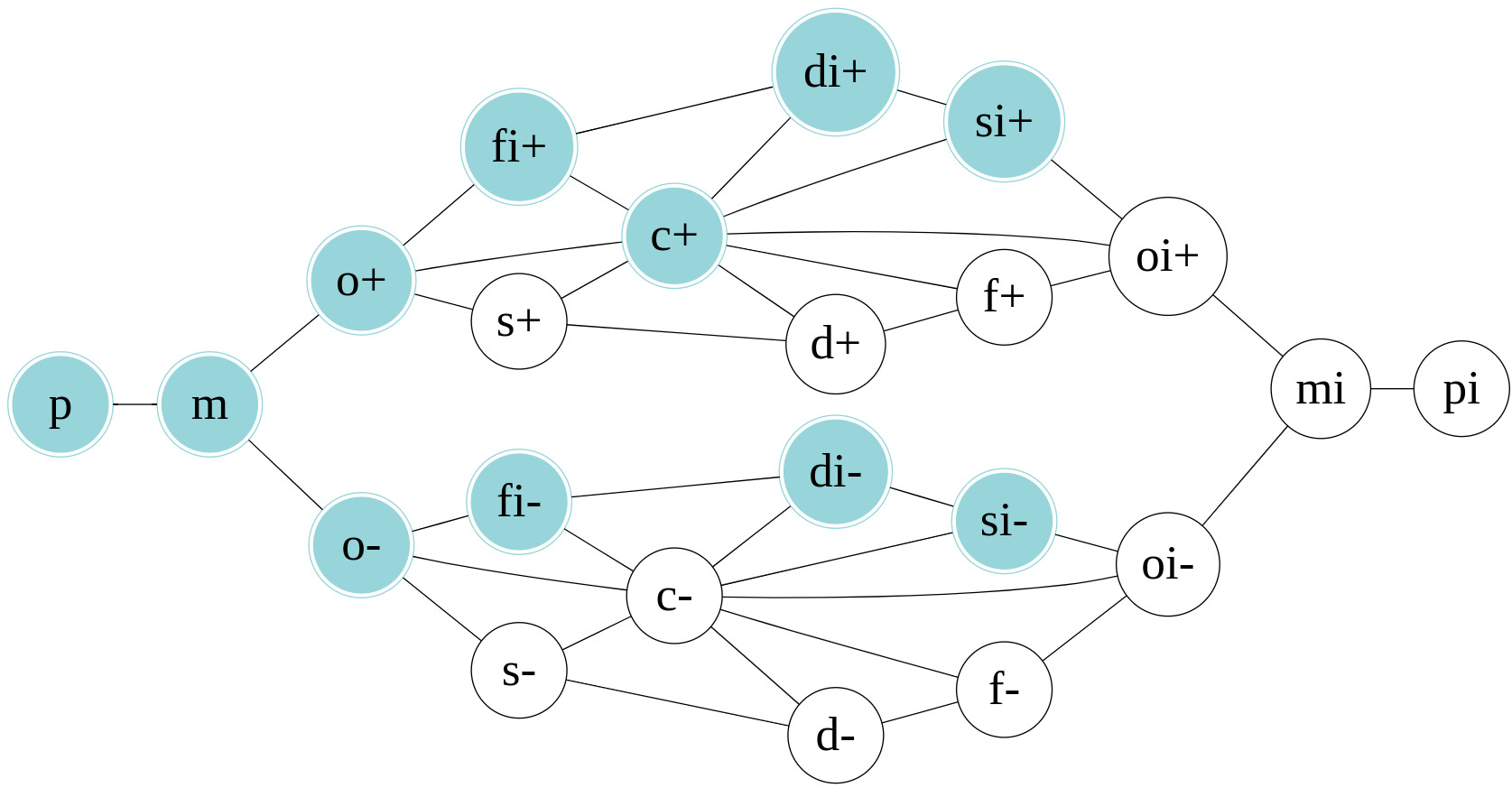}
    \caption{The neighbourhood diagram between the used IOC relations (blue circles).}
    \label{fig:ioc_neighbouring}
\end{figure*}

Moreover, it can be seen from Figure~\ref{fig:landmarkOrder2} that certain areas are spatially adjoining \textcolor{black}{to each other} while others are not. The corresponding \emph{IOC} relations of adjoining areas are therefore conceptual neighbours as they can be directly transformed into one another without encountering any other types of relations~\citep{Freksa1992TemporalRB} when a viewer starts moving. The neighbourhood constraints between the \emph{IOC} relations used in this work are shown in Figure~\ref{fig:ioc_neighbouring}.
For example, the relation \emph{`precedes' ($p$)} and \emph{`meets' ($m$)} are neighbours because viewers can directly go from an area where `\emph{A precedes B}' is observed to an area where `\emph{A meets B}' is observed. However, if a viewer wants to go from the area `\emph{A precedes B}' to another one where \emph{`A overlaps B'} is observed, they will have to go through the area/line where \emph{`A meets B'} is observed. Areas with neighbouring relations are in fact spatially connected as shown in Figure~\ref{fig:landmarkOrder2}.
By establishing the constraints on neighbouring relations, we could possibly further reduce the viewer's locations or trajectory as they start moving and continuously report the observed relations between co-visible landmarks.

\begin{figure*}[ht!] 
	\centering 
	\subfigure[The possible relations between a point landmark and a landmark with non-zero size.]{
	\label{fig:landmarkOrder_IOC2}
    {\includegraphics[width = 0.7\linewidth]{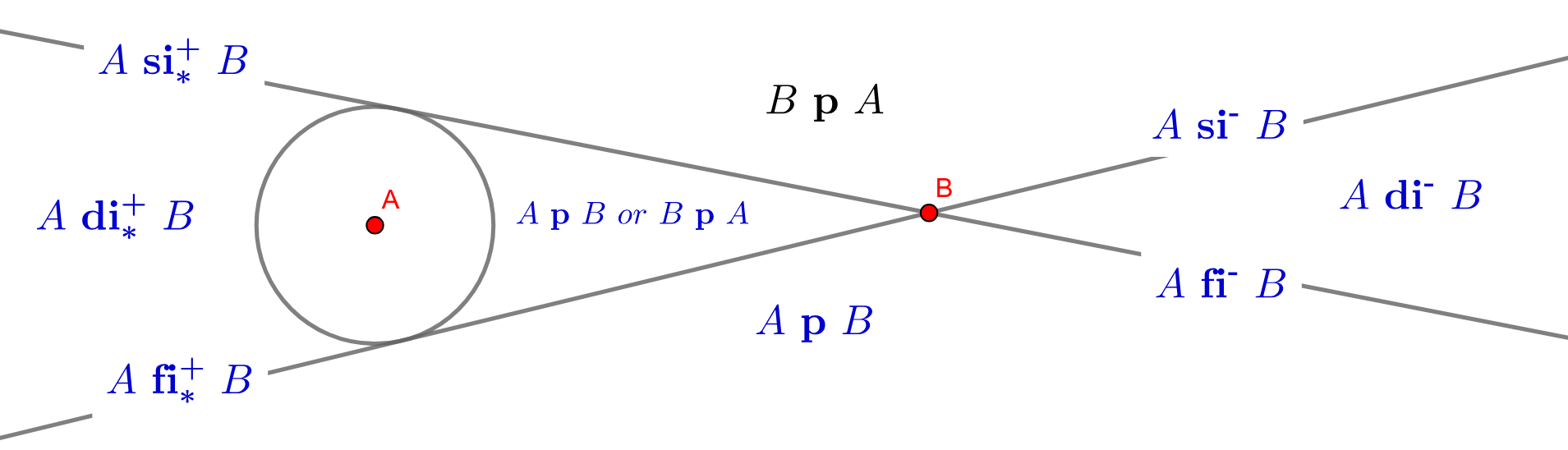}}}
    \subfigure[The possible relations between two point landmarks.]{
	\label{fig:landmarkOrder_IOC3}
    {\includegraphics[width = 0.7\linewidth]{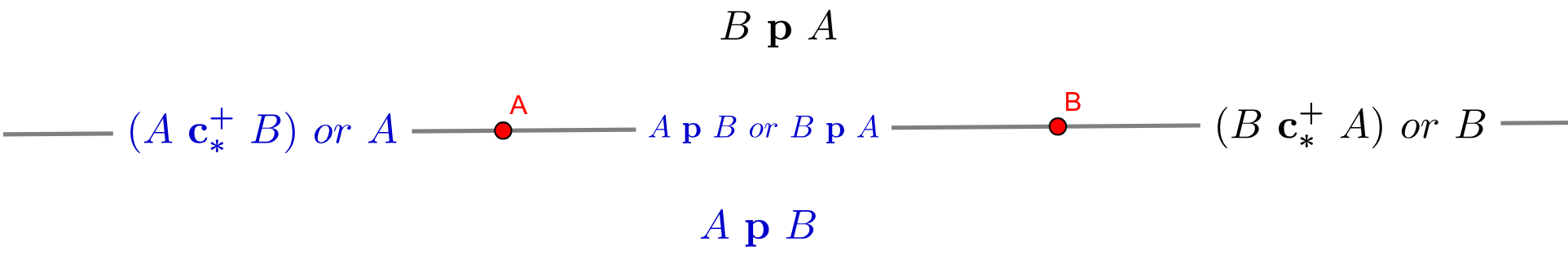}}}
    \caption{The possible relations between two landmarks using different levels of abstraction with points.}
\end{figure*}

\noindent \textbf{Abstraction of landmarks using points}. As shapes can be described using points at various levels of abstraction~\citep{Freksa1992UsingOI}, \textcolor{black}{the above relations (shown in Figure~\ref{fig:landmarkOrder2}) can be further reduced based on these abstractions}: 1) if one of the two landmarks is abstracted to a point, the 18 relations are reduced to eight, as shown in Figure~\ref{fig:landmarkOrder_IOC2}; 2) if both landmarks are abstracted to points, the number of possible relations are further reduced to four\footnote{\textcolor{black}{Note with a reduced set of relations,}
we would only identify the viewer's location much more roughly \textcolor{black}{compared to using the projected intervals as in Figure~\ref{fig:landmarkOrder2}}.}, 
as shown in Figure~\ref{fig:landmarkOrder_IOC3}. When a viewer is on different sides of the line connecting the two points \emph{AB} and facing towards the landmarks, we would expect them to observe the two landmarks in different orders, which is consistent with the observation made in~\citet{Schlieder1993} and~\citet{LEVITT1990305}. 

\begin{figure*}[ht!] 
	\centering 
    \includegraphics[width = 0.7\linewidth]{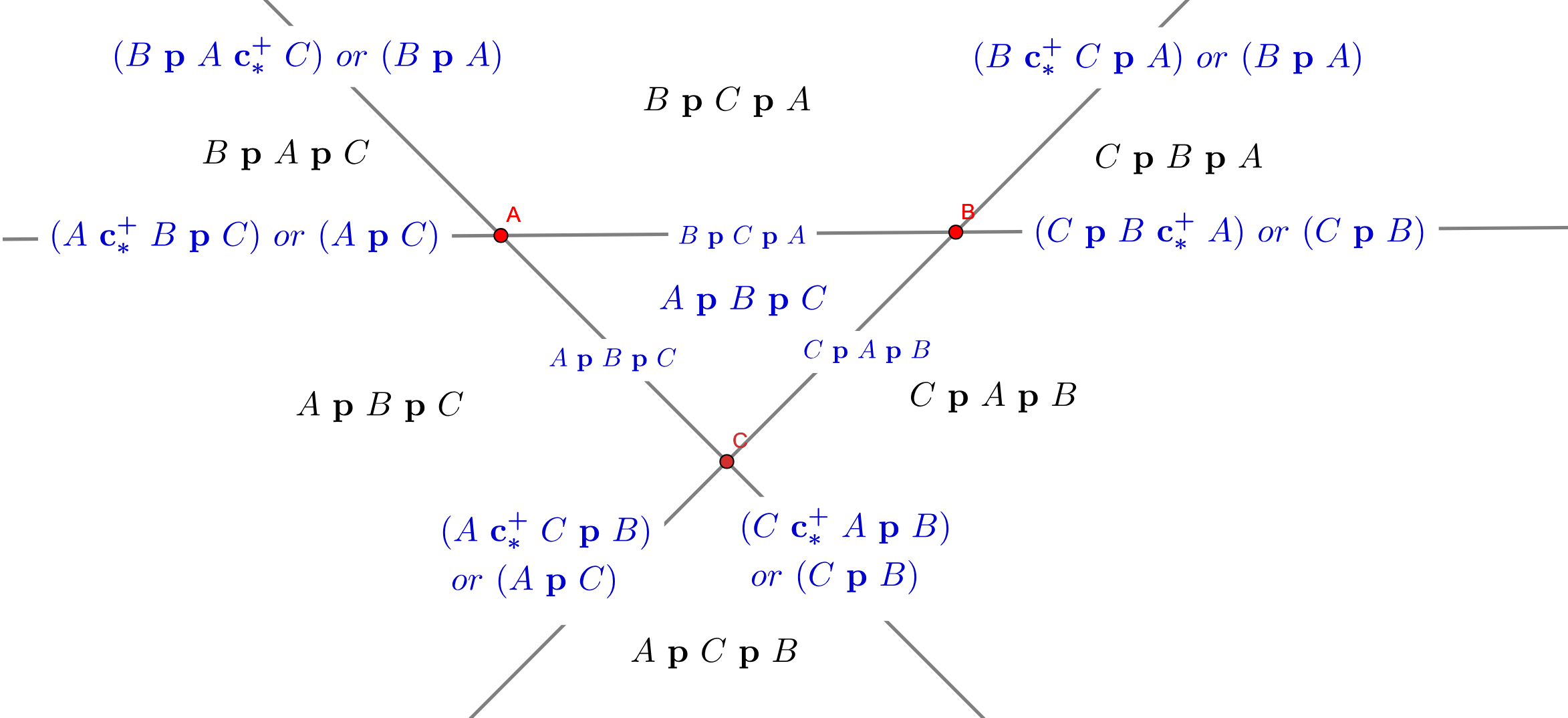}
    \caption{The possible relations of three co-visible landmarks. Note: in the area enclosed by the three landmarks, the observed order of landmarks depends on the initial viewing direction of viewers; while in other areas, we assume viewers have adjusted their orientations so that all landmarks are in the FOV.} 
	\label{fig:landmarkOrder_IOC4}
\end{figure*}

\noindent \textbf{Viewing order of more than two landmarks on a panorama with IOC relations.} When more than two landmarks ($n$) are present, the study space could be divided using $n(n-1)/2$ lines connecting every two co-visible landmarks, or the extremities of landmarks when their extent is considered,
\textcolor{black}{such that landmarks are observed in different orders in each divided area}.

For the simplistic situation when \textcolor{black}{all} landmarks are \textcolor{black}{abstracted as points}, the space could be divided into two parts using the convex hull of these landmarks in clockwise (or anti-clockwise) order. When viewers are outside or on the edge of the convex hull, they will be able to find an optimal viewing direction such that all landmarks are located in their FOV.
A 
unique viewing order of landmarks could then be identified for each of these areas (Figure~\ref{fig:landmarkOrder_IOC4})\footnote{Matlab code is available at \url{https://bitbucket.org/lwei2020/qualitativeplacesignatures} \emph{2signatures\_noSize.m}.}. When viewers are inside the convex hull, they could start from any landmark so the observed order of these landmarks might be rotated, which means the observed sequences of landmarks are cyclically equal to each other, such as \emph{ABC}, and \emph{BCA}. In this work one such rotation sequence is created as a way to reference each area
\footnote{Note that such references may not be unique.}. 
Similarly, when the size of landmarks are considered, their relative \emph{IOC} relations are identified by finding the left and right extremities of each landmark from a viewpoint, and comparing the directions of corresponding tangent pairs ($\alpha_i^l,\alpha_i^r$) in clockwise order\footnote{Code is available at \url{https://bitbucket.org/lwei2020/qualitativeplacesignatures} \emph{2signatures\_size.m}.}.


\begin{figure*}[ht!] 
	\centering 
	\subfigure[The size of landmarks is comparable to their distance apart.]{
	\label{fig:close_landmarks}
    {\includegraphics[width = 0.4\linewidth]{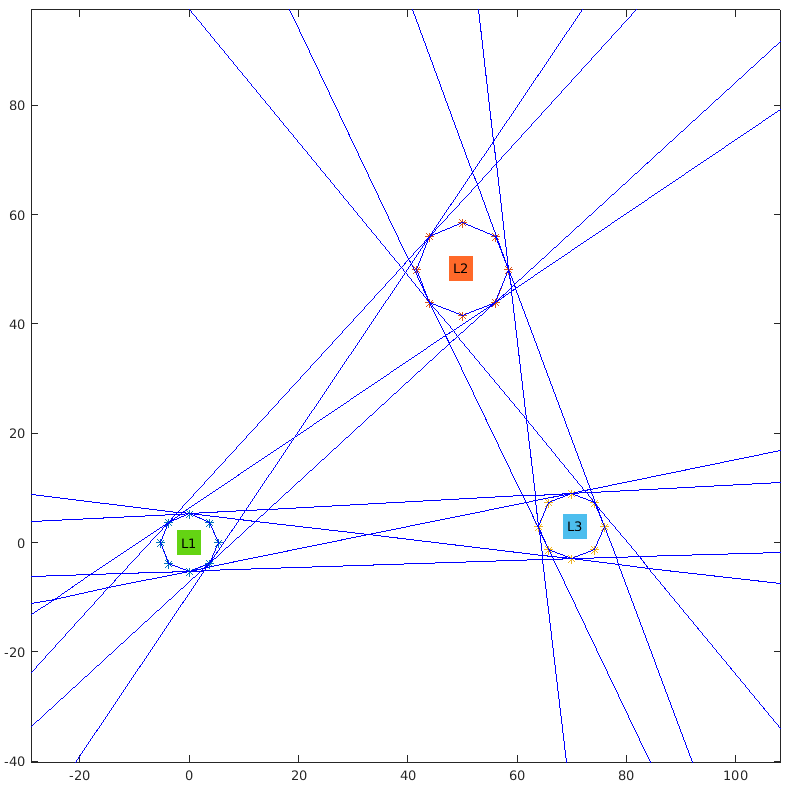}}}
    \subfigure[The size of landmarks is relatively small.]{
	\label{fig:apart_landmarks}
    {\includegraphics[width = 0.4\linewidth]{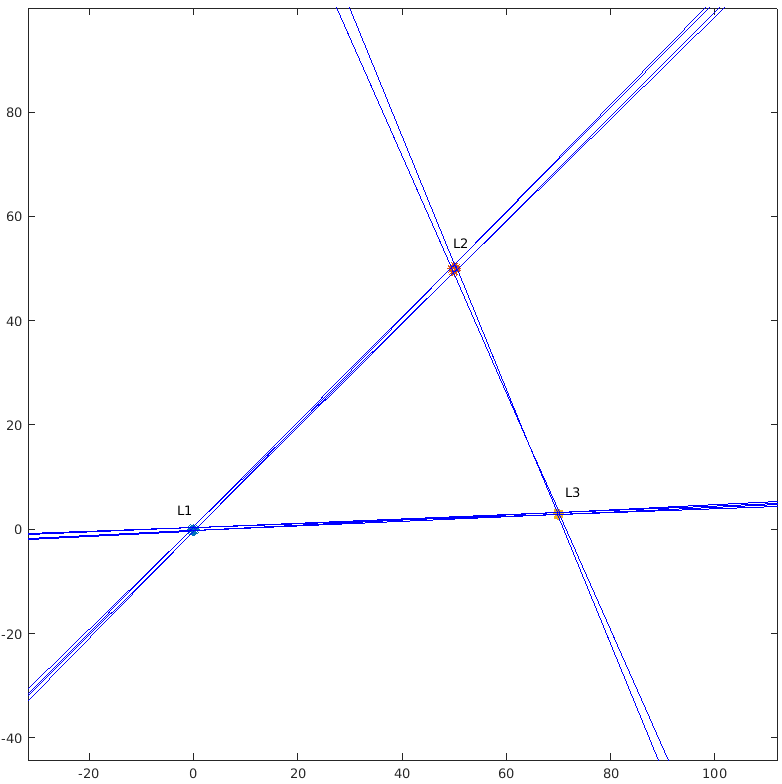}}}
    \caption{Space division using landmarks \textcolor{black}{with different distances apart}. Note that the distances between landmarks in the two figures are the same but their sizes are reduced in the second figure.}
    \label{fig:close_apart_landmarks}
\end{figure*}  

Whether to consider the size of landmarks will depend on the data available and the specific application. \textcolor{black}{For example, when the extent of landmarks is comparable to their distance apart, as shown in Figure~\ref{fig:close_landmarks}, and the specific purpose of an application is to navigate robots for visual inspection, such as for street infrastructure in urban environments~\citep{PEEL2018244}, considering the extent of landmarks and the occlusion information can differentiate locations at a much finer level, especially for those areas close to the lines connecting the centroids of landmarks.}
On the contrary, if the extent of landmarks is relatively small compared to their distance apart such as shown in Figure~\ref{fig:apart_landmarks}, and the purpose of an application is to roughly identify the initial location of a viewer, using the point representation and dividing the space with lines connecting each point pairs would be enough, as areas near those lines would be small anyway.

\begin{figure*}[htp!] 
	\centering 
	\subfigure[The same sequential relations are observed in both locations but not the first and last elements.]{     \label{fig:example_reasoning}
    \includegraphics[width = 0.47\linewidth]{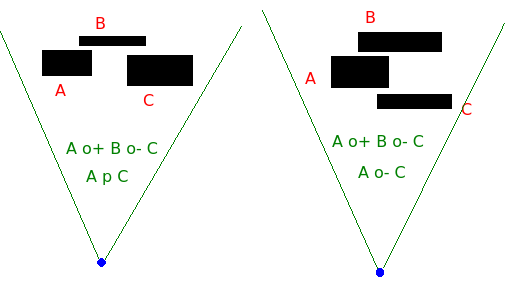}}
    \subfigure[Different sequential relations are observed in the first and second elements but if B is not observed, the two locations are indistinguishable.]{
    \label{fig:example_reasoning2}
    \includegraphics[width = 0.47\linewidth]{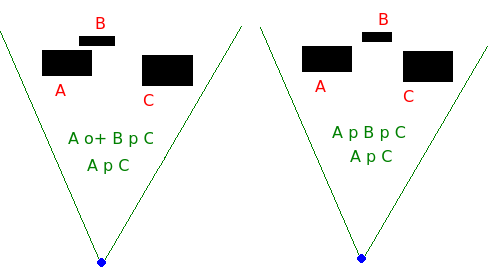}}
    \caption{Examples of different observed landmark configurations.}
    \label{fig:example_reasonings}
\end{figure*}
 
Another question is whether to encode the full set of relations between every ordered pair of landmarks or only the relations between consecutively observed landmarks.
For example, for the four scenarios shown in Figure~\ref{fig:example_reasonings}, if we record the ordering of landmarks in each scenario based on their left extremities, the ordered sequences of landmarks would be $\langle ABC \rangle$ 
for all scenarios. A viewer would not be able to differentiate their location based on this ambiguous description.
However, if we store the sequential IOC relations of these landmarks based on their size and relative distance to the viewer, the observed place signatures would be \emph{$\langle Ao^+Bo^-C\rangle$} for the first two scenarios,
\emph{$\langle Ao^+B\;p\;C\rangle$} for the third scenario, and  \emph{$\langle A\;p\;B\;p\;C\rangle$} for the last scenario. These signatures are obviously more discriminative than the previous descriptions.

But still, the descriptions of the first two scenarios are the same while actually they are different: the relation between A and C are respectively $\langle A\;p\;C\rangle$ and $\langle Ao^-C \rangle$ in the two scenarios. We can also see from the \emph{composition table} shown in Table~\ref{tab:transitivity} that based on the relations between \emph{A-B} and \emph{B-C} there are four possible relations between \emph{A-C}: \emph{\{$p,m,o^+,o^-$\}}. 
Therefore, to achieve a better discriminative ability, ideally, it would be useful to encode the relative relations between all ordered pairs of landmarks.
{When landmarks are all points, only relation $p$ and $c^+$ are needed, and \textcolor{black}{unique relations can be inferred from any combination of the two (highlighted cells in Table~\ref{tab:transitivity})}. Therefore, storing the relations between adjacent landmarks would be enough as no ambiguity will be caused by these relations.}



In the following sections, two other components of the proposed signature are introduced to further increase the `resolution' of location description.

\subsection{Adding Relative Orientations between ordered pairs of landmarks}
\label{sec:ori1} 
\textcolor{black}{The second component of the proposed place signature is the sequence of the relative orientations between ordered pairs of landmarks}. The concept of relative orientations was originally proposed by \citet{Freksa1992UsingOI} to describe the location of a point with respect to two other points using the left/right and front/back dichotomies of 15 disjoint combinations of orientation relations. 

\begin{figure*}[ht!] 
	\centering 
	\includegraphics[width = 0.3\linewidth]{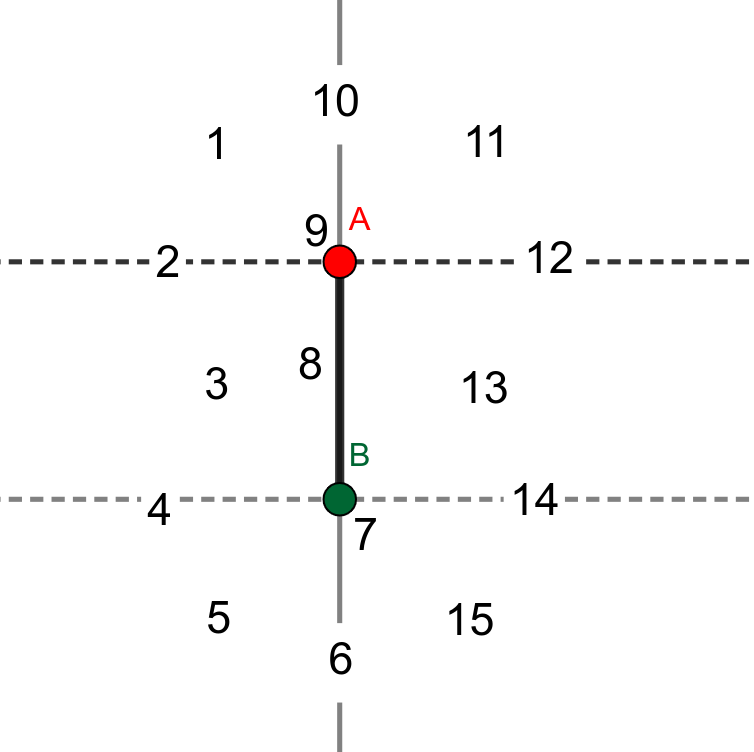}
    \includegraphics[width = 0.6\linewidth]{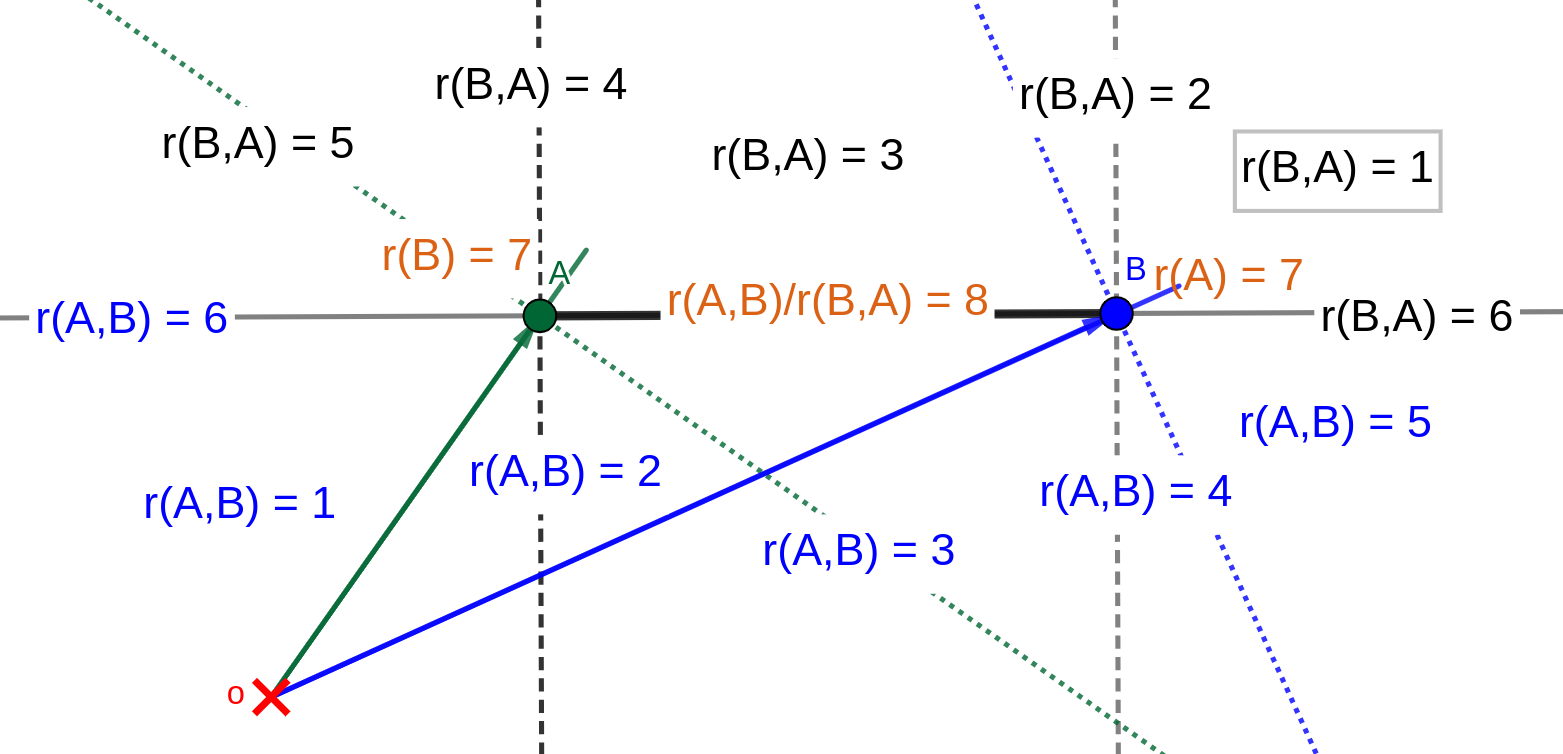}
    \caption{\textcolor{black}{The left figure shows how the location of an object can be described based on its left/right and front/back dichotomy w.r.t point A and B, modified from~\citep{Freksa1992UsingOI}.
    The right figure shows the subdivision of space from where different relative orientation relations can be observed between landmark A, B and the viewer.}
    \label{fig:freksa92}
    }
\end{figure*}   
For example, as shown in Figure~\ref{fig:freksa92} (left), objects in \textbf{area \emph{1}} are on the \emph{left-front} (LF) of point A with respect to vector $\vec{BA}$ and on the \emph{right-back} (RB) of point B w.r.t. vector $\vec{AB}$; objects in area \emph{2} are on the \emph{left-neutral} (LN) of A and \emph{right-back} of B; objects in area \emph{3} are on the \emph{left-back} (LB) of A and \emph{right-back} of B; objects in area \emph{4} are on the \emph{left-back} of A and \emph{right-neutral} (RN) of B; objects in area \emph{5} are on the \emph{left-back} of A and \emph{right-front} (RF) of B; objects in area \emph{6} are on the \emph{straight-back} (SB) of A and \emph{straight-front} (SF) of B; objects in area \emph{7} are on location B and to the \emph{straight-back} (SB) of A; objects in area \emph{8} are on the \emph{straight-back} of both A and B. Relations for the areas on the other side of the line AB (area 15, 14, 13, 12, 11, 10, 9) are similar to those in the area [1-7] except that \emph{A} and \emph{B} are switched.

Using the same division of space with the two perpendicular
lines of line \emph{AB}, one passing through \emph{A} and the second through \emph{B}, different groups of relative orientations can be observed by viewers $o$ with respect to $(A,\vec{oA}$) and $(B,\vec{oB})$ when they are located in different areas. For example, as shown in Figure~\ref{fig:freksa92} (right), 
imagine a viewer is in area \emph{1} and facing towards the two landmarks, they will observe \emph{B} on the \emph{right-front} of A by imagining a vector $\vec{oA}$ and its perpendicular line passing A; similarly, they will observe $A$ on the \emph{left-back (LB)} of B by imagining a vector $\vec{oB}$ and its perpendicular line passing B. 
Additionally, as landmarks are counted from left to right in this work and the nearer one is always considered first when multiple landmarks coincide, only eight of the original 15 orientation relations are needed in this work. 
More specifically, \textcolor{black}{the observed relations of \emph{B} with respect to $\vec{oA}$, and \emph{A} with respect to $\vec{oB}$} from the eight indexed area in Figure~\ref{fig:freksa92} (right) are respectively: {1 (RF, LB), 2 (RN, LB), 3 (RB, LB), 4 (RB, LN), 5 (RB, LF), 6 (SF, SB), 7 (SB), and 8 (SB, SB)}.

\begin{figure*}[ht!] 
	\centering 
    \includegraphics[width = 0.6\linewidth]{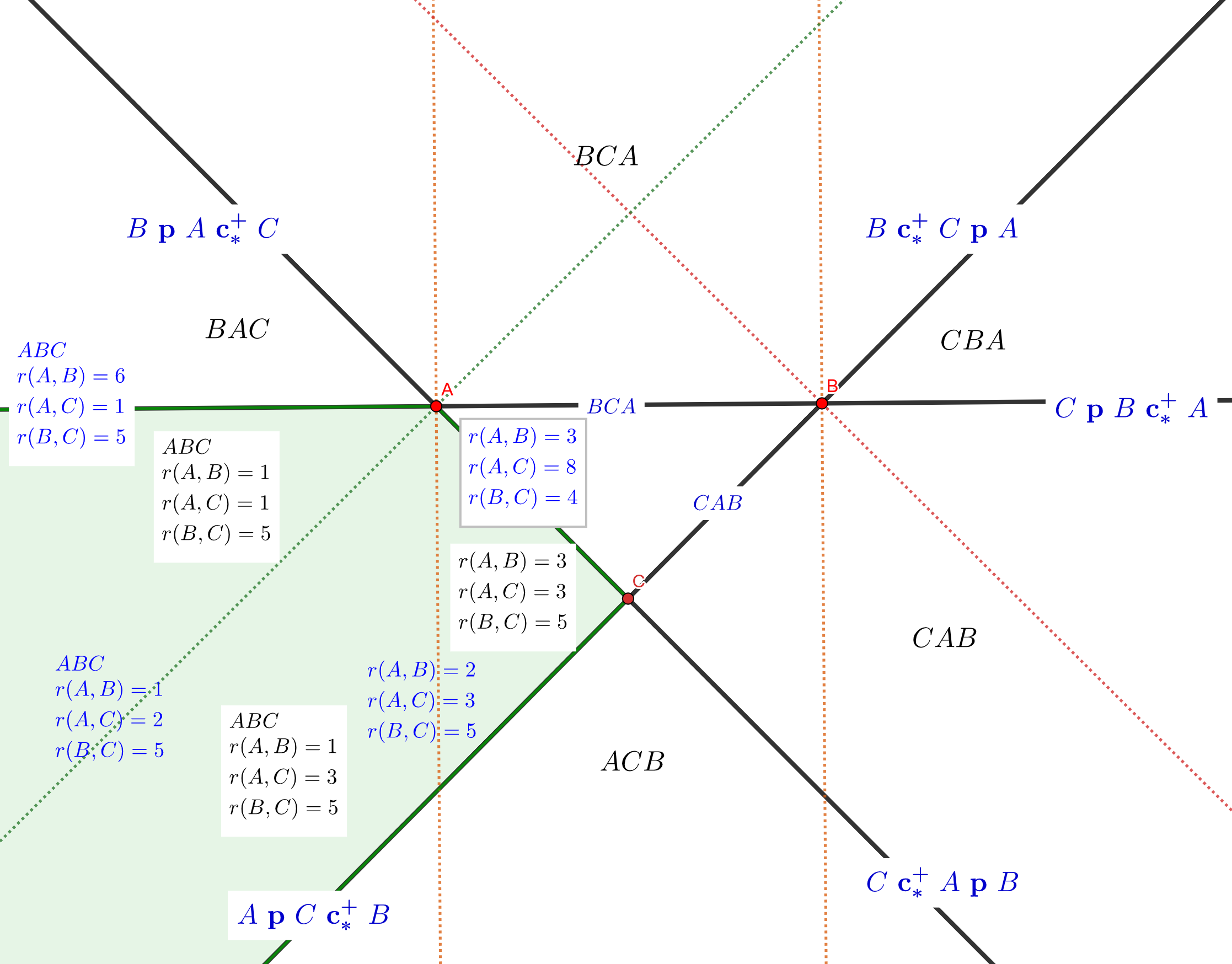}
    \caption{The refined space subdivision based on the ordering and relative orientation information of three co-visible point-like landmarks. Note that lines \emph{AC, BC} are perpendicular in this simple scenario. }
	\label{fig:landmarkOrder_IOC5}
\end{figure*}

\noindent \textbf{Usage in practice}.
Although it seems a bit tedious to define the exact relations observed in each area, in practice, after deciding the viewing order of two co-visible landmarks, i.e. \emph{r(A,B)} or \emph{r(B, A)}, a viewer will only need to select one index between $1$ and $8$ to describe their situated area. Or, to make the task even simpler, viewers will only need to describe whether they are situated \emph{on the left}, \emph{between}, or \emph{on the right} of the two ordered parallel perpendicular lines by selecting from \emph{\{1, 3, 5\}}. We will then be able to roughly differentiate their situated area.

When there are $n$ co-visible landmarks ($n\geq2$), the space previously divided using $n(n-1)/2$ lines connecting every two co-visible landmarks (Figure~\ref{fig:landmarkOrder_IOC4}) can be further divided using their perpendicular lines\footnote{In the case that the extent of landmarks are considered, the right extremity of the first seen landmark and the left extremity of the following landmark can be used to identify their relative orientations.}, totalling in $3n(n-1)/2$ dividing lines. The number of final areas will depend on the configuration of co-visible landmarks. For example, for the simple case shown in Figure~\ref{fig:landmarkOrder_IOC5} with three landmarks and line \emph{AC} and being perpendicular to \emph{BC}, the area marked in green where landmarks \emph{ABC} are sequentially observed can be further divided into seven areas (three regions and four on lines) \textcolor{black}{each annotated
with different combinations of relative orientation indices}.

\begin{figure*}[ht!] 
	\centering 
    \subfigure[Inferred relative closeness of two landmarks A and B.]{\label{fig:freksa92_2_middle}\includegraphics[width = 0.4\linewidth]{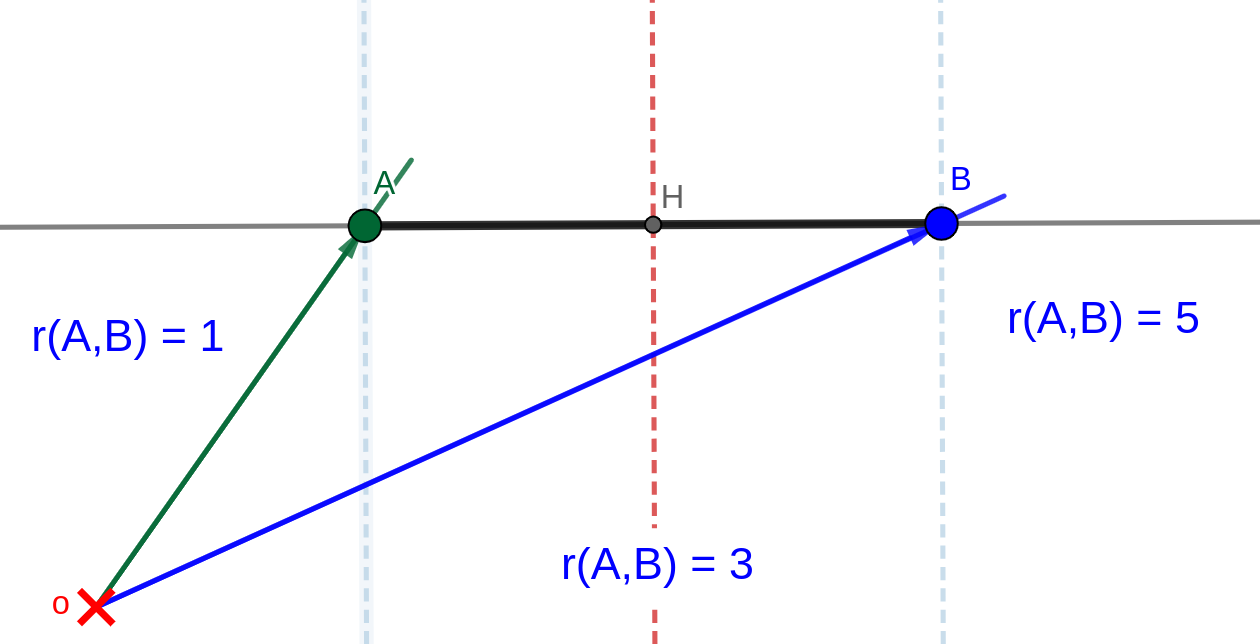}}
    \subfigure[Differentiating between different qualitative distances.]{\label{fig:freksa92_4}
    \includegraphics[width = 0.4\linewidth]{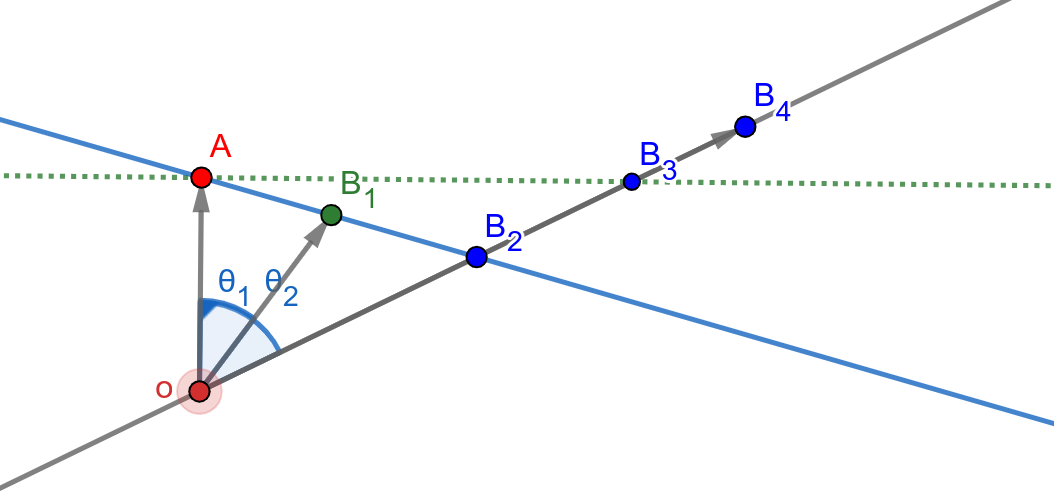}}
    \caption{a) The space is divided by the perpendicular line crossing the middle point of line segment \emph{AB}. A viewer on the left-hand side of the middle line will be closer to the first seen landmark while a viewer on the other side will be closer to the following landmark; b) The relative orientations can distinguish between four locations from where $A-B_1$, $A-B_2$, $A-B_3$ and $A-B_4$ are respectively observed.}
\end{figure*}


One advantage of using \emph{relative orientation} relations is that they are able to distinguish between different qualitative distances. For example, as shown in Figure~\ref{fig:freksa92_2_middle}, a viewer on the left-hand side of the perpendicular line crossing the middle point of line segment will always be closer to the first seen landmark while a viewer on the other side will be closer to the following landmark. 

Another example is shown in Figure~\ref{fig:freksa92_4}. \textcolor{black}{Assuming a landmark of type \emph{A} and one of type \emph{B} are co-visible in three different locations $o$\footnote{They are drawn at the same location in this figure for illustration purposes}, the relative location between $A$ and the viewer are the same in these scenarios, while the location of $B_2$, $B_3$ and $B_4$ are different}, 
then it would be possible for us to differentiate between these locations as the indices of relative orientations between $A-B_2$, $A-B_3$ and $A-B_4$ are respectively \emph{3 (RB-LB)}, \emph{2 (RN-LB)}, and \emph{1 (RF-LB)}; we can also infer that $B_2$ is the closest one to the viewer among the three.
However, these relations do not resolve the orientation information more finely~\citep{Freksa1992UsingOI}. For example, assume there is another location from where $A-B_1$ could be seen. The same configuration of relative orientations could be observed between $A-B_1$ and $A-B_2$, though the angles $\theta_1$ and $\theta_2$ between the lines of sight of the two landmarks are different.
Therefore, in the next section, the angles between the lines of sight of ordered landmarks are introduced as the third component of the proposed place signature.

\subsection{Adding qualitative angles between the lines of sight of ordered landmark pairs}
\label{sec:ori2}

\begin{figure}[htp] 
    \centering  
    \includegraphics[width = 0.7\linewidth]{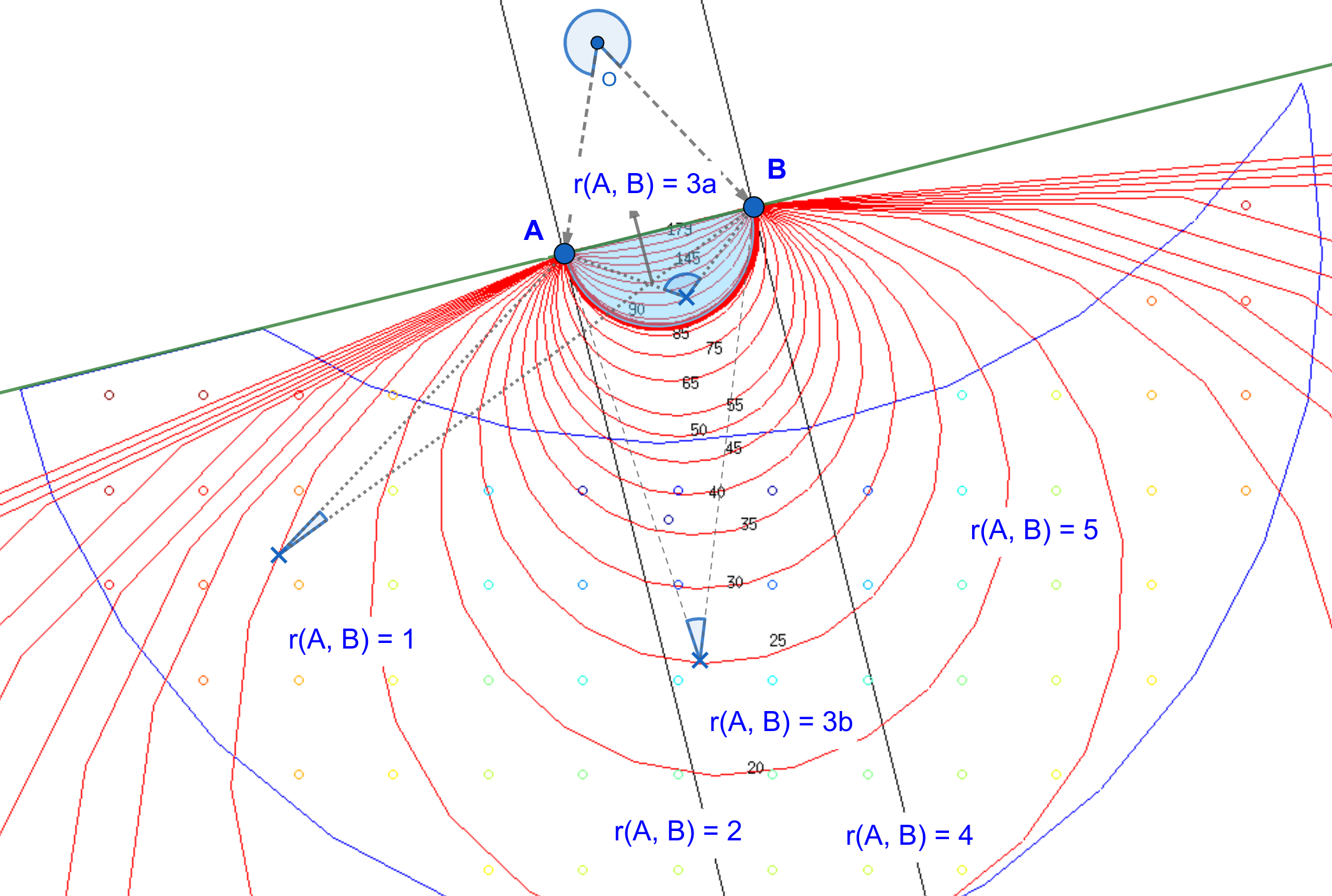}
    \caption{The locations from where constant angles $\theta_{ij}$ can be observed between two landmarks are plotted as red contour lines with the corresponding angles marked. Note that when the angle is closer to $180^o$, the contour lines get much denser so they are drawn for every $5^o$ (when $\theta\leq55^o$) and $10^o$ (when $\theta\geq65^o$) for illustration purpose. Note: the area outlined in blue is an exemplar place cell before further division.}
    \label{fig:constant_angle}
\end{figure}

As previously suggested by~\citet{LEVITT1990305}, the set of points (i.e. viewers' locations) from where a constant angle $o^o\leq \theta_{ij}\leq180^o$ can be observed between the lines of sight of two landmarks is constrained to circular arcs in 2D space, which can be plotted as contour lines as shown in Figure~\ref{fig:constant_angle} with the corresponding angles marked in black\footnote{Note for visualisation purposes, the contour lines are plotted for every $5^o$ when angles are less than $65^o$, and every $10^o$ when angles are above.}. It can be seen from this figure that: 
\begin{itemize}
    \item when $\theta_{ij}=0^o$ or $180^o$, the viewer must be co-linear with line $AB$\footnote{\textcolor{black}{Note that in this situation, the relative orientation and angular information can be used to infer each other uniquely, i.e. $(r=6)\longleftrightarrow(\theta_{ij}=0^o)$, $(r=8)\Longleftrightarrow(\theta_{ij}=180^o)$.}};
    \item \textcolor{black}{when $\theta_{ij}=90^o$, the \textcolor{black}{corresponding contour line of $\theta_{ij}$} is the half-circle centred at the middle point of segment \emph{AB}. The corresponding contour lines of $90^o< \theta_{ij}\leq180^o$ are always between this half-circle and segment $AB$, and also between the two perpendicular lines of line \emph{AB} passing through point \emph{A} and \emph{B};}
    \item \textcolor{black}{If viewers approach line $AB$ by moving along one of its perpendicular lines that pass between A and B, then $\theta_{ij}$ will get continuously closer to $180^o$; but if the perpendicular is outside the two lines passing through A and B, then the observed angle can go larger, and then smaller, as one approaches the lines AB.
    }
    The corresponding circular arc\textcolor{black}{/contour line} becomes quite large and the gap between two nearby arcs becomes wider, suggesting a higher ambiguity of viewers' possible location(s) between the two contour lines. 
\end{itemize}

Since humans are often not good at judging the exact value of an angle and it is not always feasible to use a separate measuring tool, qualitative angles between the lines of sight of landmarks are used in this work by judging approximately whether an angle is acute ($0^o<\theta<90^o$), obtuse
($90^o<\theta<180^o$)~\citep{Latecki93orientationand}, \textcolor{black}{or exactly 90 degrees}. \textcolor{black}{With this strategy}, the region between the two perpendicular lines (Figure~\ref{fig:constant_angle}) with relative orientation $r(A,B)=3$ is further divided:
\begin{enumerate}
    \item if the observed angle between the two landmarks is \textbf{\emph{obtuse}} (noted as $1$), then the viewer must be in area $3a$; 
    \item if the observed angle between the two landmarks is \textbf{\emph{acute}} (noted as $0$)
    and their relative orientation is $r(A,B)=3$, then the viewer must be in area $3b$; 
    \item if the observed angle is exactly 90 degrees, then the viewer must be on the half-circle $AB$ centred at the middle point of segment \emph{AB}.
\end{enumerate} 

By combining the above three types of spatial relations, $4n(n-1)/2$ lines are used to divide the space if all ordered pairs are considered. This number is reduced to $4(n-1)$ if only adjacent pairs are considered.
Each such divided area is called a \emph{place cell} and 
has an associated
place signature\textcolor{black}{, consisting of three ordered sequences of qualitative relations between co-visible landmarks}. For example, for the place cell (area outlined in blue) shown in Figure~\ref{fig:constant_angle}, \textcolor{black}{the left bottom area} has a signature
\emph{S = (AB, p, 1, acute)}, where
\emph{(AB, p)} \textcolor{black}{means that}
\emph{A proceeds B}, \emph{r(A,B)=1} \textcolor{black}{means that} \emph{B is on the right-front of A and A is on the left back of B}, and \emph{`acute'} \textcolor{black}{means that} the observed relative angle between A and B is less than $90^o$.

\subsection{Hedging place signatures in cyclic order}
\label{sec:augment}

\begin{figure}[ht!] 
    \centering  
    \subfigure[The place cells created by the straight lines connecting the three landmarks. \textcolor{black}{The enclosed area (green) has an observed sequence $\langle BCA \rangle$, i.e. one of the permutations of the landmark ordering.}
    ]{\label{fig:example1}{
    \includegraphics[width = 0.45\linewidth]{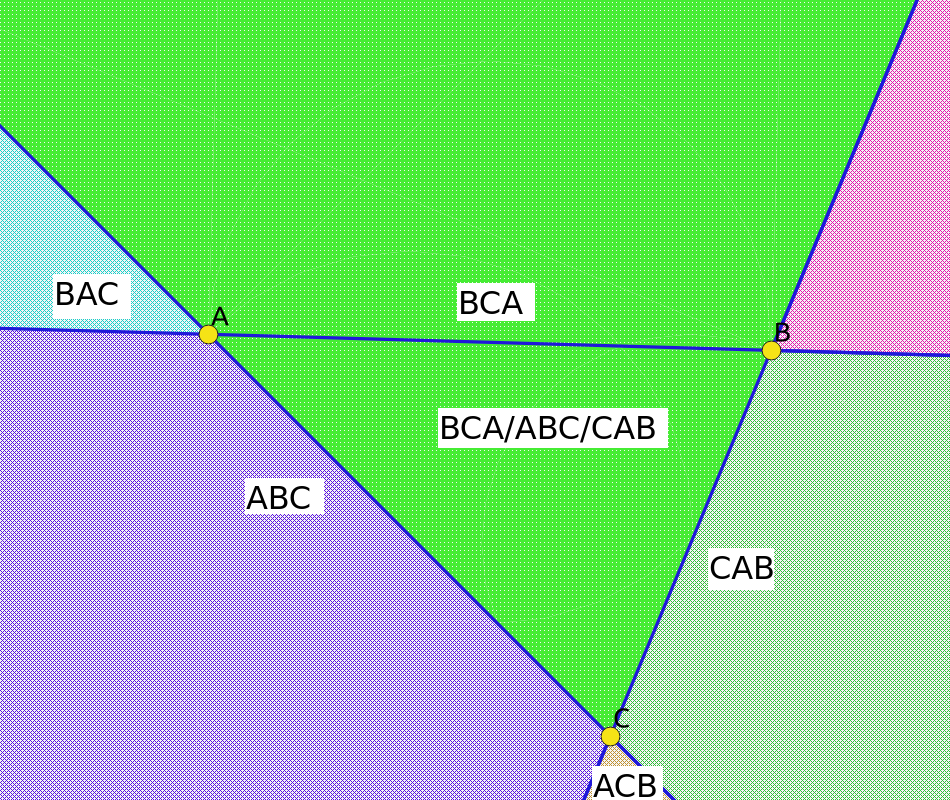}}}
    \subfigure[The place cells created by the straight lines and perpendicular lines of the three landmarks. The areas marked in green are inseparable with signature $\langle BCA, 333 \rangle$.]{\label{fig:example2}{
    \includegraphics[width = 0.45\linewidth]{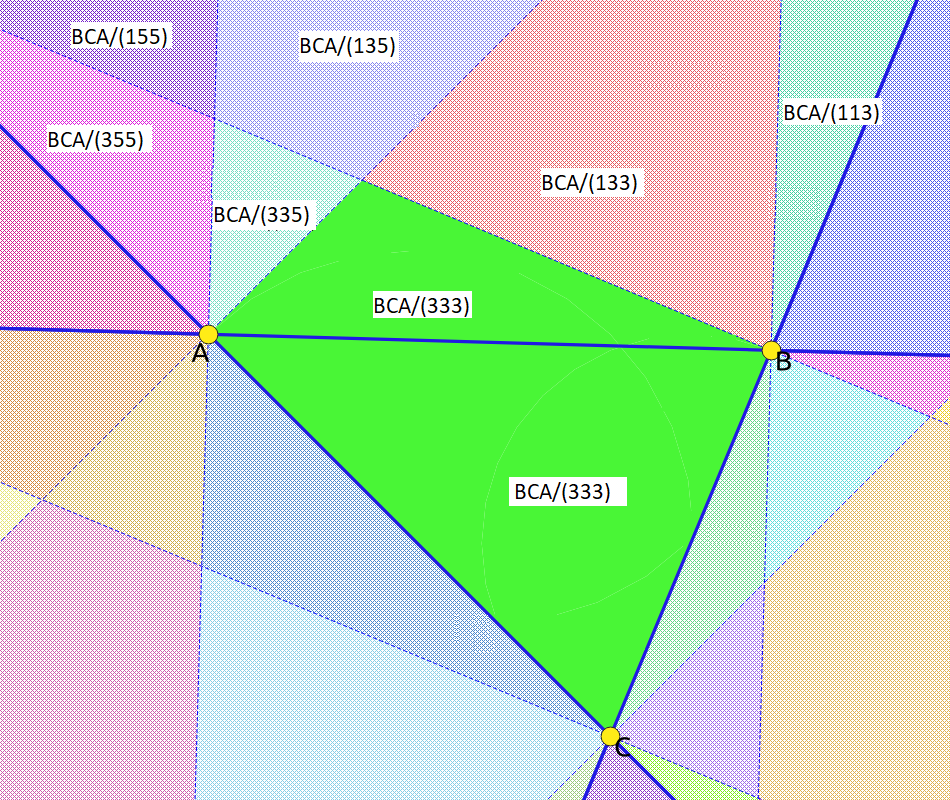}}} 
    \subfigure[\textcolor{black}{After adding the circular lines, the enclosing and outside parts of the green and blue area are still inseparable.}]{\label{fig:example3}{
    \includegraphics[width = 0.45\linewidth]{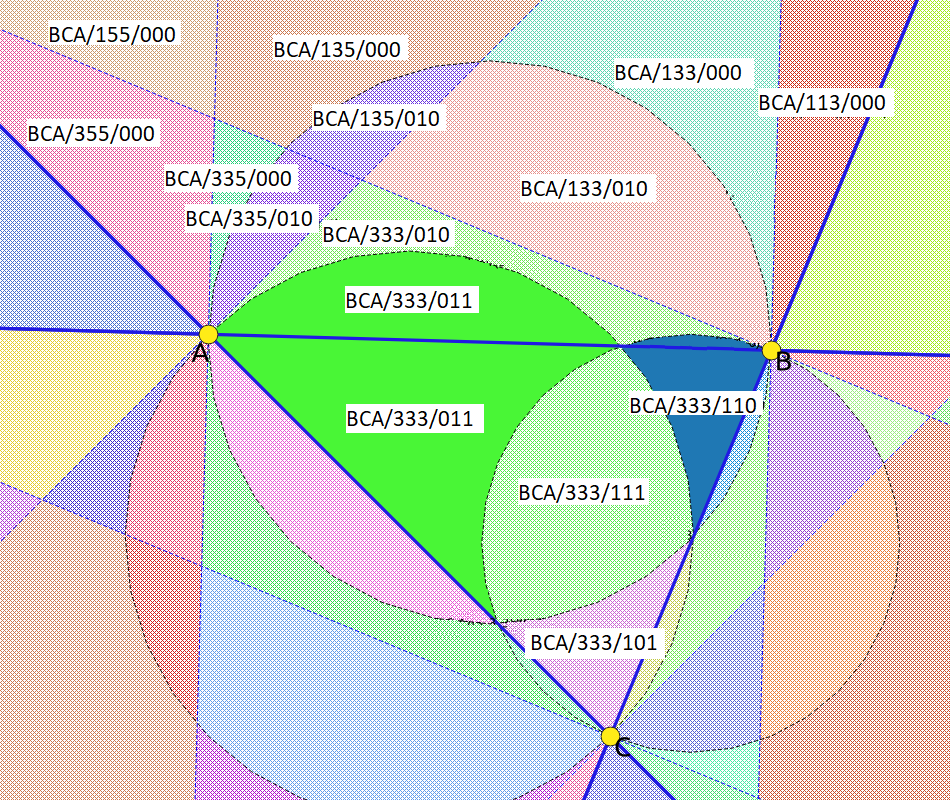}}} 
    \subfigure[\textcolor{black}{After adding the enclosing index, the ambiguous areas are separated.}]{\label{fig:example4}{
    \includegraphics[width = 0.45\linewidth]{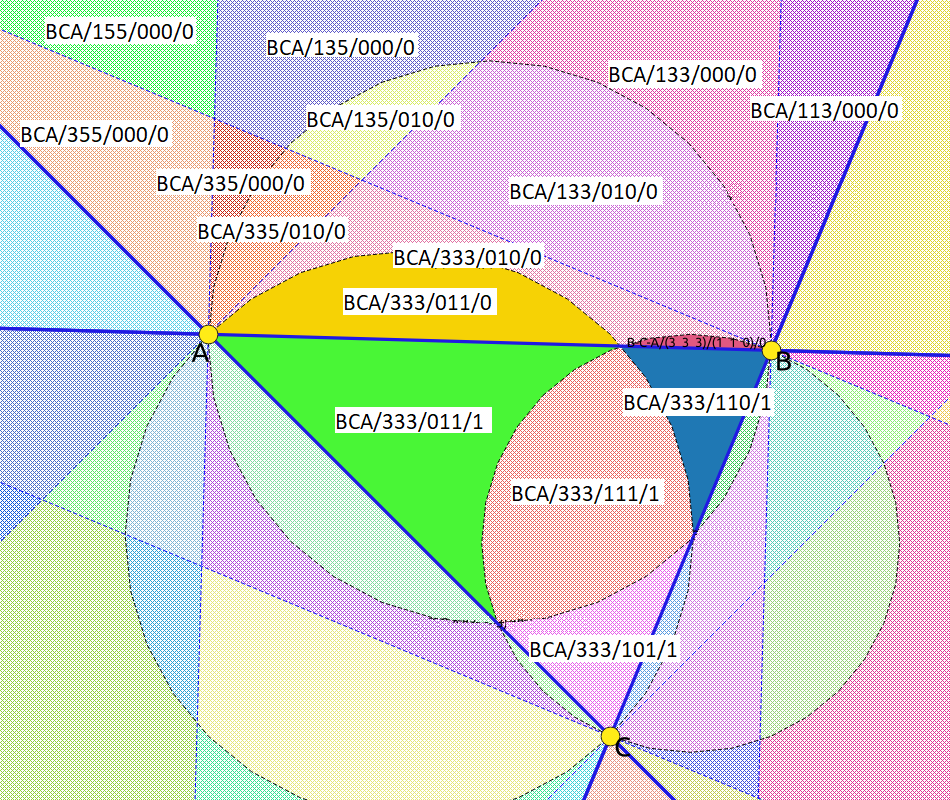}}}
    \caption{An example of place cells created using different combinations of dividing lines. Note the shown relative orientations and qualitative angles are given for landmark pairs in the order of 1-2, 1-3, and 2-3.}
    \label{fig:illus1}
 \end{figure}
 
In certain locations, there exist a best viewing direction(s) such that all landmarks are located in the FOV, thus
a
unique place signature can be observed.
In other locations, viewers may find landmarks are distributing around them and can start from any direction before turning clockwise, so the observed sequences of landmark types and relations could be rotated. For example, the sequences seen from the enclosed middle area in Figure~\ref{fig:example1}
could be $\langle ABC \rangle$, $\langle BCA \rangle$, or $\langle CAB \rangle$, the same as from the adjacent regions\footnote{Note the starting element of these rotated sequences will only depend on the viewing direction not the exact location in the enclosed region. If we assume a viewer inside the enclosed area is facing \emph{B}, then $\langle BCA \rangle$ will be observed, the same as the one from the adjacent top area marked in green.}.

To distinguish between the adjacent areas with same sequences of landmarks, \citet{Schlieder1993} proposed to augment each landmark sequence with the complementary directions 
of all landmarks (as illustrated in Appendix-Figure~\ref{fig:augVec})\footnote{\textcolor{black}{In more detail, given a list of co-visible landmarks $P_1 \cdots P_n$ from a location $o$, the panorama is defined based on the $2n$ directed lines of $\vec{oP_i}$ and $\vec{P_io}$, and encoded as a sequence of upper-case letters (for the original landmarks) and lower-case letters (for the complementary directions of the original landmarks) in clockwise circular order.}}. But this method can only distinguish between adjacent areas when a similar heading direction is assumed as a viewer moves across regions, which is reasonable for landmarks-guided robot navigation. 
In our work, there is no constraint on viewers' starting directions and no movement between regions is strictly required, so this method is not applicable. 

\textcolor{black}{Instead, by considering the relative orientations of landmarks using perpendicular lines, the ambiguous non-enclosing area is first refined (Figure~\ref{fig:example2}) though the enclosed area is still indistinguishable (highlighted in green); then,by considering the qualitative angles between landmarks using half-circles, part of the enclosed area are separated (Figure~\ref{fig:example3}) except for those directly adjacent to the boundary (highlighted in bright green and dark blue).
}
A simple solution suggested is to flag each region based on the distribution of the visible landmarks with respect to the viewer, which is considered as the last component of place signatures. For example, a region is flagged as $enclosed = 0$ if visible landmarks are all located on one side of the viewer (i.e. the clockwise angle between the first and last landmarks is less than $180^o$); and flagged as $enclosed = 1$ if visible landmarks are distributed around the viewer (i.e. the clockwise angle between the first and last landmarks is greater than $180^o$). 
It can be seen in Figure~\ref{fig:example4} that all ambiguous adjacent areas are separated with the suggested solution. 


\subsection{Practical steps for creating and managing a reference database of place signatures}
\label{sec:database}

\begin{figure*}[htbp] 
	\centering 
    \subfigure[Occlusion.]{
    \label{fig:mv_visibility}
    \includegraphics[width = 0.25\linewidth]{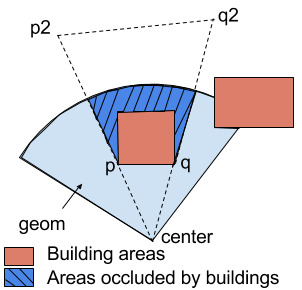}  } 
    \subfigure[Storing cyclically equal place signatures.]{
    \label{fig:rotation}
    \includegraphics[width = 0.65\linewidth]{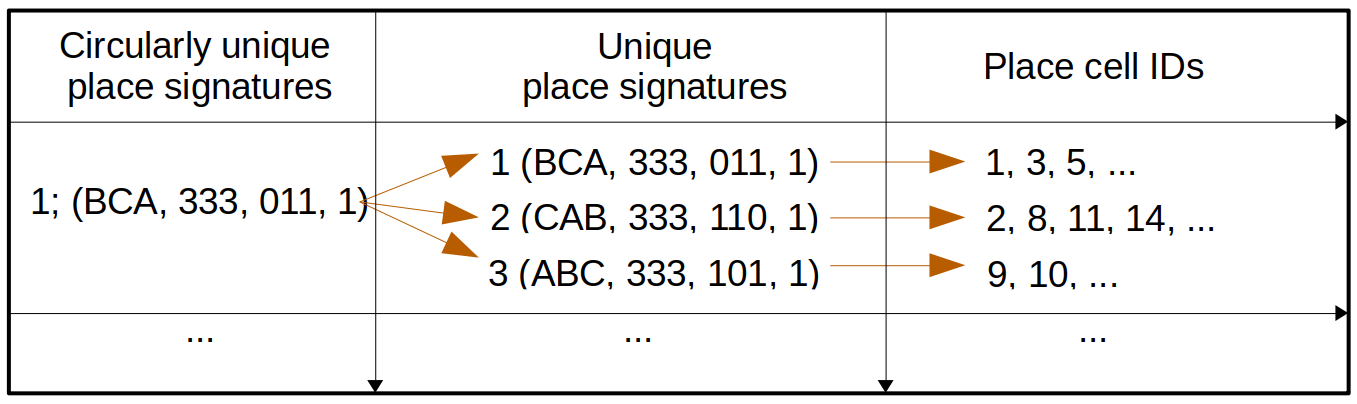}
    }
    \caption{a) removing the visibility area of landmarks occluded by buildings; b) storing cyclically equal place signatures (any rotated version could be stored as a representative).}
\end{figure*}

\noindent \textbf{Steps for creating a reference database of place cells and signatures}\footnote{Exemplar code is available in \url{https://bitbucket.org/lwei2020/qualitativeplacesignatures} \emph{1divideSpace\_noSize.sql} and \emph{1divideSpace\_size.sql}.}. 
Assume the location and attributes of a set of landmarks is givens for a study area, and the visibility range of each landmark has been set up as discussed in Section~\ref{sec:landmarks}, the \emph{visibility areas} occluded by buildings are first removed based on the line-of-sight of each landmark (Figure~\ref{fig:mv_visibility}) using Algorithm~\ref{alg:euclid} (Appendix); then, the intersections of all \emph{visibility areas} are calculated to identify co-visible landmarks. After that, 
\begin{enumerate}
    \item if landmarks are considered as \emph{points}, each of the intersection areas is divided successively using the lines connecting each pair of co-visible landmarks, the two corresponding perpendicular lines of each connecting line, and the circles centred at the middle point of each connecting line segment.
    \item if the \emph{extent} of landmarks is considered, landmarks $A, B$ on the X-Y plane are represented by their convex hulls and each of the intersection areas is divided using the extended lines of the upper-upper, lower-lower, upper-lower and lower-upper tangents of each pair of co-visible landmarks, the two perpendicular lines of line $ab$ (where \emph{a, b} are the centroids of the two polygons) passing the two outer intersections $a_1b_1$, and circles centred at the middle points of line segment $a_2b_2$ (where $a_2b_2$ are the inner intersections of line $ab$ with the two polygons).
\end{enumerate}
In each of the resulting areas, the same ordering, relative orientation, and qualitative angles can be observed between landmark pairs, so a random point is chosen from each such area\footnote{using postGIS function \emph{ST\_PointOnSurface(geometry g1).}} to compute the corresponding place signatures. Note when the size of landmarks is considered, the two tangents from the viewer to each landmark are used to identify the ordering relations between visible landmarks.
Finally, unique place signatures are identified and place cells sharing the same place signatures are indexed for ease of information retrieval in the next stage.

Additionally, the unique place signatures that are rotations of each other are considered as cyclically equal, and stored using linked tables for future location retrieval. 
To check whether a sequence $s1$ is a rotation of $s2$, we first check whether they are of the same length; if so, we concatenate one sequence with itself then check whether it contains the other sequence. If so, they must be a rotation of each other. 
Note that all components of two place signatures need to be cyclically equal with corresponding shifts of the starting elements. 
One such example is shown in Figure~\ref{fig:rotation}.

\begin{figure}[htp] 
    \centering  
    \includegraphics[width = \linewidth]{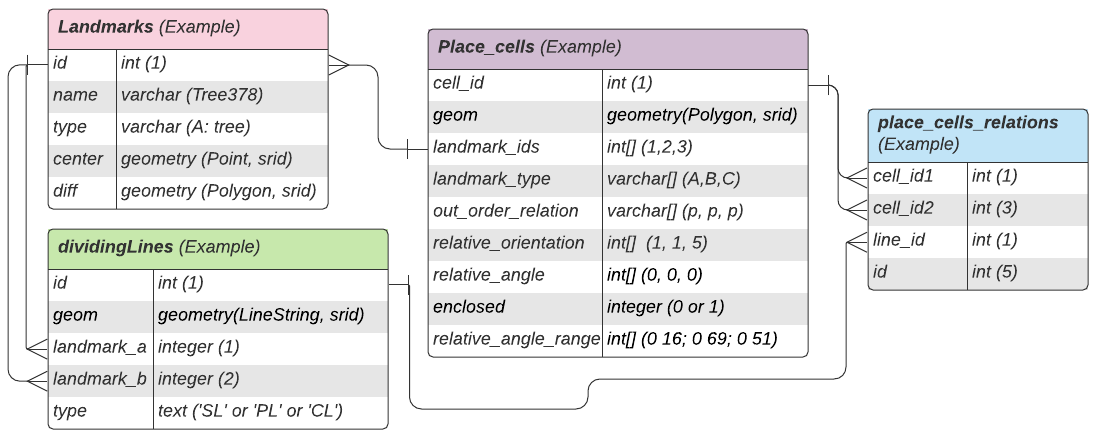}
    \caption{The database diagram of landmarks, place cells and place signatures. Note the relative relations 
    are stored between landmark 1-2, 1-3, and 2-3 sequentially.}
    \label{fig:database}    
 \end{figure}

\noindent \textbf{Reference database management}. The created place cells and signatures can be managed in a relational database, as shown in Figure~\ref{fig:database}, using a table \emph{Landmarks} storing the information of individual landmarks, a table \emph{Place\_cells} storing the information of place cells, a table \emph{dividingLines} storing all lines used to divide the space with their type \{\emph{SL, PL, CL}\} attached, and a table \emph{place\_cells\_relations} storing the adjacency relations between cells with link to the associated dividing line.  
 
Based on the \emph{type} of a dividing line, we can infer that:
\begin{enumerate}
    \item if viewers walk across a \textbf{\emph{Straight Line (SL)}} connecting two visible landmarks $A, B$, the observed orders of the two landmarks will be reversed in adjacent areas; 
    \item if viewers walk across one of the two \textbf{\emph{Perpendicular Lines (PL)}} of the \emph{Straight Line} connecting $A, B$, the observed \emph{relative orientations} between the two landmarks will change between 1 and 3, or 3 and 5;
    \item if viewers walk across a
    \textbf{\emph{Circular Line (CL)}}, the observed \emph{qualitative angles} between the two corresponding landmarks will change between acute and obtuse;
    \item if viewers walk across the visibility \textbf{\emph{Boundary Line (BL)}} of a landmark, the landmark will be removed or added into the visible landmark list.
\end{enumerate}

\noindent \textbf{Reference database updating to add new landmarks}.
When a new landmark is added into the study area, only those place cells within or with intersections to the \emph{visibility area} of the new landmark need to be examined. These areas will be further divided using the three types of dividing lines of the corresponding visible landmarks, and the place signatures of the updated place cells will be calculated accordingly.

\noindent \textbf{Impact of the uncertainty of reference landmark locations}. \textcolor{black}{Due to the inaccuracy of GPS devices used in data capturing, errors in map digitising, remote sensing surveys, etc, it is not uncommon to see uncertainty in GIS maps~\citep{LongleyPaul2005Gis, gis_uncertainty}, . In this work, we assume the semantic information of landmarks are well defined in the maps, only the uncertainty in landmark locations is discussed.}

Assume the degree of uncertainty in landmarks location is much smaller than their visibility range, this uncertainty will primarily affect how the spatial relations discussed above \textcolor{black}{will be} observed by viewers rather than which landmarks will be observed. 
By modelling the location uncertainty of each landmark using independent multivariate Gaussian distributions on \emph{X (Easting)} and \emph{Y (Northing)} directions, and assuming the errors on both directions are uncorrelated, the uncertainty of any two co-visible landmarks $A(x_1, y_1)$ and $B(x_2, y_2)$ can be propagated to the three types of space dividing lines \emph{SL, PL, CL} using first-order Taylor series propagation, as detailed in Appendix~\ref{sec:location_uncertainty}. 
\textcolor{black}{As each type of dividing line corresponds to a certain type of qualitative spatial relation,} we may expect viewers to observe a relation different to those stored in the database with a different likelihood which is depending on the viewers' location w.r.t. individual dividing line.
Using this procedure, the likelihood of viewers observing an inconsistent spatial relation between any two co-visible landmarks in a reference place cell can be pre-defined.
 
\noindent \textbf{A simplified version of place signatures using relations between successively observed landmarks}.
Assume \textcolor{black}{there are} \emph{n} co-visible landmarks to a viewer, there would be $\frac{n(n-1)}{2}$ sets of relative relations between all ordered pairs of landmarks. For example, a signature with 35 visible landmarks will have $595$ sets of relative relations. Though providing this full set of information would make a place signature more unique ( as discussed in Section~\ref{sec:ioc_order}), it may be a great burden for humans to identify all such relations. \textcolor{black}{Depending on the applications,} it might be \textcolor{black}{worth just providing} the $(N-1)$ groups of relations between successively observed landmarks. Note that with a database of complete place signatures generated (as shown in Figure~\ref{fig:database}), this step is equivalent to extracting the $k^{th}$ elements from the relation vectors, where 
\begin{equation}
k= 1-\frac{i(i-1)}{2} + (i-1)N, \; i = 1, \cdots, N-1; N\geq2.
\end{equation}
This simplified version will be used in the following location retrieval method.

\section{A coarse-to-fine Location Retrieval Method Using Visible Landmarks based Place Signatures}
\label{sec:locationRetrieval}

\textcolor{black}{To identify a viewer's location, they first need to report their observations} by
starting from the left-most landmark in their FOV\footnote{In more detail, if all landmarks can be placed in the FOV, viewers can start from the left-most landmark; otherwise, start from the landmarks nearest to the direction of true north if this is known; otherwise start from any landmark.}, and continuously providing the types of following landmarks as well as their relative orientations and qualitative angles by turning clockwise until returning to the starting landmark. This step could potentially be automated if a camera(s) is used to capture the scene\footnote{Note that though the information of ordering, occlusion as well as relative angles between 2D landmarks can be extracted from a single (panoramic) image, extracting the relative orientations between consecutive landmarks may need 3D cues.}. In this work, we assume such a place signature $ps_j$ is ready for use and the reference database is complete\footnote{or at least we assume that landmarks in urban environment are modified gradually and the reference database is reviewed regularly such that the observed place signatures will be similar to those stored ones.}, the location retrieval task turns to finding those place cell(s) in the database with the most similar signature $PS_i$ to the observed one, which is equivalent to finding those $PS_i$ with the smallest distance to the queried place signature, written as:
\textcolor{black}{
 \begin{equation}
PS_i = \{\alpha: \alpha \in Database \wedge (\forall \beta \in Database: dist(\beta, \gamma) \geq dist(\alpha, \gamma))\}
 \end{equation} 
 where $\gamma$ is the observed signature, and $dist(,)$ measures the similarity of two signatures (see Section~\ref{sec:edit_distance} and Section~\ref{sec:proposed_method} below).}
The ideal situation is that viewers can observe all surrounding landmarks and their relations correctly so an exact match can be found in the database, but this is often not the case due to inaccurate perception and possible uncertainty in landmark locations. For example, a landmark could be occluded by cars or identified as a wrong type by viewers, or certain non-existent landmarks could be reported by mistake. 

\textcolor{black}{
In the following sections, we will first discuss the pros and cons of a basic distance metric for real-time place signature matching (Section~\ref{sec:edit_distance}), then introduce
a coarse-to-fine location retrieval method based on signature indexing (Section~\ref{sec:proposed_method}).}

\subsection{The pros and cons of a basic distance metric: Edit Distance}
\label{sec:edit_distance}

As each place signature is composed of three ordered sequences of landmark types, relative orientations and qualitative angles, the effect caused by perception errors are exactly the same as deleting, inserting or substituting characters in strings. \emph{{Edit Distance} or {Levenshtein distance}}~\citep{Navarro_aguided}, a commonly used distance metric for string matching, would be a good candidate metric for comparing place signatures.

\begin{figure}[htp] 
    \centering  
    \subfigure[A place signature  \emph{$s_1$ = $\langle ABC, ppp, 115, 000 \rangle$}]{\label{fig:test_signature}
      \includegraphics[width = 0.45\linewidth]{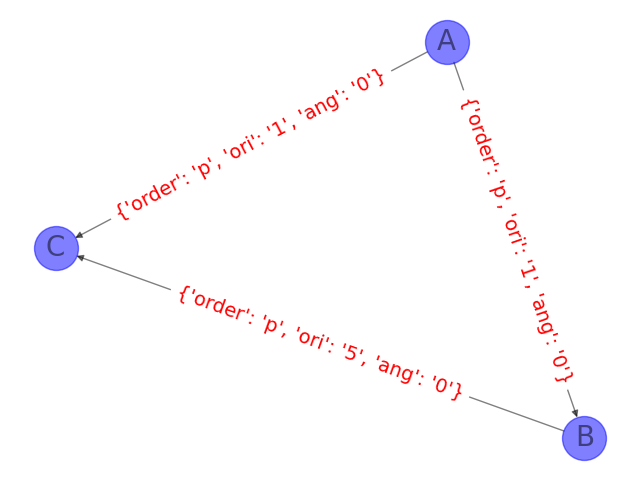} }
  \subfigure[Distance from reference cells to the queried one. ]{
  \label{fig:result_test_signature}
      \includegraphics[width = 0.45\linewidth]{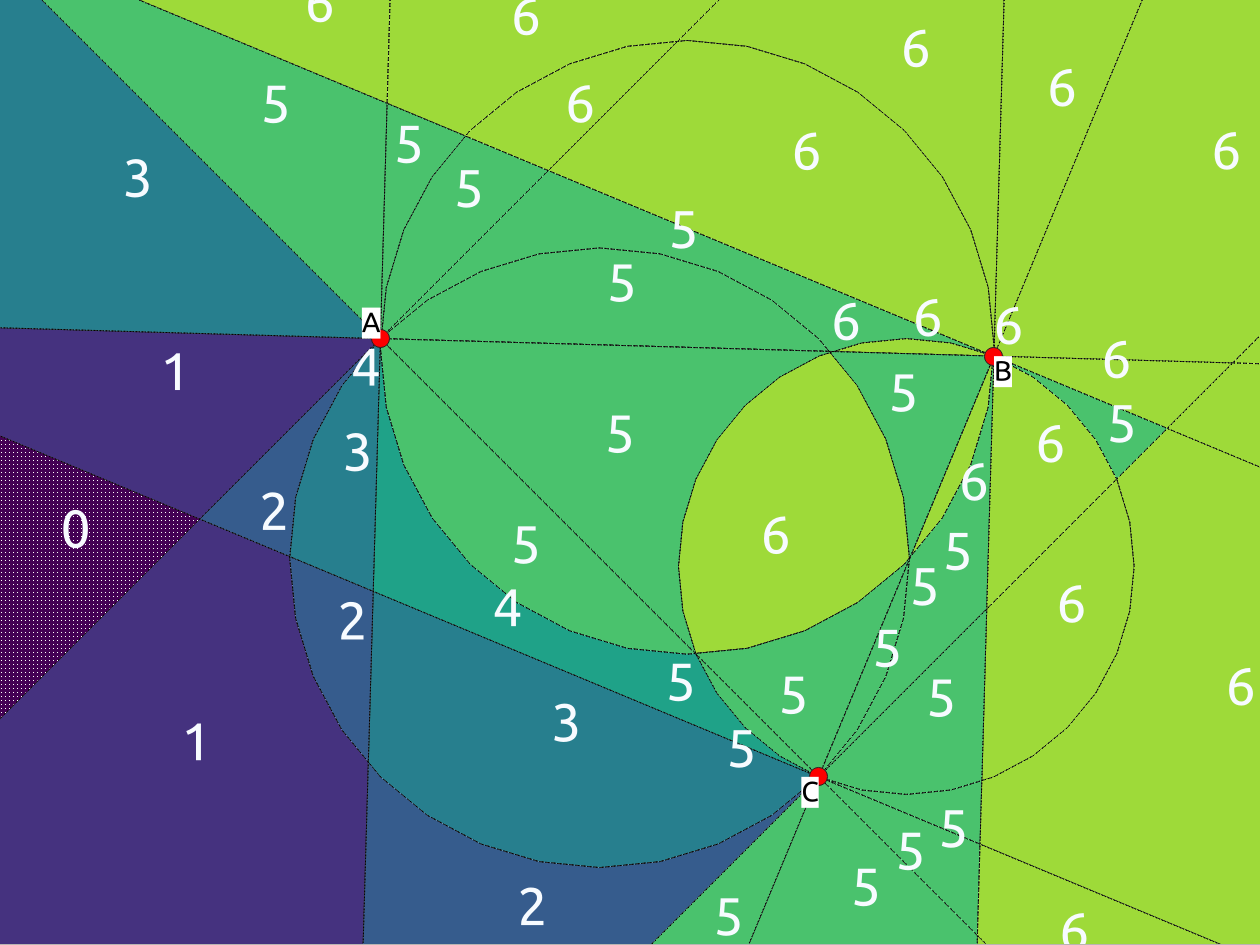}}
    \caption{An example of using \emph{edit distance} to compare place signatures. It can be seen that the corresponding place cell of the signature shown in Figure (a) is correctly identified as the one with the smallest distance (dist = 0).}
    \label{fig:example_dist}
 \end{figure}

It measures the similarity of two strings by counting the minimum number of deletion, insertion, substitution or transposition~\citep{Damerau1964} of characters required to transform the source string into the target one. 
For example, given an observed (and complete) place signature with three landmarks shown in Figure~\ref{fig:test_signature}, its distances\footnote{When the complete set of relations between all ordered pairs of landmarks need to be considered, \emph{graph edit distance}~\citep{Abu-Aisheh2015} can be used by representing landmarks as nodes  with attached attributes (e.g. type) and representing ordering relations between landmarks as directional edges  with other types of spatial relations attached. Then, finding the \emph{edit distance} between two graphs is to find the minimum number of required edit operations on nodes and edges to transform one graph into another.} to all reference signatures are shown on the corresponding place cells in Figure~\ref{fig:result_test_signature}. It can be seen that the queried place cell is correctly identified with a distance $0$.

However, {it is time-consuming to employ \emph{edit distance} for real-time place signature matching, especially for large-size databases with long place signatures}. As mentioned earlier, a simplified place signature with $n$ landmarks 
has an associated
sequence of $n$ landmark types, and two sequences of at least ($n-1$) spatial relations between ordered landmark pairs.
Given two place signatures with $n_1$ and $n_2$ landmarks, it will take quadratic time $O(3n_1n_2)$ to compare the two signatures\footnote{In addition to comparing the relations between consecutive landmarks, it will take $O(2n_1(n_1-1)n_2(n_2-1)/2^2) ~= O(2(n_1n_2)^2)$ to compare the relations between all landmark pairs.}.
If there are $\mathbf{P}$ distinct place signatures in the reference database and the maximum number of landmarks in a place signature is $\Bar{n}_2$, the total comparing time would be $O(3n_1\Bar{n}_2*\mathbf{P})$, which is linear to the size of the reference database, the length of the queried place signature, and the maximum length of reference place signatures.

For example, given a randomly selected place signature with 17 landmarks $\langle FFGGGGFGBGGGGGGJI \rangle$ from the \emph{Leeds dataset} (will be detailed in Section~\ref{sec:leed_data}), it took $0.04$ seconds to calculate its \emph{edit distance} to another randomly selected ignature with three landmarks, \textcolor{black}{running} on a laptop of \emph{Intel® Core™ i7-7500U CPU @ 2.70GHz} with one processor\footnote{Using \emph{MatLAB R2021a}.}. If we assume all reference place signatures contain an average of $35$ landmarks and $\mathbf{P}=1,178,445$
(i.e. numbers are from the \emph{Leeds dataset}), it would take approximately $0.04*35/3*1,178,445$ seconds (around $152.76$ hours) on the same machine to search through the whole database. In fact, it even took \emph{316 seconds} to finish this procedure using parallel processing on a High Performance Computing facility with 12-cores\footnote{ARC4, University of Leeds, link: \url{http://www.arc.leeds.ac.uk}, last accessed: 2022-07-21.}.
This time complexity makes \emph{edit distance} impractical for real-time place signature matching.
Therefore, it is important for us to investigate more efficient methods to quickly reduce the number of candidates such that more time-expensive yet more accurate methods could be used. 

In the following sections, a coarse-to-fine location retrieval method is proposed by gradually reducing the number of candidates using \emph{weighted MinHash}, \emph{Jaccard distance of bags}, and \emph{Edit distance} by considering the uncertainty in landmarks perception. 


\subsection{The proposed coarse-to-fine location retrieval method}
\label{sec:proposed_method}

\noindent \textbf{Preparation step: representing place signatures as vectors of numbers using k-mers}. 
To facilitate the use of other distance measures, the original place signatures are first mapped to vectors of numbers using K-shingles (k-grams)~\citep{mining2014} (or k-mers in Bioinformatics~\citep{arbitman:hal-03219482}), which are substrings of length $k$ contained in a document or a sequence. Each component of a place signature is represented as a vector of k-mers by selecting a certain \emph{k} (or a combination of different \emph{k}s).
For example, given ten types of landmarks represented by \emph{\{A, B, C, D, E, F, G, H, I, J\}}, there could be up to $10^1$ distinct 1-mers and $10^2$ 2-mers in a sequence of visible landmark types. The vector of 1-mer \emph{term counts (tc)} in a sequence $\langle AFFJBAAAGBFF \rangle$ would be $\langle 4200041001 \rangle$, as shown in Table~\ref {tab:representation_example}.
\begin{table}[htb!]
\sffamily
\begin{center} 
\resizebox{0.9\textwidth}{!}{
\begin{tabu}{ lccccccccccccccccc  }
k=1&A&B&C&D&E&F&G&H&I&J&\multicolumn{7}{r}{sum=12, $f_{k=1}=\frac{12}{10^1}$}\\  \rowfont{\color{blue}}
tc:&4&2&0&0&0&4&1&0&0&1&\\
\rowfont{\color{brown}}
tf&0.33&0.17&0&0&0&0.33&0.08&0&0&0.08&\\
\rowfont{\color{cyan}}
ta&1&1&0&0&0&1&1&0&0&1&\\ 
\rowfont{\color{magenta}}
btc:&1&1&0&0&0&1&0&0&0&0&\\
\hline
k=2&AA&...&AF&AG&...&BA&...&BF&...&FF&...&FJ&GB&...&JB&...&sum=11, $f_{k=2}=\frac{11}{10^2}$\\
\rowfont{\color{blue}}
tc&2&0&1&1&0&1&0&1&0&2&0&1&1&0&1&0\\
\rowfont{\color{brown}}
tf&0.18&0&0.09&0.09&0&0.09&0&0.09&0&0.18&0&0.09&0.09&0&0.09&0\\
\rowfont{\color{cyan}}
ta&1&0&1&1&0&1&0&1&0&1&0&1&1&0&1&0\\
\rowfont{\color{magenta}}
btc&1&0&1&1&0&1&0&1&0&1&0&1&1&0&1&0
\end{tabu}} 
\end{center}
\caption{An example of representing a sequence of landmark types $\langle AFFJBAAAGBFF \rangle$ as vectors of numbers.}
\label{tab:representation_example}
\end{table}

\noindent Other vector representations are also available (Table~\ref{tab:representation_example}), such as the \emph{frequency of terms (tf)} by dividing \emph{tc}s with the total counts of terms appeared in a sequence, the \emph{appearance of terms (ta)} by counting whether each term appears (1) or not (0) in a sequence, or the \emph{binarized term counts (btc)}~\citep{arbitman:hal-03219482}
by calculating the average count of all distinct k-mers in a sequence and keeping those elements with a count below the average as 0 and others as 1, etc.
\textcolor{black}{The representation using \emph{ta} is selected in this work as it is shown in the experiments (Section~\ref{sec:experiments}) that it provides the best location retrieval performance compared to other representations.}

Similarly, the sequence of relative orientations can be converted to vectors of numbers using $K_1=\Sigma_{k\in[k_1,k_2]}3^k$ terms of {\sffamily{\{1~2~3;~11~12~13~21~22~23~31~32~33;...\}}}\footnote{Note there are eight relative orientation indices in total, four on lines (2, 4, 6, 8), one on points (7), and the remaining three (1, 3, 5) for regions which are used for representation.}; and the sequence of qualitative angles can be represented using $K_2=\Sigma_{k\in[k_1,k_2]}2^k$ terms of {\sffamily{\{0~1;~00~01~10~11;...\}}}, \textcolor{black}{where $[k_1,k_2]$ set a range of values for $k$ between $k_1$ and $k_1$}.
Then, we can either concatenate the three vectors or use them sequentially for distance matching .
Though using them sequentially can reduce instantaneous memory requirements, 
there is a potential that true positives could be filtered out in an earlier stage because it is unknown which part of the signature could be incorrectly observed. Therefore, the three vectors are concatenated as one vector for the following analysis.  

After mapping each place signature into a vector of a fixed-length $K=K_1+K_2+K_3$, alternative distance metrics 
include \emph{cosine distance}~\citep{doi:10.1080/13658816.2013.790548, shahmirzadi2018text, semanticclusters2016}, \emph{Hamming distance}~\citep{arbitman:hal-03219482}, \emph{Jaccard distance}~\citep{mining2014}, etc. 
The time complexity of calculating these distances between two vectors is $O(P*K)$\footnote{More explicitly, \textcolor{black}{\emph{O}(Edit distance) $\geq$} \emph{O}(Jaccard bags) $\geq$ \emph{O}(Jaccard Distance) $\geq$ \emph{O}(Hamming distance) $\geq$ \emph{O}(Cosine distance)). Detailed comparison will be given in Section~\ref{sec:time-theory}.}, which is better than \emph{Edit distance} but still linear to the size of distinct vectors.
As observations are subject to errors and an exact match may not exist in the database, it would be useful if `similar' reference place signatures are placed together so we will only to examine the distances to \textcolor{black}{these similar ones}.
\vspace{0.2cm}

\noindent \textbf{Step 1: initial fast screening using locality sensitive Hashing (\emph{LSH})}. Being different from other Hashing methods to avoid hashing collision, \emph{LSH} (or \emph{approximate Hashing}) methods hash vectors such that similar ones are more likely to be hashed to same buckets and dissimilar ones into different ones~\citep{arbitman:hal-03219482,mining2014, 10.1093/bioinformatics/btz354}. 
One most used \emph{LSH} method is \emph{MinHash} which efficiently approximates Jaccard distance~\citep{mining2014} by randomly generating $n$ hash functions to simulate the random permutations of term \emph{ids}, then mapping each input vector to a vector of $n$ minimum hash values. 
This method can only be applied to unweighted vectors with binary values, such as vectors of term appearance (ta) (an example was given above in Table~\ref{tab:representation_example}). 

To take account of the exact number of terms appeared in each place signature, \emph{Weighted MinHash}~\citep{Ioffe2010,NIPS2016_c2626d85} is used in this work to map vectors of term counts (tc) to vectors of $n_{hash}$ hash values such that the probability of drawing identical samples for a pair of inputs is identical to their Jaccard similarity. The algorithm is summarised in Algorithm~\ref{alg:minhash}. 

\begin{algorithm}
\caption{
Weighted MinHash~\citep{Ioffe2010}.}
\label{alg:minhash}
\begin{multicols}{2}
\begin{algorithmic}[1]
 \Procedure{w\_MinHash}{$x$, $n_{hash}$} 
 \State{\% generate random hash variables}
     \State \texttt{K := length(x)}
     \For{\texttt{i = 1 to $n_{hash}$}}
     \State Sample \texttt{$r_{i}$, $c_{i}\sim$ Gamma(2,1)}
     \State Sample \texttt{$\beta_{ij}\sim$ Uniform(0,1)}
     \EndFor
    \State{\% calculate Hash values for each vector $x$} 
    \For{\texttt{i = 1 to $n_{hash}$}}
    \For{iterate over $x_j$ s.t. $x_j>0$}
    \State \texttt{$t_j = floor(\frac{\log x_j}{r_{ij}}+\beta_{ij})$} 
    \State \texttt{$y_j=\exp({r_{ij}(t_j-\beta_{ij})})$}
    \State \texttt{$z_j=y_j*\exp({r_{ij}})$}
    \State \texttt{$a_j = c_{ij}/z$}    \EndFor 
    \State \texttt{$h^* = \min_j a_j $}
    \State \texttt{hashPairs[i] = ($h^*,t_{h^*}$)}
    \EndFor
    \State \textbf{return} \texttt{hashPairs}
 \EndProcedure
\end{algorithmic}
\end{multicols}
\end{algorithm} 
Using this method, the hash vectors of all reference place signatures are pre-computed offline and place signatures with the same hash vectors are placed in a same group, resulting in $P_l$ unique reference hash vectors, where $P_l\leq \mathbf{P}$. 
When an observation is provided by viewers, it is first hashed to a vector of $n_{hash}$ values using the same set of hash functions, then compared with all reference hash vectors by counting the number of positions with different hash values, noted as $n_0$. By setting a threshold \textcolor{black}{to the proportion of different positions} $t=\frac{n_0}{n_{hash}}, 0\leq t \leq 1$, those reference vectors with a proportion less than the threshold are considered as `similar' candidates. 

The comparison process requires $O(d)$ time to compute the hash values of a vector of $d$ non-zero values, and $O(P_l*n_{hash})$ time to compare with all reference hash vectors, which is linear to the number of hash functions $n_{hash}$ and the number of buckets $P_l$ (i.e. the number of unique vectors of hash values). 
Then, candidates in the `similar' buckets can be further examined in the following steps.

\textcolor{black}{
Note that since \emph{Weighted MinHash} is an approximation of \emph{Jaccard distance} (which requires exhaustive searching through the whole database), detailed comparison of their retrieval performance in terms of query time and precision-recall will be given in the experiment section~\ref{sec:evaluation_lsh}\footnote{\textcolor{black}{The recall and precision rate of exhaustive searching using \emph{Edit distance} is
not directly compared due to its impractical
time complexity as discussed in Section~\ref{sec:edit_distance}.}}.
Overall, \emph{LSH} can significantly reduce the number of candidates in a fraction of the time compared to using \emph{Jaccard distance}, and feeding this reduced set of candidates into the next stage with an exact matching metric will only take the corresponding proportion of the exhaustive searching time while maintaining the same precision rate. \textcolor{black}{For example, the \emph{LSH} approach can reduce the average number of candidates per query to half the size of the database at a recall rate $1$ using nearly 10\% of the searching time of using \emph{Jaccard distance}. This suggests that by using at most $1/10+\mathbf{1/2}=3/5$ of the exhaustive searching time, the same precision rate can be achieved with \emph{LSH}. When a slightly smaller recall rate is considered, e.g. $0.97$, the number of candidates can be reduced to $1/30$ of the database size, which totalled in $1/10+\mathbf{1/30} = 2/15$ of the exhaustive search time}}.

\vspace{0.2cm}
\noindent \textbf{Step 2: Candidates refinement using \emph{Jaccard distance of bags} with an adaptive distance threshold}.
The \emph{Jaccard distance} of two sets $s_1$ and $s_2$ is one minus the Jaccard similarity of the two sets, defined as the ratio of the size of their intersection and the size of their union~\citep{mining2014}, written as: 
\begin{equation}
J(s_1,s_2) = 1 -\frac{|S_1\cap S_2|}{|S_1\cup S_2|}=1-\frac{\sum min(a_1,a_2)}{\sum (a_1+a_2)}
\end{equation} 
When the number of terms appeared in each sequence is considered, the above Jaccard distance turns to \emph{Jaccard distance of bags} (Jaccard bags): the intersection of two sets is the sum of the minimum number of each term appeared in the two sequences, and their union is the total number of terms~\citep{mining2014}. For example,
the 1-mer tc vectors of two sequences \emph{`ACBC'} and \emph{`ADCA'} are respectively \ $a_1=[1~1~2~0]$ and $a_2=[2~0~1~1]$, and their Jaccard bags is $Jb(s_1,s_2) = 1 -\frac{\sum[1~0~1~0]}{\sum [3~1~3~1]}= 6/8$.

After calculating the Jaccard distance to all candidates found in the \emph{weighted MinHash} step, the \emph{k-nearest candidates} or those with a distance less than a threshold can be kept as the refined candidates. 
Although setting up a high distance threshold would allow severely deformed place signatures being corrected retrieved (recalled), it will inevitably bring in more false positives for those queries with smaller perception errors. Therefore, an adaptive threshold is chosen for each queried place signature by taking a fixed threshold $t_1$ (e.g. $t_1 = 0.59$), or the $l^{th}$ (e.g. $l = 50$) lowest distance $m(l)$ between the queried and reference candidates, whichever is smaller, $t_i = min(t_1, m(l))$.
    
Through above procedures, the number of candidates can be quickly reduced to an acceptable level, enabling the usage of \emph{Edit distance} to examine whether the number of candidates could be further reduced while retaining the same recall rate.
\vspace{0.3cm}

\noindent \textbf{Step 3: Further candidates refinement using \emph{Edit Distance} by considering the uncertainty in landmarks perception}. As each place signature is composed of three sequences, the \emph{edit distance} between two place signatures $ps_1, ps_2$ is the weighted sum of the distances of the three components:
\begin{equation}
d(ps_1, ps_2) = w_1*d(type_1, type_2) + w_2*d(ro_1, ro_2) + w_3*d(ra_1, ra_2)
\end{equation}
\textcolor{black}{All weighting factors are set equally as $1/3$ in this work.}
The costs of \emph{deletion, insertion} and \emph{substitution} of elements were defined by considering the likelihoods of errors in landmarks perception. For example, we could imagine that it is often more likely for us to miss an existing landmark (e.g. due to occlusion) than to `discover' a non-existent one\footnote{Assuming the reference database is complete.}. Therefore, if a viewer reports a sequence $\langle ABC \rangle$ and there are two reference signatures $\langle ABCD \rangle$ and $\langle AB \rangle$, \textcolor{black}{although the edit changes to them are both one,} it would be more likely for the viewer to be in a place where $\langle ABCD \rangle$ should be observed by missing a `D' than in a location where $\langle AB \rangle$ should be observed by adding a `C'. This suggests that, generally, the edit cost of \emph{deleting} a landmark should be lower than the cost of \emph{inserting} a landmark. Similarly, we can imagine that once we observed a landmark, it would be very unlikely for us to misidentify its general type (given a limited number of options), at least more unlikely than missing the landmark. These observations can help us set up a global constraint on the edit cost of \emph{deleting, inserting} and \emph{substituting} landmarks in this work:
    \begin{equation}
    Constraint\;1:\; C_{subs}\geq C_{ins} \gg C_{del} 
    \end{equation}
In addition to the global constraint, for each of the above three types of perception errors/edit changes, the cost can vary for landmarks of different attributes. For example, imagine a situation that a viewer reports a sequence \emph{ACD}, two candidate signatures \emph{`A{\textcolor{black}B}CD'} (`bin' stands for bin) and \emph{`A{\textcolor{black}J}CD'} (`J' stands for tree) would be ranked equally with one edit change of \emph{deletion} for both. But could \emph{`A{\textcolor{black}B}CD'} be slightly more likely to be the correct match given that \emph{bins} are usually much shorter than trees thus more likely to be occluded by other objects? 

In general, the bigger a landmark is on viewers' retina or the image plane of cameras (which means taller, wider or closer), and more salient it is (depending on factors like colour, pattern, static/flashing etc.), the less likely it will be missed or being identified as a wrong type. Therefore, the higher the weight of change cost of a node/landmark will be. This relation could be expressed as: $
        w^n_1\propto \{height, \; size, \;visual\;salience\}
$. 
For example, the above example of \emph{bin} and \emph{trees} deletion can be expressed as: $w_1^{bin}C_{del}<w_1^{tree}C_{del}$. A more detailed discussion on the uncertainty in landmarks and place signature perception is given in Appendix~\ref{sec:perception_error}. 
After calculating the \emph{edit distance} between a query place signature and all candidates, the final candidates are identified by choosing those with \emph{k-smallest \emph{distance}s} to the queried one, or those with a distance below a threshold $t_e = b*len(ps_q)$, where $b$ is a proportion value (e.g. $ b=[0~1/6~1/6~3/6~4/6~5/6~1]$) and $len(ps_q)$ is the length of the queried place signature.

\vspace{0.2cm}   
\noindent \textbf{Assigning \textcolor{black}{a} probability to retrieved location hypotheses}. After finding the final candidate(s) for each queried place signature, we can represent our estimate of the viewer's initial location with a probability density function distributed over the corresponding place cells of those candidate reference place signatures using a set of M particles $loc_{t}=(loc_{t}^{1},\cdots,loc_{t}^{M})$.
Each particle contains the information of a candidate place cell and is considered as a hypothesis of the viewer's location. A uniform probability can be assigned to the viewer's possible locations inside each place cell w.r.t the area of the place cell, as $P(S_i)=1/{|X_k|}$
where $X_k$ is the total volume/area of a place cell. This information can be used to further refine the viewer's location or trajectory as they start moving and continue providing their observations.

\section{Experiments and Evaluation}
\label{sec:experiments}

To evaluate the proposed location retrieval method, a reference database of place cells and place signature was created for the city centre of \emph{Leeds} in the UK following the methods presented in Section~\ref{sec:signatureDefinition} \textcolor{black}{and~\ref{sec:locationRetrieval}}; \textcolor{black}{then, a set of place signatures were randomly selected and modified} to simulate observations and used to evaluate the proposed location retrieval method
by examining the query precision, recall and time complexity.

\begin{figure}[htb!] 
    \centering  
    \includegraphics[width = 0.8\linewidth]{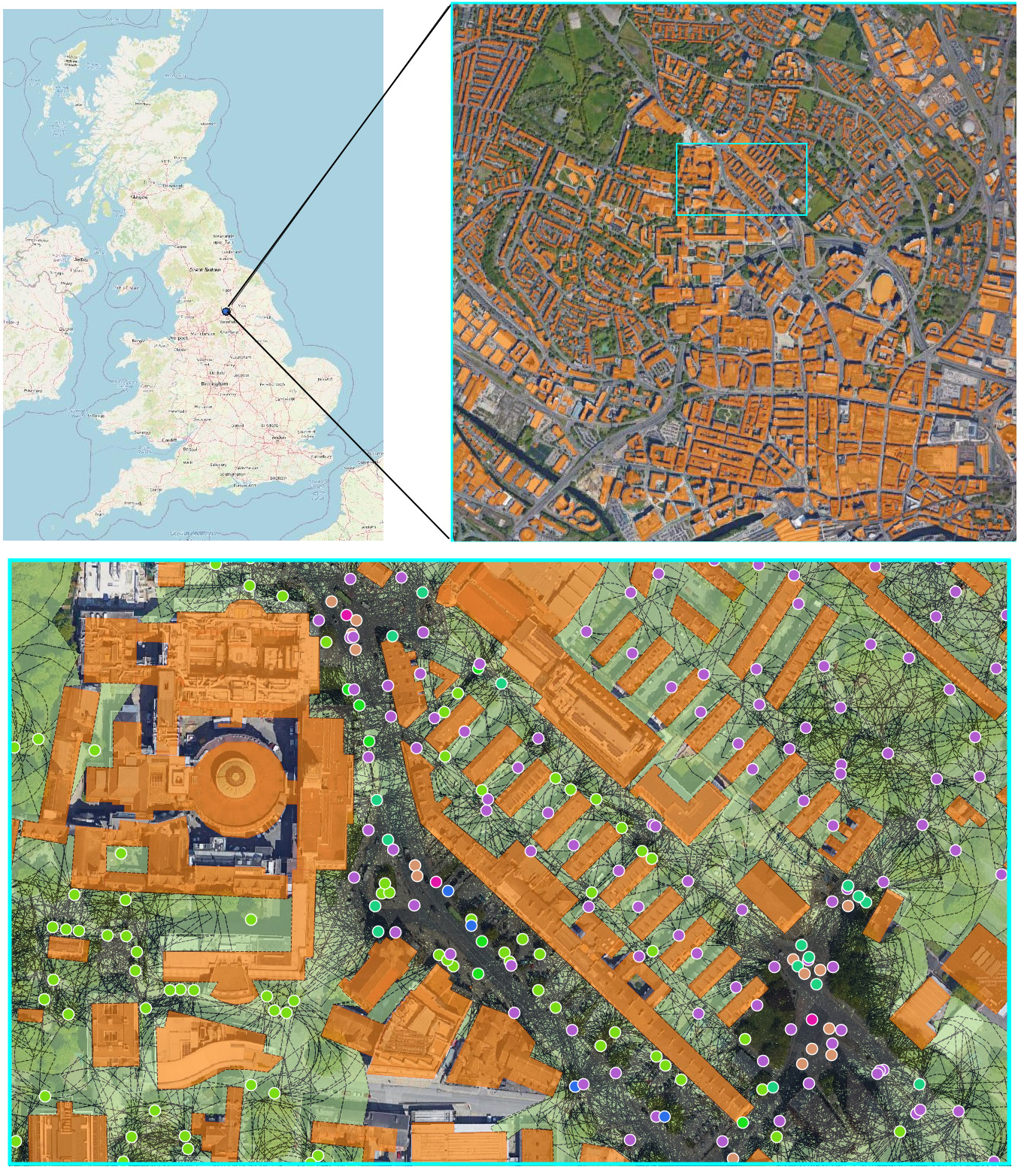}
    \caption{\textcolor{black}{Overview} of the \emph{Leeds Dataset}. Top left) the location of the Leeds dataset in the UK. \textcolor{black}{Top right) the studied area in Leeds city centre. A detailed view of the landmarks/place cells in the the area outlined in cyan is given in the bottom figure. Bottom) A zoomed view of the exemplar area divided by visibility boundary lines and lines connecting each visible landmark pair.}}
    \label{fig:leeds_data}    
\end{figure}
\subsection{Creating a reference database of place cells and place signatures}
\label{sec:leed_data}

The studied area in this work is situated in the city centre of \emph{Leeds} in the UK, as shown in Figure~\ref{fig:leeds_data}. It covers an area of $6.47km^2$, including $3.82km^2$ \emph{free} space after excluding those occupied by buildings.
As semantic and spatial discrepancy\footnote{Semantic discrepancy: different names or attributes are used to describe a same object. \textcolor{black}{For example, \emph{street lights} are stored in a table called \emph{street\_light} in \emph{Find open data}, but labelled as \emph{`highway' = `street\_lamp'} in the \emph{OSM} map.} Spatial discrepancy: the recorded locations of a same objects are (slightly) different in different datasets.} were identified in landmark datasets \textcolor{black}{sourced from the website} \emph{Find open data} and \emph{OpenStreetMap (OSM)}, and detailed alignment between them is out the scope of this work, only the data from \emph{Find open data} was used. 
The landmarks include street lights\footnote{Links to the used landmark datasets are listed in Appendix~\ref{sec:links}.}
, traffic signals
, bins
, trees
, bus stops
, 
etc. 
Although there is no limit of the \textcolor{black}{number and type} of landmarks \textcolor{black}{being} used in the proposed method, \textcolor{black}{a total of} \emph{8108} landmarks of ten types were used in this experiment, respectively represented using one character in $\langle ABCDEFGHIJ \rangle$, as summarised in Table~\ref{tab:landmarks_Leeds_probability}.

\begin{table}[htb!]
\begin{center} 
\caption{Summary of the ten types of landmarks used in the experiments.}
\label{tab:landmarks_Leeds_probability}
\begin{tabular}{ |>{\em}l| >{\color{blue}} c | >{\color{blue}}c|>{\em}l| >{\color{blue}} c | >{\color{blue}}c| } \hline 
\textbf{Landmark} &\textbf{Symbol}&Number (\%)&\textbf{Landmark} &\textbf{Symbol}&Number (\%)\\ \hline
bicycle\_parking & A & {53 (0.65)}&road sign &F &{888 (10.95)}\\  \hline
bin  &B &{348 (4.29)}&street light &G & {4,566 (56.31)}\\  \hline
bollards &C &{356 (4.39)}&toilets &H & {3 (0.037)}\\  \hline
bus stop &D &{245 (3.02)}&traffic signals &I & {113 (1.39)}\\  \hline
memorial &E &{2 (0.02)}&tree &J &{1,534 (18.92)}\\  \hline
\multicolumn{5}{|c|}{Total}&{8,108 (100)}\\ \hline \hline
\end{tabular}  
\end{center} 
\end{table}

\begin{table}[htb!]
\begin{center} 
\caption{Statistics of the place cells and place signatures in the \emph{Leeds dataset}. \emph{Symbol} stands for landmark type, \emph{ro} stands for relative orientation, and \emph{ra} stands for relative angle.}
\label{tab:landmarks_Leeds}
\begin{tabular}{ |>{\em}l|| >{\color{blue}} c | >{\color{blue}}c|>{\color{blue}}c|>{\color{blue}}c|>{\color{blue}}c||  } \hline \hline
\textbf{\makecell[c]{Attributes}} &\textbf{\makecell[c]{\emph{symbols}}}& \textbf{\makecell[c]{\emph{ro}}}&\textbf{\makecell[c]{\emph{ra}}}&\textbf{\makecell[c]{\emph{symbols}+\\\emph{ro}+\emph{ra}}}& \textbf{\makecell[c]{+\\enclosed}}\\ \hline
\makecell[c]{N. of distinct\\ place signatures} & 1,178,445  &1,916,974&9,011&2,224,059&2,232,311\\ \hline 
\makecell[c]{Max. coverage of a\\ single signature ($km^2$)} & $0.2359$&$0.2032$&$0.2099$&$0.0817$& $0.0817$\\ \hline
\makecell[c]{Avg. coverage of a\\ single signature ($m^2$)} & $3.23$
&$1.98$&$422.86$&$1.71$&$1.71$\\ \hline 
\makecell[c]{Signature with the \\largest coverage} & $\langle GGG \rangle$   &$\langle 3 \rangle$&$\langle 0 \rangle$&\makecell{$\langle GG,3,1 \rangle$}&\makecell{$\langle GG,3,1,0 \rangle$}\\ \hline
\hline
\end{tabular}  
\end{center} 
\end{table}
The `free' space was divided into 
\emph{2,224,059} 
place cells of distinct place signatures with three components, including the observed ordered sequences of landmark types (symbols), relative orientations and qualitative angles. 
Some statistics of the created dataset are given in Table~\ref{tab:landmarks_Leeds}. For example, the average number of co-visible landmarks in a place cell is $35$, and the maximum number is \emph{168} \textcolor{black}{due to close distribution of parallel street lights in certain areas}. 
As streetlight is the only type of landmarks observed in certain areas (in this dataset), the signature with the maximum coverage is $\langle GGG\rangle$ of three continuous streetlights. When relative orientation and angles are considered, the maximum coverage of a single signature is reduced to one third of the above number.

\begin{figure}[!ht] 
\centering  
\subfigure[The top-50 place signatures when the ordered sequences of \emph{landmark types} are used. Abbreviations for landmarks are: A (bicycle\_parking), B (bin), C (bollards), D (bus stop), E (memorial), F (road sign), G (street light), H (toilets), I (traffic signals), and J (tree).]{
\includegraphics[width =0.9 \linewidth]{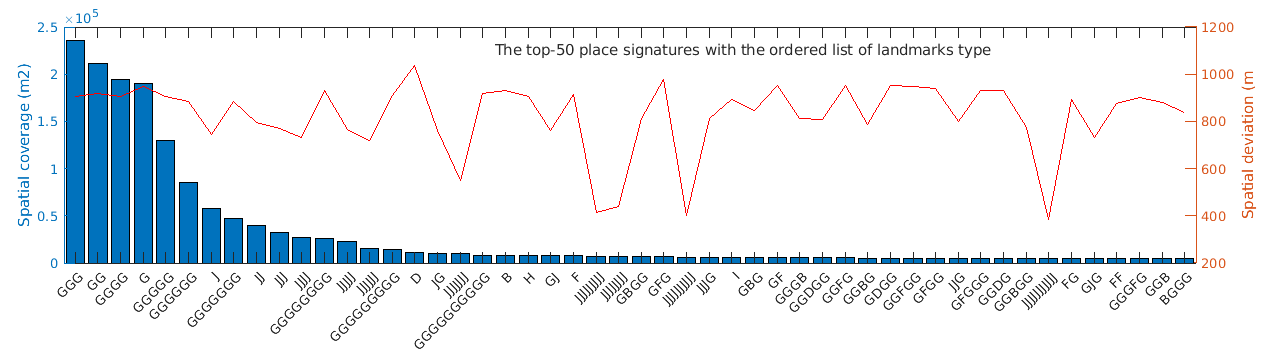}}
\subfigure[The top-50 place signatures when \emph{relative orientation}s are used. Note: when only one landmark is observed, the \emph{relation orientation} is NULL.]{
\includegraphics[width =0.9 \linewidth]{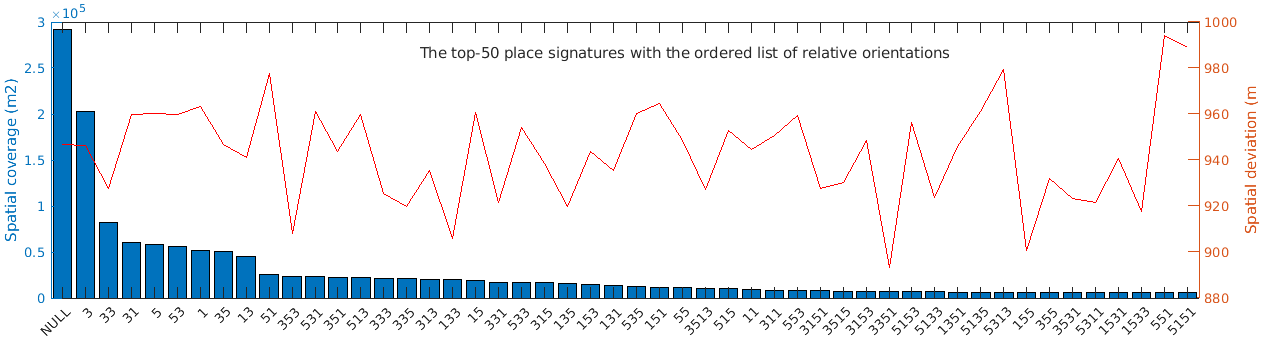}}
\subfigure[The top-50 place signatures when \emph{qualitative angle}s are used, which could be '0' (acute angle), '1' (obtuse angle) or NULL when only one landmark is observed.]{
\includegraphics[width =0.9 \linewidth]{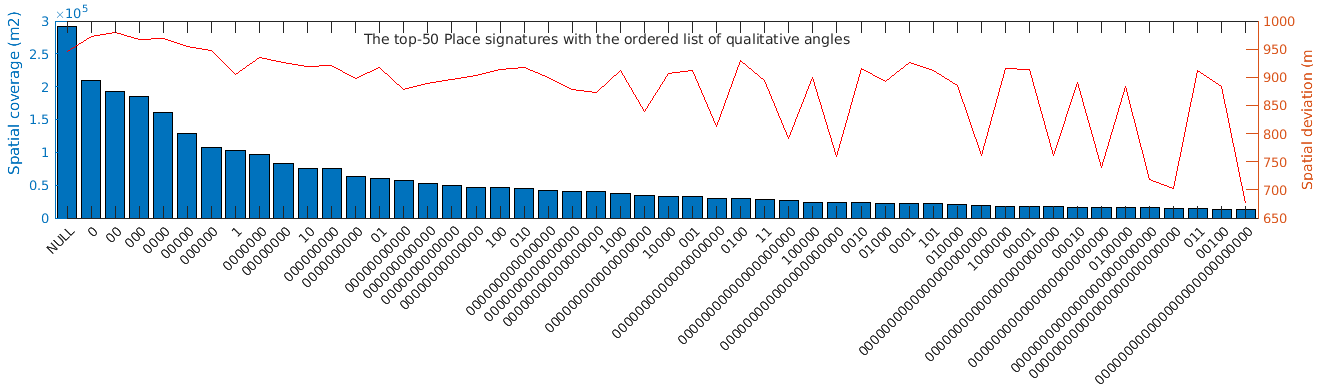}}
\subfigure[The top-50 place signatures when all three components are used.]{
\includegraphics[width =0.9 \linewidth]{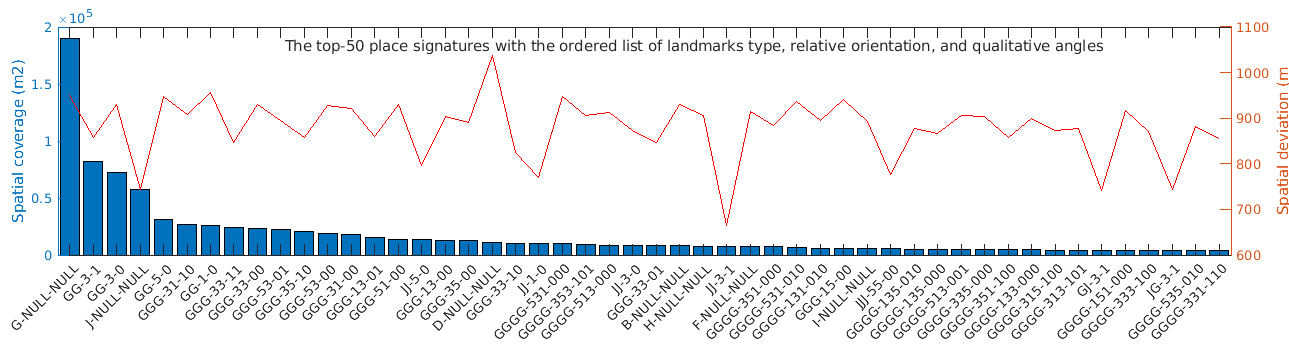}}
\caption{The spatial coverage and spatial deviation of the top-50 place signatures with the largest coverage when individual component of place signatures is used for comparison.}
\label{signatureFrequency}
\end{figure}

Furthermore, \textcolor{black}{
as the size of individual place cell might be different, the count of place cells sharing the same signature may not well describe its potential location ambiguity. To better understand the discriminating ability of each place signature, we define their} spatial coverage as the summed area of all place cells sharing this signature, noted as ${sc^j}$; and their spatial deviation
as the standard deviation of the centroids $(c_{x}, c_{y})$ of these place cells, written as $sd^j=\sqrt{var(c_{x}) + var(c_{y})}$. 
\textcolor{black}{Generally, place signatures with a small spatial coverage and deviation are relatively centralised, while those with a large deviation are Loosely distributed and viewers may observe them from many scatter locations.}
The maximum coverage of a single signature is $81,738.7m^2$ and the average is $1.71m^2$.
\textcolor{black}{The top-50 signatures with the largest coverage are displayed in Figure~\ref{signatureFrequency}}.

\textcolor{black}{In the following sections}, the overall performance of the proposed location retrieval method is evaluated step by step using randomly selected and modified place signatures from the above dataset.

\subsection{Evaluation criteria and observation simulation}
\label{sec:criteria}

\subsubsection{Evaluation criteria for location retrieval}
In this work, given the observed place signatures at $N$ locations, a list of `similar' reference signatures will be retrieved from the database. \textcolor{black}{The {\emph{precision}}, {\emph{recall}} rates and \emph{time complexity} of location retrieval methods are evaluated.} 
\textcolor{black}{A retrieved candidate is considered as a \emph{True positive (TP)} if it is the correct correspondence of the queried signature, otherwise a \emph{False Positive (FP)}. If none of the retrieved candidates contain the queried one, this query is considered as a \emph{False Negative (FN)}}. Then, the \textbf{\emph{recall}} 
and \textbf{\emph{precision}} rates are defined as below:
\begin{equation}
    Recall = \frac{\sum TP}{\sum(TP+FN)}=\frac{\sum TP}{N},~ \\
    Precision = \frac{\sum(TP)}{\sum(TP+FP)}
\end{equation}    
At a given recall rate, the method with a higher precision rate would suggest that fewer irrelevant candidates (\emph{FN}) are retrieved; while at a given precision rate, the method with a higher recall rate would suggest that more true positive candidates can be identified for the same amount of candidates. 


\begin{table}[htb!]
\begin{center} 
\caption{The probability of different types of landmarks being \emph{missed/deleted} in a trial of observation.}
\label{tab:probility_perception}
\begin{tabular}{ |>{\em}l| >{\color{blue}} c | >{\color{blue}}l|>{\em}l|>{\color{blue}}c|>{\color{blue}}c| } \hline 
\textbf{Landmark} &\textbf{Symbol}& {$p_1$}&\textbf{Landmark} &\textbf{Symbol}& {$p_1$}\\ \hline
bicycle\_parking & A & {0.2}&road sign &F&{0.1}\\  \hline
bin  &B &{0.2}&street light &G&{0.05}\\  \hline
bollards &C &{0.3}&toilets &H & {0.3}\\  \hline
bus stop &D &{0.1}&traffic signals &I &{0.05}\\  \hline
memorial &E &{0.2}&tree &J &{0.1}\\  \hline
\end{tabular}  
\end{center} 
\end{table}
\vspace{-1.0cm}

\subsubsection{Simulation of observations with errors} 
The observed signatures by viewers are simulated by first randomly selecting $N=1000$ place signatures\footnote{Any other numbers or multiple sets of place signatures could be used as well.} from the reference database, then distorted 
to mimic different types of perception errors.
To do this, we assume the probability of a landmark being deleted ($p_1$) varies between different types of landmarks, while the probability of a landmark being substituted ($p_2$) or being inserted ($p_3$) in each trial of observation are the same for all types of landmarks. For example, \emph{bollards} are assigned  a higher probability of $p_1$($0.3$) compared to \emph{traffic signals} ($0.05$) as they are often much shorter than traffic signals which are often have  flashing signals attached\footnote{Note future experiments will be needed to understand the actual probabilities in different environments/scenarios.}.
\textcolor{black}{The assigned values of $p_1$ are given in Tab.~\ref{tab:probility_perception}, and : $p_2$ and $p_3$ are set as $0.01$ in the experiments.}

\textcolor{black}{
For each type of landmark in the selected place signatures, a random number ${N_i^{sub}}$, $N_i^{del}$ and $\bar{N_i}^{ins}$ are respectively generated from a binomial distribution $binornd(N_i,p_i)$ ($i=2,1,3$) to simulate how many landmarks of this type are to be substituted, missed, and inserted, where $N_i$ is the total number of this type of landmarks in the selected place signatures\footnote{Note \emph{insertion} is simulated after substitution and deletion so the number of landmarks is recalculated as  $\bar{N_i}$.}. 
Then, $N_i^{sub}$ and $N_i^{del}$ unique random integers are generated respectively between $[1, N_i]$  to simulate which of this type of landmarks are to be substituted and deleted; and $\bar{N_i}^{ins}$ numbers are generated between $[1,\bar{N_i}]$ to suggest where landmarks are to be inserted after. After that, each of the $N_i^{sub}$ landmarks are replaced by a randomly generated different landmark type; each of the $N_i^{del}$ landmarks as well as their related spatial relations are deleted from the original signatures; and landmarks and relevant relations are inserted after each of the $\bar{N_i}^{ins}$ landmarks with randomly generated numbers between $[1,10]$, $\{1,3,5\}$ and $\{0,1\}$\footnote{Note that two of such spatial relations with respect to the previous and following landmarks need to be inserted if a landmark is inserted after the first element and before the last element of a landmark sequence; otherwise, only one element of each such spatial relations needs to be inserted.}.}

\textcolor{black}{After these steps, the $N$ modified place signatures are used in all following experiments to evaluate the location retrieval performance.}



\subsection{Evaluation of the proposed location retrieval method}
\label{sec:compare_distance}

The proposed location retrieval method is evaluated step by step by first comparing \emph{Jaccard bags} with other distance metrics, then comparing \emph{Weighted MinHash} with another approximated Hashing method, following by evaluating the proposed adaptive distance metric and the contribution of using \emph{Edit distance} by considering the uncertainty of landmarks perception. 

\subsubsection{Theoretical comparison of different distance metrics}
\label{sec:time-theory}

Given a query place signature with $n_1$ landmarks and $K$ ($\sum_k{10^k+3^k+2^k})$ terms to represent the three components of a place signature, the following distance metrics are evaluated,  
including:
\begin{enumerate}
    \item The \emph{Edit distance} between original place signatures. It will take $O(\mathbf{P}*3n_1\Bar{n}_2)$ time to search across the whole database of $\mathbf{P}$ reference place signatures, where $\Bar{n}_2$ is the maximum length of a reference place signature, where $n_1$ and $\Bar{n}_2$ are the number and average number of landmarks in the query and reference place signatures;
    \item The \emph{cosine distance} between \emph{term frequency} (tf), \emph{term counts} (tc), or \emph{tf-inverse document frequency} (tf-idf) vectors~\citep{doi:10.1080/13658816.2013.790548, shahmirzadi2018text, semanticclusters2016}. In \emph{tf-idf}, the frequencies of terms appeared in a sequence are weighted by their \emph{Inverse Document Frequency} in the corpus of sequences 
    $    idf(t)=1+log_{10}\frac{N}{n_t+1}$,  
    where $N$ is the number of reference signatures and $n_t$ is the number of them containing the term $t$. 
    Given an observed place signature, it is first represented as a vector $v_q$ of $K$ \emph{tf} or \emph{tf-idf} values\footnote{Note that the inverse document frequency $idf(t;P)$ of terms appeared in a corpus was pre-calculated and assumed to be consistent, therefore we only need to calculate the term frequency in each queried signature.} by concatenating the three components, then compared with all reference \emph{tf} or \emph{tf-idf} vectors $v_p$ using cosine distance: $
    cos(v_p, v_q) = 1-\frac{v_p\cdot v_q}{\lVert v_p\rVert \lVert v_q \rVert}$. 
    It will take $O((P_c+2)*K)$ time \footnote{(as it takes time $O(P_c*K)$ for dot production, and time $O(2K)$ for calculating the norm of a queried vector. Note norms of the reference vectors were pre-calculated).} by using this measure where $P_c$ is the number of distinct \emph{tf} (or \emph{tf-idf, tc}) reference vectors and $K$ is the number of terms;  
    
    \item The \emph{Hamming distance} between \emph{term appearance} (ta) or \emph{binarized term counts} vectors 
    using logical exclusive function \emph{xor} as both vectors are binary. It will take $O(P_h*2K)$ time using this measure where $P_h$ is the number of distinct reference \emph{Hamming} vectors; 
    
    \item The \emph{Jaccard distance} between binary \emph{term appearance} (ta) vectors using logical \emph{xor} and \emph{or} functions as: 
    $J(s_1,s_2) = \frac{\sum xor(a_1,a_2)}{\sum or(a_1,a_2)}$. It will take $O(P_{ta}*4K)$ time by using this measure where $P_{ta}$ is the number of distinct reference \emph{ta} vectors. 
    
    \item The \emph{Jaccard bags} between \emph{tc} vectors (Section~\ref{sec:proposed_method}). It will take $O(P_{tc}*4K)$ time by using this measure where $P_{tc}$ is the number of distinct reference vectors of \emph{tc}. 
\end{enumerate}
   
As the length $n_2$ of original place signatures can be as high as $168$ (as shown in Table~\ref{tab:landmarks_Leeds}), $\mathbf{P}$ is generally large, and the number of distinct term count vectors $P_{tc}$ is generally higher than the number of distinct binary vectors of term appearance $P_{ta}$, it can be expected that \emph{O}(Edit distance) $\geq$ \emph{O}(Jaccard bags) $\geq$ \emph{O}(Jaccard Distance) $\geq$ \emph{O}(Hamming distance) $\geq$ \emph{O}(Cosine distance)).

\subsubsection{Evaluation of the recall and precision rate of different distance measures}
\label{sec:evaluation_measures}
\textcolor{black}{To compare the performance of different distance measures, a threshold between $[0,1]$ is selected for each distance measure other than \emph{Jaccard bags}, whereas a threshold between $[0.5,1]$ is selected}. Reference place signatures with a distance below the selected threshold are considered as `similar' candidates of a queried sample. Then, the recall rate, the average number of candidates per sample (which is equivalent to the inverse of \emph{Precision} rate), and the average time per query using different distance measure are compared. \textcolor{black}{At a same recall rate, a lower average number of candidates would suggest a better performance}.

\begin{figure}[htp!] 
\centering
\includegraphics[width = \textwidth]{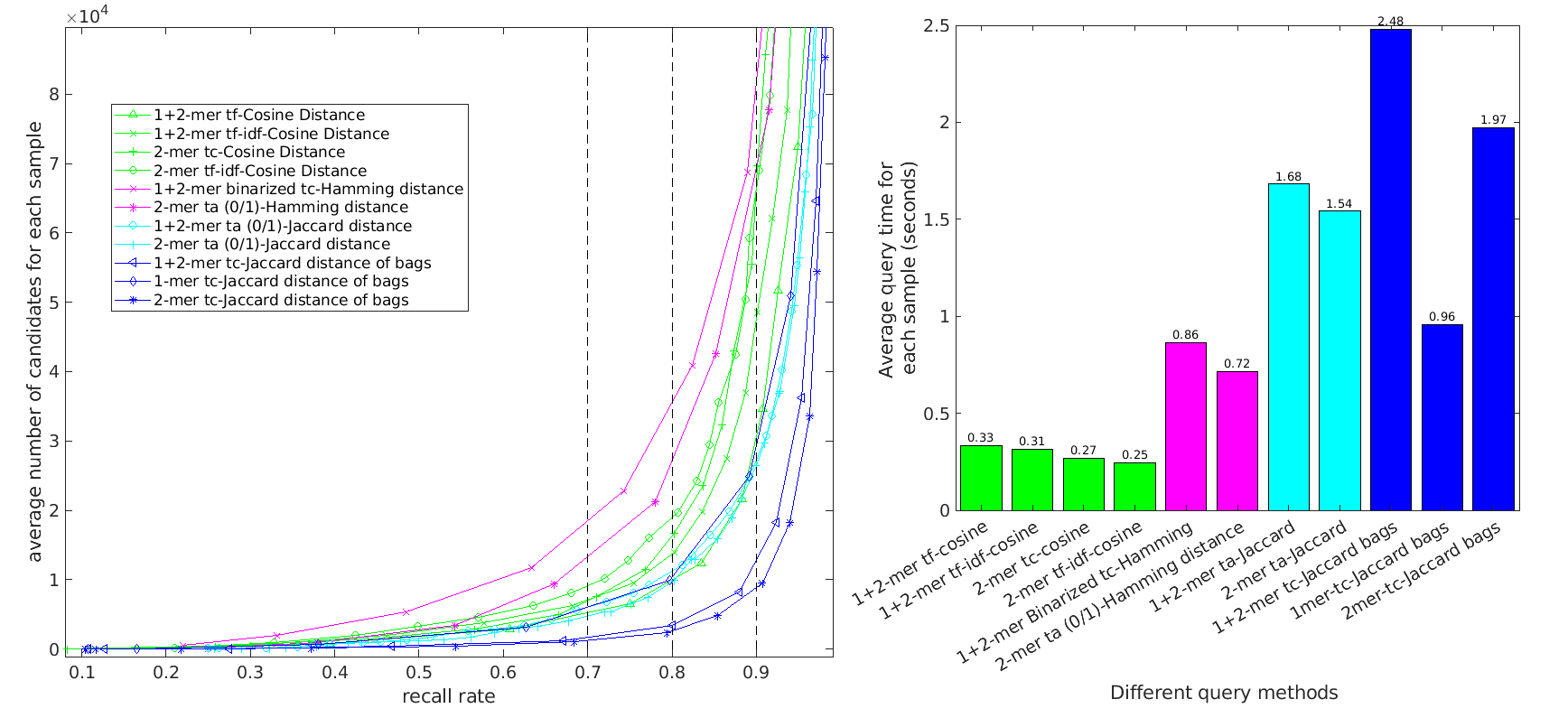}
\caption{Comparison of multiple distance measures for location retrieval using 1000 randomly selected and distorted place signatures. \textbf{Left)} The recall rate and the average number of candidates per quert; \textbf{right)} the average query time per sample.}
\label{fig:num_candidates1}
\end{figure} 

As shown in Figure~\ref{fig:num_candidates1} (left), the three methods using \emph{Jaccard bags} with \emph{tc}s (blue lines) gave the smallest number of candidates at a same recall rate, followed by the methods using \emph{Jaccard distance} with term appearance (cyan lines), cosine distance (green lines) and Hamming distance (magenta lines). With regards to the performance of different k-mers, generally, when the same distance measure is considered, the methods using \emph{2-mer} suggest lower numbers of candidates than those using \emph{1-mer}, or the combination of \emph{1-mer} and \emph{2-mers}. 
This may be because in some sense the ordering information between consecutive landmarks is kept in 2-mer terms but completely lost in 1-mer terms\footnote{Note that k-mers with $k>2$ are not used in this experiment due to the high memory demands. For example, selecting $k=3$ would convert each place signature into a $(10^3 +3^3+2^3= 1035)$-dimensional vector, which requires approximately ten times the memory for storing all reference data compared to using $k=2$ $(10^2 +3^2+2^2= 113)$.}. 
While for the average time per query, as seen from the right part of Figure~\ref{fig:num_candidates1}, both \emph{Jaccard distance} and \emph{Jaccard bags} take more time to query through the whole database (cyan and blue bars) than other methods, which is consistent with our theoretical analysis in Section~\ref{sec:time-theory}. 
Note \emph{edit distance} is not directly compared at this stage due to its unpractical time complexity as discussed in Section~\ref{sec:locationRetrieval}. However, once the number of candidates is reduced to an acceptable level using other methods, its performance will be compared in a later section. \textcolor{black}{All experiments were run on a PC with an \emph{Intel® Core™ i7-7500U CPU @ 2.70GHz} and one processor\footnote{using \emph{MatLAB R2021a}.}}.

\subsubsection{Evaluation of the locality sensitive Hashing methods on location retrieval}
\label{sec:evaluation_lsh}

\begin{figure}[htp!] 
\centering
\includegraphics[width = \linewidth]{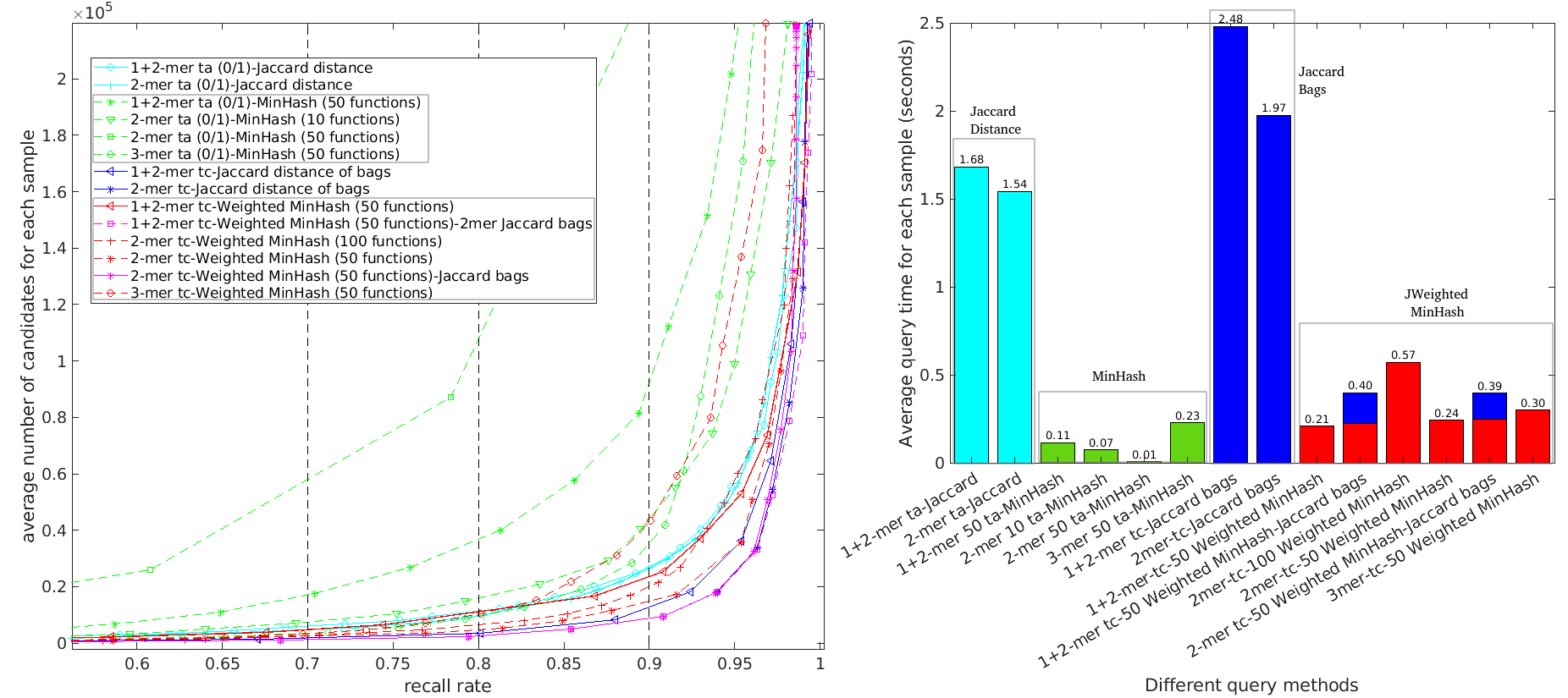}
\caption{Comparison of the performance of Jaccard Distance, Jaccard bags, MinHash and Weighted MinHash in location retrieval using $1000$ random place signatures, Left) The recall rate and the average number of candidates per query. (Right): the average query time per sample by using different methods. 
\label{fig:num_candidates2}
}
\end{figure}

\textcolor{black}{
As \emph{Jaccard distance} and \emph{Jaccard bags} based methods provide the best performance in recall-precision, experiments in this section evaluate whether locality-sensitive hashing techniques can reduce the initial query time \textcolor{black}{while keeping the performance in location retrieval}.
To do this,} the \emph{tc} vectors of reference place signatures are mapped to \emph{Weighted MinHash} vectors using $n$ random hash functions, and the \emph{term appearance} vectors are mapped to unweighted \textbf{MinHash} vectors~\citep{mining2014}. 
Both methods require time $O(P_l*n)$ which is linear to the number of hash functions $n$ and the number $P_l$ of unique hash vectors. As there are generally more unique \emph{tc} vectors than \emph{term appearance} vectors, we would expect \emph{Weighted MinHash} take slightly more time than the non-weighted method \textcolor{black}{if same numbers of hash functions are used}. \textcolor{black}{The recall rates and the average number of candidates per query of the two methods are shown in Figure~\ref{fig:num_candidates2}} were drawn by choosing different thresholds \textcolor{black}{for the proportion of different buckets} between $[0,1]$.
    



As shown in  Figure~\ref{fig:num_candidates2} (right)), the query time per sample was significantly reduced to \textcolor{black}{nearly ten percent of the original exhaustive searching time} by using approximate hashing methods (green bars and red bars). 

With regarding to their performance in identifying candidates, it can be seen from Figure~\ref{fig:num_candidates2} (left) that tests using \emph{weighted MinHash} (blue/magenta lines) generally suggest lower numbers of candidates than \emph{MinHash} (green lines) at the same level of recall rate. More specifically, the \emph{weighted} method with 50 random functions on \emph{2-mer} \emph{tc} vectors (red dashed line with stars) gave the lowest number of candidates when the recall rate is below $0.97$ while using the same method on \emph{1+2-mer} terms (red solid line with triangles) gave the lowest number of candidates when the recall rate is above $0.97$. 
\textcolor{black}{This suggests that \emph{2-mer} \emph{tc}s are enough for most cases while the combination of \emph{1-mer} and \emph{2-mer} terms are needed for some extreme cases.}

To keep as many true positives as possible, a large threshold is selected for \emph{Weighted MinHash} method with \emph{1+2-mer} terms, followed by \emph{Jaccard } on \emph{2-mer terms} (as seen in Figure~\ref{fig:num_candidates1}) for location refinement. 
The average number of candidates was brought down to $1/30$ of the size of the database at recall rate $0.97$, and half the size of the database at recall rate $1$. Then, feeding these candidates into the next 
exhaustive step will only take the corresponding proportion of the original computation time but retain the same levels of recall rate. 
For example, it can be seen from Figure~\ref{fig:num_candidates2} (left) that after combing these two steps, the curve of recall and average candidate numbers (shown as a magenta dashed line with squares) is almost the same as the exhaustive method
using \emph{Jaccard bags} on \emph{2-mer} terms (blue line with stars), but the average query time per sample was reduced from $1.97s$ to $0.40s$, shown as stacked red/blue bars in Figure~\ref{fig:num_candidates2} (right).
Note that the average number of candidates still increases sharply when the recall rate approaches $1$. This is because the same distance threshold was used on \emph{Jaccard Bags} for all queried samples and a large distance threshold will inevitably bring in more false positive for certain queries. In the next section, the proposed adaptive threshold on Jaccard bags is evaluated.

\subsubsection{Evaluating the contribution of adaptive distance thresholds on Jaccard distance of bags}

\begin{figure}[htp] 
\centering
\includegraphics[width = \linewidth]{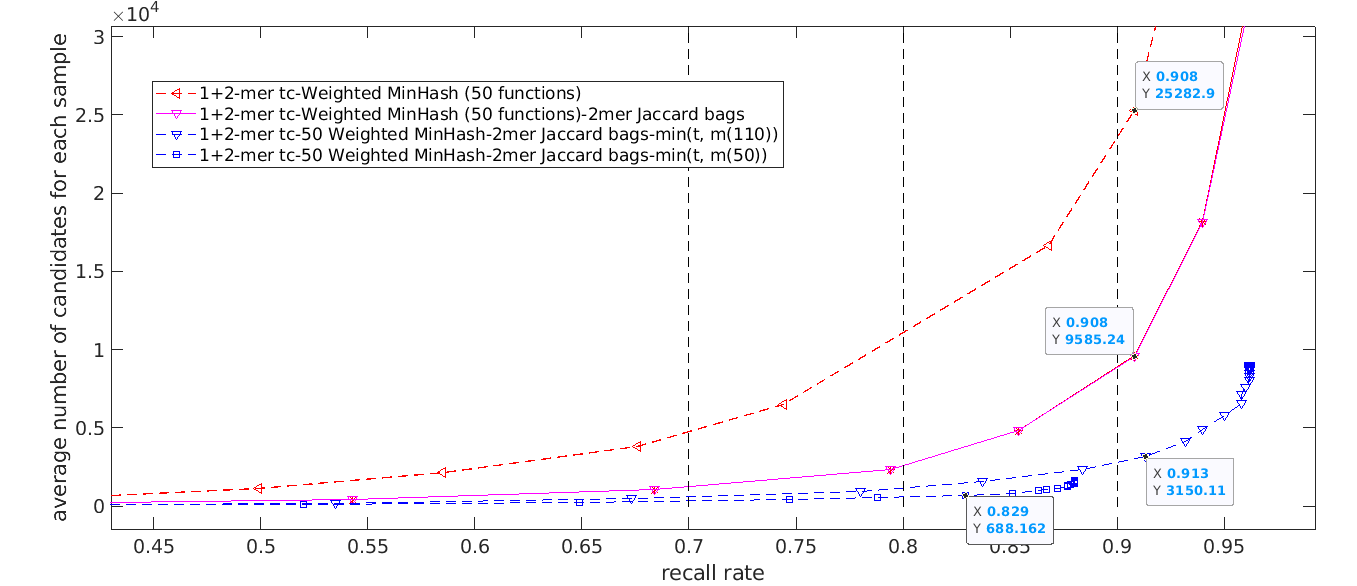}
\caption{\textcolor{black}{Comparison of the performance of Jaccard
distance of bags on place signature matching by using or not using an adaptive distance threshold.}
}
\label{fig:num_candidates3}
\end{figure}

In this experiment, an adaptive distance threshold is chosen for Jaccard bags for each queried sample by taking the smaller value of a fixed threshold $t$ and the $l^{th}$ lowest distance $m(l)$ between the queried sample and all candidates. The results of using $l=50$ and $l=110$ with a list of fixed values of $t$ between $0.5$ and $0.9$ are shown in Figure~\ref{fig:num_candidates3} as dashed lines in blue. It can be seen that the average numbers of candidates are significantly reduced. For example, at a recall rate $0.9$, the average number of candidates was reduced to around $3000$ by using $t_i = min([t, m(110)])$ on \emph{Jaccard Bags}, which is only one-third of the number when using a fixed threshold (line in magenta), and one-eighth of the number when using \emph{weighted MinHash} only (red dashed line). 

\subsubsection{Evaluating the contribution of edit distance by considering the uncertainty in landmarks perception}

\begin{figure}[htp] 
\centering
\includegraphics[width = \linewidth]{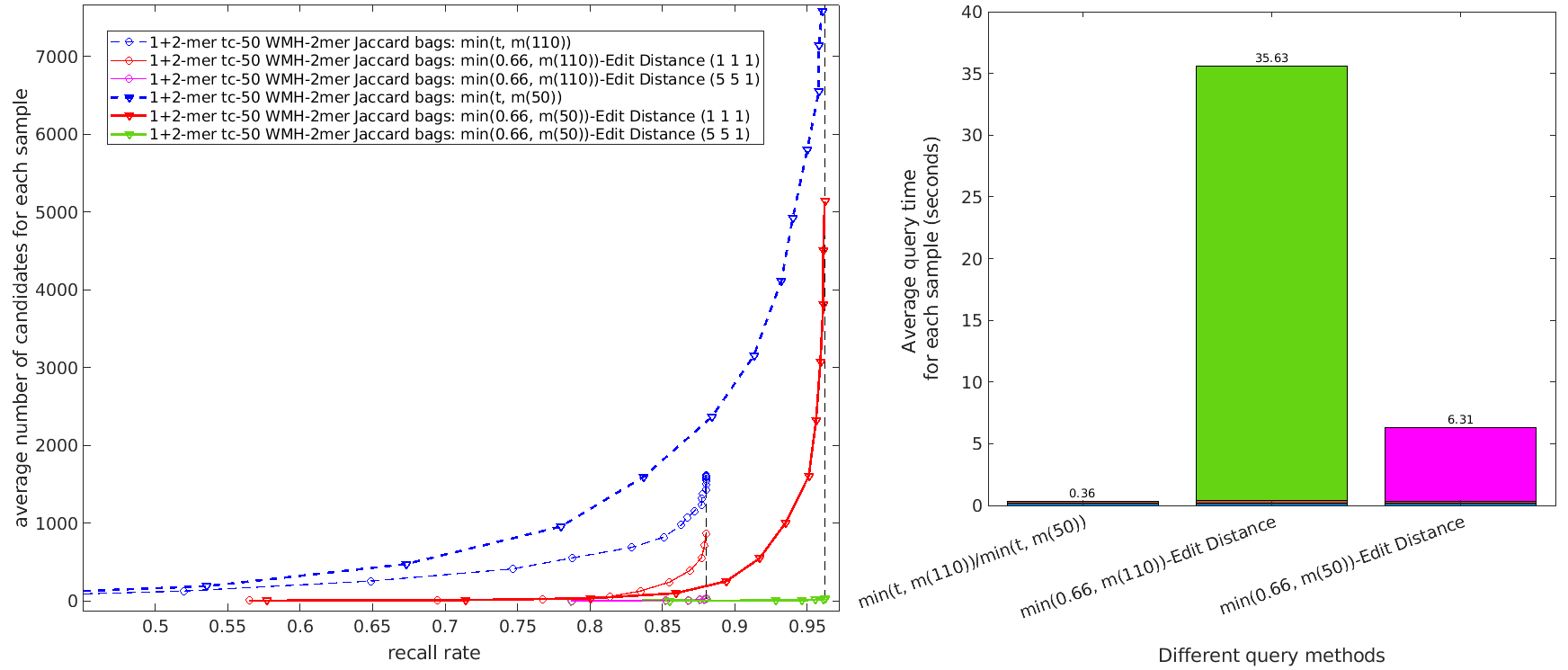}
\caption{The recall rate and average number of final candidates after using \emph{edit distance}. (Left) the results after adding the basic \emph{edit distance} are shown as red lines, and the results by considering the difference in landmark perception are shown in magenta and green. (Right) the average query time per sample. }
\label{fig:num_candidates_edit}
\end{figure}

\textcolor{black}{The candidates can be further refined using \emph{edit distance}. The results by setting all costs as one (i.e. $C_{subs}=C_{ins}=1$), and by considering the difference in landmarks perception with a higher substitution and insertion cost (i.e. $C_{subs}=C_{ins}=5$, $C_{del}=1$) are shown in Figure~\ref{fig:num_candidates_edit}.}
It can be seen that whether a large or a small threshold is chosen for \emph{Jaccard bags}, the corresponding maximum recall rates can all be achieved after adding the \emph{edit distance} while the average numbers of final candidates are reduced, especially by setting \textcolor{black}{different edit costs}. For example, by using $min(t,m(110))$, the average number of candidates was reduced from $7000$ to $36$ while keeping the maximum recall rate.
Although it is more time-expensive using using $min(t,m(50))$ as more candidates need to be examined by \emph{edit distance}, the maximum recall rate is slightly higher. 
Therefore, depending on the required precision of specific applications and the available computing resources, a different $l$ could be selected.

\subsubsection{Comparison with other methods}
\label{sec:compare-other}

\textcolor{black}{
As the concept proposed by \cite{Weng_2020} is close to our approach, the same \emph{Leeds} dataset was tested on that method by comparing the performance on place cells generation and signature query. After sampling the study area as regular grids for every $10$ meters on the \emph{East-West} and \emph{North-South} directions, $257$x$253$=$65021$ point locations and their corresponding place signatures (i.e. sequences of visible landmarks and quantified angle indices w.r.t North) were created. The average coverage of the reference place signatures is $248m^2$ compared to $1.71m^2$ using our approach (Tab.~\ref{tab:landmarks_Leeds}).
} 
\textcolor{black}{We then queried through this reference database using the same set of $1000$ simulated viewer observations of randomly selected and modified place signatures. Note that as accurate measurement of angles are not assumed to be available in our work, only the sequences of landmark types were used for query.}
\textcolor{black}{
With the exhaustive searching method, it took about $45.9$ seconds in average to calculate the edit distance between a query and all reference signatures. And the averaged query time was $0.36$s (before applying edit distance) and $6.31$s (after adding edit distance) in our approach by comparing all components of place signatures.}
\textcolor{black}{
Furthermore, only $4275$ of the created reference signatures using \cite{Weng_2020} were exactly the same as those $1178,445$ signatures encoded with our method (Tab.~\ref{tab:landmarks_Leeds}), this is partly because occlusions from buildings was not considered, and partly because the fixed-distance space division method only encode the observations from the centre point of each $10$x$10m^2$ square. It is difficult to quantify the difference between the observations from each centre point and elsewhere inside each grid as this will depend on the number, type and distribution pattern of landmarks surrounding each individual place cell. Our approach provides a much more complete and accurate description of the environment. It also provides much richer information to help infer viewers' movement if they start crossing different types place cell dividing lines, as previously discussed in section~\ref{sec:database}. }

\textcolor{black}{Concerning other vision-based place recognition methods relying on geo-referenced images (or LiDAR)  datasets, we think that they are not directly comparable to our approach, first because they are challenging to be scaled up in terms of the availability of the reference datasets and of the complexity of the retrieval process through the visual features being used; second because although they may provide a more precise location or 6D pose, this is often achievable within a smaller search area; but this does make them complementary to our approach which can be exploited upstream to reduce the search area quickly and thus reducing the overall time complexity.}

\section{Conclusion}
\label{sec:conclusion}

In this work, a qualitative place signature is proposed to describe locations using the \textcolor{black}{perceived qualitative} spatial relations between co-visible landmarks from viewers' perspective.
A framework is proposed to divide the space such that consistent place signatures can be observed inside each place cell; and a coarse-to-fine location retrieval method is proposed to identify viewers' possible location(s) by efficiently reducing the number of candidates using \emph{weighted MinHash} and \emph{Jaccard bags}, hypotheses refinement using \emph{edit distance} by considering the uncertainty in landmarks perception.
A reference database was created for the city of \emph{Leeds} in the UK using openly available landmark datasets and observations were simulated to evaluate the proposed location retrieval method. The results suggest that by using \emph{weighted MinHash} and \emph{Jaccard bags} with \emph{adaptive distance thresholds} for initial screening, the number of false positives can be significantly reduced to an acceptable level in less than a second; while incorporating the \emph{edit distance} by considering the difference in perception error could further reduce the number of candidate locations by keeping the high recall rate.
\textcolor{black}{As the proposed approach only requires storing the configuration of high-level landmarks and utilising an approximate Hashing step for fast screening, it is easy to scale up thus can be exploited as an upstream approach for other location/pose refinement techniques.} This technique could be used indoors, given a suitable database of landmarks and it will help in urban canyons where there are not enough satellites in view.
For future work, \textcolor{black}{we plan to create a more complete and coherent reference landmarks dataset by resolving the semantic and spatial discrepancy in different datasets using methods such as ontology alignment~\citep{Stoilos2005, li2009} and {geometry matching}~\citep{du2017}, to test the proposed method in virtual reality environment by considering landmarks with extended sizes, and to extend the proposed scheme from single location retrieval to trajectory identification by considering the movements of agents.}

\section*{Acknowledgement(s)}
\textcolor{black}{The authors thank the valuable discussions with Dr 
Bahman Soheilian and Dr Mathieu Bredif}.



\section*{Funding}
This research received no specific grant from any funding agency in the public, commercial, or not-for-profit sectors.

\bibliographystyle{plainnat}

\appendix
\section*{Appendices}
\addcontentsline{toc}{section}{Appendices}
\renewcommand{\thesubsection}{\Alph{subsection}}

\subsection{Links to websites and datasets mentioned in the main text}
\label{sec:links}

For all websites and datasets mentioned in the main text, their links are listed below based on order of appearance and the last accessed date is 2022-07-21.
{\small
\begin{itemize}
    \item {UK Ordnance Survey Roadside Asset Data Services}: 
    \url{https://osonline.maps.arcgis.com/apps/View/index.html?appid=61436bfcb44e4acaa99014a8f723e0e5}.
    \item {Find Open Data}: \url{https://data.gov.uk/}. 
    \item {OpenDataParis}: \url{http://opendata.paris.fr}.    
    \item {OpenStreetMap}: \url{https://www.openstreetmap.org}.
    \item Street lights: \url{https://data.gov.uk/dataset/3bcfdacd-705c-4370-a60a-64e17a1c9e03/street-lights-unmetered}.
    \item Traffic signals: \url{https://data.gov.uk/dataset/12c387e1-e65f-4d9e-a576-05dbc8f1d038/traffic-signals-in-leeds}.
    \item Bins: \url{https://data.gov.uk/dataset/adb73cd0-08ee-4963-b3a1-6f3066fcce0c/litter-bin-locations}.
    \item Trees : \url{https://data.gov.uk/dataset/d8fc56a4-4e57-4e22-b202-eff063545f72/trees-in-leeds-city-centre}.
    \item Bus stops: \url{https://datamillnorth.org/dataset/west-yorkshire-bus-stops}.
    \item Traffic light crossing points: \url{https://data.gov.uk/dataset/b1a820fe-6aff-404f-85cd-ff3484142040/pedestrian-crossing-points}.
    \item War memorials: \url{https://data.gov.uk/dataset/c99488fb-6244-4b84-8345-b7fda39bbd28/war-memorials}.
    \item Bike parking bays : \url{https://data.gov.uk/dataset/de5bc73e-ad4a-4282-b74a-a762204dad4f/leeds-city-centre-bike-bays}.
    \item Public convenience/toilets: \url{https://data.gov.uk/dataset/dee4f4dd-bea7-43c2-b6b4-098f7b91396b/public-toilets}.
\end{itemize}
}

\subsection{An illustration of the method proposed by~\citet{Schlieder1993}}
\label{sec:additional_illus}

\begin{figure}[ht!] 
    \centering  
    \includegraphics[width=0.7\linewidth]{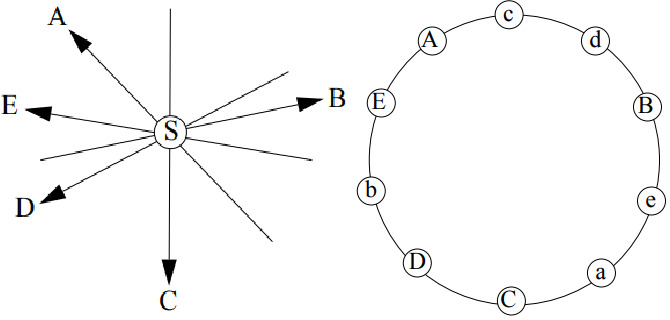}
    \caption{Given a location $S$ and five co-visible landmarks \{ABCDE\}, the panorama proposed by~\citep{Schlieder1993} can be written as $\langle AcdBeaCDbE\rangle$ where the lower case letters represent the complementary ($180^o$) directions w.r.t. the landmarks A, B, C, D, E.}
    \label{fig:augVec}
 \end{figure} 

\subsection{Algorithm to remove the visibility area occluded by buildings}
\label{sec:viewshed}

\textcolor{black}{Given the visibility zone of a landmark as a polygon stored in \emph{geom} (after removing areas directly in buildings), and the \emph{centre} of the landmark, the visibility area occluded by building can be calculated as follows:}
\renewcommand{\thetable}{A.\arabic{table}}
\setcounter{table}{0}
\begin{algorithm}
\caption{Remove visibility areas behind any buildings}\label{alg:euclid}
\begin{algorithmic}[1]
 \Procedure{mb\_visibility}{$geom, center$} 
      \State \texttt{rMax := max(dist(center,points))} for all $points$ on the visibility zone \texttt{geom}
      \State \texttt{ring := ST\_Dump(Boundary(geom))}
      \For{\texttt{$k=1:size(ring)-1$}} \Comment{loop over all edges on boundary}
        \State \texttt{p:=ring(k);\t q:=ring(k+1)}
        \State \texttt{dp:=dist(center,p);\t dq:=dist(center,q)}        
        \If{\texttt{dp=0 Or dq=0}}
        	\State \texttt{Continue} 
        \EndIf
        \State \texttt{dp := 2*rMax/dp; dq := 2*rMax/dq}
        \State \texttt{p2.x := p.x+dp*(p.x-center.x); p2.y := p.y+dp*(p.y-center.y)}
        \State \texttt{q2.x := q.x+dq*(q.x-center.x); q2.y := q.y+dq*(q.y-center.y)}        
        \State \texttt{occluder := Polygon([p,q,q2,p2,p])}  
        \State $res := st\_diff(geom,occluder)$
      \EndFor       
      \State \textbf{return} $res$
 \EndProcedure
\end{algorithmic}
\end{algorithm}

\begin{table}[htb!]
\begin{center} 
\label{tab:transitivity}
\caption{The composition table of \emph{IOC} relations.} 
 \begin{tabular}{|l||*{11}{c|}}\hline
\rowcolor{yellow!50}
\setrow{\bfseries}&\multicolumn{11}{|c|}{B$r_2$C}\\\hline
\rowcolor{yellow!50}
\setrow{\bfseries}A$r_1$B
&\makebox[1em]{\cellcolor{blue!25}\textbf{$p$}}&\makebox[1em]{$m$}&\makebox[1em]{$o^+$}
&\makebox[1em]{$o^-$}&\makebox[1em]{$si^+$}&\makebox[1em]{$si^-$}&\makebox[1em]{$di^+$}&\makebox[1em]{$di^-$}&\makebox[1em]{$fi^+$}&\makebox[1em]{$fi^-$}&\makebox[1em]{\textbf{\cellcolor{blue!25}$c^+$}}\\\hline\hline
\cellcolor{blue!25}{$p$} &\cellcolor{blue!25}$p$&$p$&$p$&$p$&$p$&$p$&$p$&$p$&$p$&$p$&\cellcolor{blue!25}$p$\\\hline
$m$ &$p$&$p$&$p$&$p$&$m$&$m$&$p$&$p$&$p$&$p$&$m$\\\hline
$o^+$ &$p$&$p$&\makecell{$p$\\$m$\\$o^+$}&\makecell{$p$\\$m$\\$o^{+,-}$}&\makecell{$di^+$\\$fi^+$\\$o^+$}&\makecell{$di^{+,-}$\\$fi^{+,-}$\\$o^{+,-}$}&\makecell{$p$\\$m$\\$di^+$\\$fi^+$\\$o^+$}&\makecell{$p$\\$m$\\$di^{+,-}$\\$fi^{+,-}$\\$o^{+,-}$}&\makecell{$p$\\$m$\\$o^+$}&\makecell{$p$\\$m$\\$o^{+,-}$}&$o^+$\\\hline
$o^-$  &$p$&$p$&\makecell{$p$\\$m$\\$o^{+,-}$}&\makecell{$p$\\$m$\\$o^-$}&\makecell{$di^{+,-}$\\$fi^{+,-}$\\$o^{+,-}$}&\makecell{$di^-$\\$fi^-$\\$o^-$}&\makecell{$p$\\$m$\\$di^{+,-}$\\$fi^{+,-}$\\$o^{+,-}$}&\makecell{$p$\\$m$\\$di^-$\\$fi^-$\\$o^-$}&\makecell{$p$\\$m$\\$o^{+,-}$}&\makecell{$p$\\$m$\\$o^-$}&\makecell{$o^{+,-}$}\\\hline
$si^+$ &\makecell{$p$\\$m$\\$di^{+,-}$\\$fi^{+,-}$\\$o^{+,-}$}&\makecell{$di^{+,-}$\\$fi^{+,-}$\\$o^{+,-}$}
&\makecell{$di^+$\\$fi^+$\\$o^+$}
&\makecell{$di^{+,-}$\\$fi^{+,-}$\\$o^{+,-}$}&$si^+$&\makecell{$si^{+,-}$}&$di^+$&\makecell{$di^{+,-}$}&$di^+$&\makecell{$di^{+,-}$}&$si^+$\\\hline

$si^-$ &\makecell{$p$\\$m$\\$di^{+,-}$\\$fi^{+,-}$\\$o^{+,-}$}&\makecell{$di^{+,-}$\\$fi^{+,-}$\\$o^{+,-}$}&\makecell{$di^{+,-}$\\$fi^{+,-}$\\$o^{+,-}$}&\makecell{$di^-$\\$fi^-$\\$o^-$}&{$si^{+,-}$}&$si^-$&\makecell{$di^{+,-}$}&$di^-$&\makecell{$di^{+,-}$}&$di^-$&$si^{+,-}$\\\hline
$di^+$ &\makecell{$p$\\$m$\\$di^{+}$\\$fi^{+}$\\$o^{+}$}&\makecell{$di^{+}$\\$fi^{+}$\\$o^{+}$}&\makecell{$di^+$\\$fi^+$\\$o^+$}&\makecell{$di^{+,-}$\\$fi^{+,-}$\\$o^{+,-}$}&$di^+$&$di^{+,-}$&$di^+$&$di^{+,-}$&$di^+$&$di^{+,-}$&$di^+$
\\\hline
$di^-$ &\makecell{$p$\\$m$\\$di^{+,-}$\\$fi^{+,-}$\\$o^{+,-}$}&\makecell{$di^{+,-}$\\$fi^{+,-}$\\$o^{+,-}$}&\makecell{$di^{+,-}$\\$fi^{+,-}$\\$o^{+,-}$}&\makecell{$di^-$\\$fi^-$\\$o^-$}&$di^{+,-}$&$di^-$&$di^{+,-}$&$di^-$&$di^{+,-}$&$di^-$&$di^{+,-}$
\\\hline
$fi^+$ &$p$&$m$&$o^+$&$o^{+,-}$&$di^+$&$di^{+,-}$&$di^{+}$&$di^{+,-}$&$fi^{+}$&$fi^{+,-}$&$fi^+$\\\hline
$fi^-$ &$p$&$m$&$o^{+,-}$&$o^-$&$di^{+,-}$&$di^-$&$di^{+,-}$&$di^{-}$&$fi^{+,-}$&$fi^{-}$&$fi^{+,-}$\\\hline
\cellcolor{blue!25}$c^+$ &\cellcolor{blue!25}$p$&$m$&$o^+$&$o^{+,-}$&$si^+$&$si^{+,-}$&$di^{+}$&$di^{+,-}$&$fi^{+}$&$fi^{+,-}$&\cellcolor{blue!25}$c^+$\\\hline
\end{tabular}
\end{center} 
\end{table}

\subsection{The uncertainty of perception on landmarks and spatial relations}
\label{sec:perception_error}
\renewcommand{\thefigure}{B.\arabic{figure}}
\setcounter{figure}{0}
\renewcommand{\thetable}{B.\arabic{table}}
\setcounter{table}{0}

\subsubsection{Defining the likelihood of landmarks perception errors}
 
The likelihood of a landmark being incorrectly perceived (e.g. deleted/missed due to occlusion) may vary between different types of landmarks \textcolor{black}{when other conditions are the same}. For example, compared with road signs, city trees are generally less likely to be occluded by vehicles as they are usually taller and wider. 
Other than the impact $w^n_1$ of the visual attributes of landmarks (as discussed in Section~\ref{sec:proposed_method}), such as landmark \textbf{height, width/size, visual salience}, two additional factors can be considered:
\begin{enumerate}
    
    \item \textbf{$w^n_2$: weight of the substitution cost in landmarks perception based on the detailed level of landmark semantics}. In addition to the general `type' of a landmark, a levelled representation strategy \emph{T|sT|n} can be used to further enrich their semantic information by combining their general type (\emph{T}), e.g. `shop'), sub-type (\emph{ST}, e.g. `restaurant', `laundry'), and name (\emph{n}) if there are any. As it is generally easier for a human to get the more general information correct than the details, for example, it is easier to identify a \emph{bin} than providing its exact \emph{material}, the (weight of) the cost of the change of more general information would be higher than more detailed information. The total substitution cost ccan be updated as:
    \begin{equation}
        C^{subst} = w^{n,1}_2C_{node}^{type} + w^{n,2}_2cost_{sub\_type}+ w^{n,3}_2cost_{name}
        \label{eq:node_subst_cost2}
    \end{equation}
    For example, if we set $w^{n,1}_2=0.6, w^{n,2}_2=0.3, w^{n,3}_2=0.1$, the total substitution costs between the eight observed landmarks shown in Table~\ref{tab:exm_node_cost_detail} and the reference \emph{\{T|ST|N\}} are listed. The smaller the substitution cost is, the more similar two landmarks will be.
\begin{table}[htb!]
\begin{center} 
\label{tab:exm_node_cost_detail}
\parbox{12cm}{\caption{Example oft the landmark substitution cost w.r.t. a reference landmark \emph{L: $\langle T|ST|N \rangle$}. The smaller the substitution cost is, the more similar two landmarks will be.} }
  \begin{tabular}{ | l| l | r|  } \hline 
  Observation & \makecell[l]{Non-weighted   cost} & \makecell[l]{Weighted substitution cost} \\ \hline
  $T_2|ST_2|N_2$ &$1+1+1=3$&$0.6+0.3+0.1=1$ \\  \hline      
  $T_2|ST_2|N$ &$1+1+0=2$&$0.6+0.3+0.1*0=0.9$ \\  \hline 
  $T_2|ST|N_2$ &$1+0+1=2$&$0.6+0.3*0+0.1=0.7$ \\  \hline  
  $T_2|ST|N$ &$1+0+0=1$&$0.6+0.3*0+0.1*0=0.6$ \\  \hline  
  $T|ST_2|N_2$ &$0+1+1=2$&$0.6*0+0.3+0.1=0.4$ \\  \hline  
  $T|ST_2|N$ &$0+1+0=1$&$0.6*0+0.3+0.1*0=0.3$ \\  \hline 
  $T|ST|N_2$ &$0+0+1=1$&$0.6*0+0.3*0+0.1=0.1$ \\  \hline 
  $T|ST|N$ &$0+0+0=0$&$0.6*0+0.3*0+0.1*0=0$ \\  \hline
\end{tabular}  
\end{center} 
\end{table}
\item \textbf{$w^n_3$: weight of the change cost based on the closeness of landmarks to the viewer.} While the difference brought in by the \textbf{height, size and visual salience} attributes can be roughly predefined using the above strategy, 
the difference brought in by the closeness of landmarks to a viewer is location-dependent. When a viewer is closer to a landmark, there should be less chance for the landmark to be incorrectly perceived, therefore with a higher cost of change. Note that there is no need for a viewer to provide the distance information. By assessing the location of landmarks and the centroids of place cells, a weight $w^n_3$ can be assigned to the change cost of each landmark, which is inversely proportional to the distance between a landmark and a viewer as, written as $w^n_3\propto \frac{1}{distance(viewer, landmark)}$.    
As a maximum visible range $D_{max}^i$ was predefined for each type of landmarks (i = signs, traffic lights, etc.), the corresponding weighting value $w^n_3$ can be defined as:
    \begin{equation}
        w^n_3\propto \frac{D_{max}^i-D^i}{D_{max}^i}
    \end{equation} 
where a small $D^i$ means that the location is close to the landmark, thus it should be very unlikely for the landmark to be incorrectly perceived ($w^n_3\approx1$); while a $D^i$ close to $D_{max}$ means that the locations is near the boundary of the visibility range of the landmark, thus more likely to be incorrectly perceived ($w^n_3\approx0$). We could also impose a tiny weight in this case.
\end{enumerate} 
 
\subsubsection{Defining the likelihood of perception errors on spatial relations}
\begin{figure}[htp] 
    \centering  
    \subfigure[The likelihood of edge perception errors for two 100-metres apart landmarks \emph{A, B}.]{\label{fig:edge_u1}
      \includegraphics[width = 0.48\linewidth]{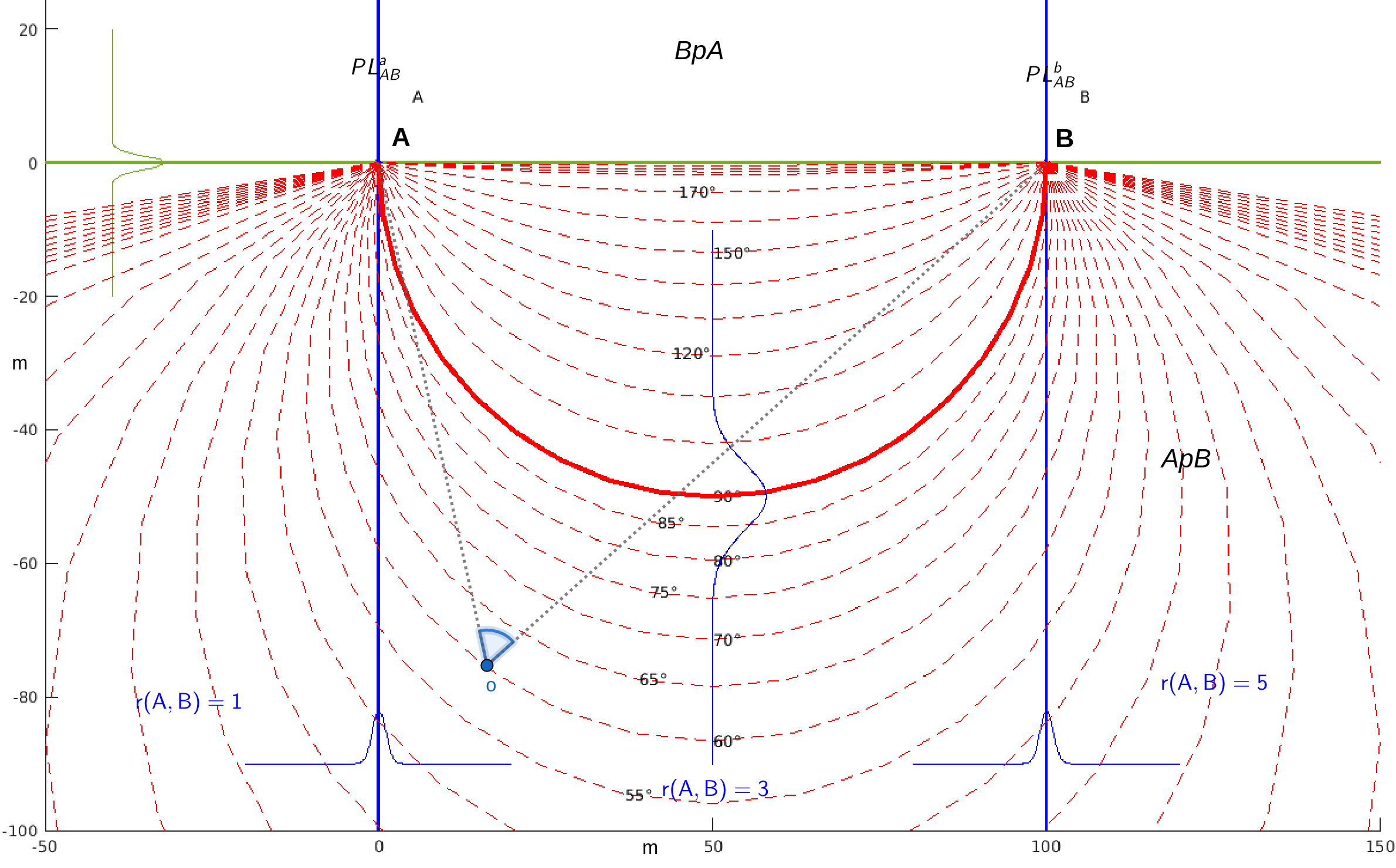} }
  \subfigure[The likelihood of edge perception errors for two two-metres apart landmarks \emph{A, B}.]{
  \label{fig:edge_u2}
      \includegraphics[width = 0.48\linewidth]{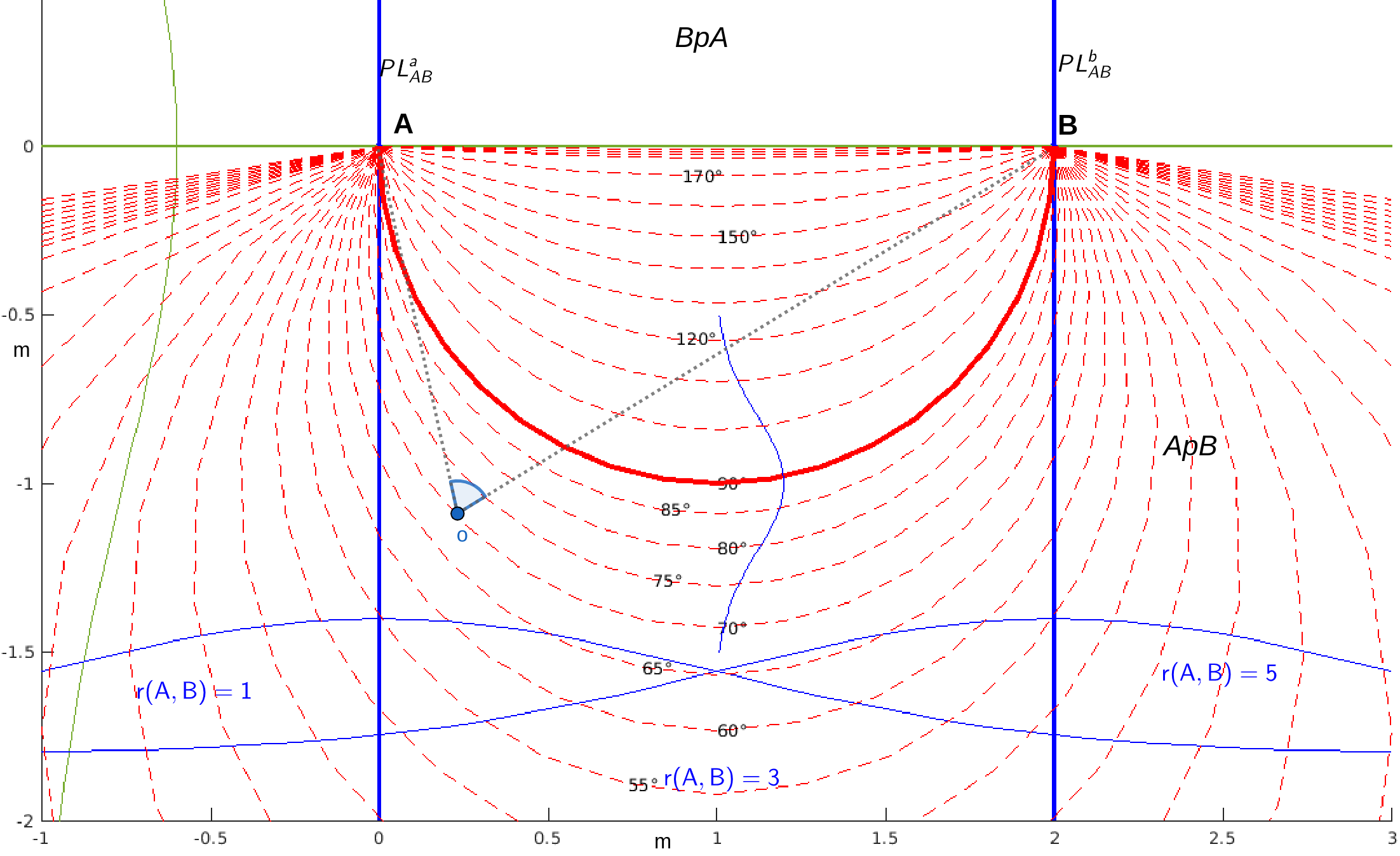}}
    \caption{The likelihood of perception errors on the ordering relation, relative orientations and qualitative angles are approximated as Gaussian distributions w.r.t. the distance to the corresponding dividing line or the angular difference to $90^o$. Note the plotted probability density functions in the figures are only shown for illustration purpose.}
    \label{fig:edge_uncertainty}
 \end{figure}

As discussed earlier in Section~\ref{sec:signatureDefinition}, three types of lines between each pair of co-visible landmarks are used to divide the space into place cells such that the corresponding type of spatial relations can be consistently observed in each cell, including \emph{Straight line} for \emph{ordering relation}, \emph{Perpendicular Lines} for \emph{relative orientations} and \emph{Circular line} for \emph{relative angles}.  
As shown in Figure~\ref{fig:edge_uncertainty}, when a viewer is approaching either of these dividing lines, they would be less confident in their judgement of the corresponding spatial relation being observed. The more uncertain this observation is, the more likely the observation could be different from those stored in the reference database, thus the lower the (weight of the) change cost of this relation would be. 

More specifically, we can model these uncertainties as independent Gaussian distributions w.r.t. the  viewer's distance to the corresponding dividing line or the angular difference to $90^o$. The three probability density functions of \emph{ordering relation}, \emph{relative orientation} and \emph{qualitative angle} at a location $o$ can be written as:
\begin{equation}
    f(o)_1 = \frac{1}{\sigma_1\sqrt{2\pi}}{\exp^{-\frac{1}{2}\left(\frac{d(o,\overleftrightarrow{AB})}{\sigma_1}\right)^{\!2}}}
    \label{eq:per_un1}
\end{equation}
where $d(o,\overleftrightarrow{AB})$ is the distance between a location $o$ and line $\overleftrightarrow{AB}$\footnote{When viewers are between the two perpendicular lines, i.e. \emph{relative orientation} index $r(A,B)=3$, it would not unlikely for them misperceive the order of the two landmarks; but when they are outside of this area, i.e. $r(A,B)=(1,\;5)$, they could be less confident of the observed order as they approach the line connecting the two landmarks.}.
\begin{align*}f(o)_2 = max(f(o)_2^1,\;f(o)_2^2)
    &= max(\frac{1}{\sigma_2^1\sqrt{2\pi}}{\exp^{-\frac{1}{2}\left(\frac{d(o,PL^{AB}_a)}{\sigma_2^1}\right)^{\!2}}}, \frac{1}{\sigma_2^2\sqrt{2\pi}}{\exp^{-\frac{1}{2}\left(\frac{d(o,PL^{AB}bb)}{\sigma_2^2}\right)^{\!2}}})\\
    &=
    \frac{1}{\sigma_2\sqrt{2\pi}}{\exp^{-\frac{1}{2}\left(\frac{min(d(o,PL^{AB}_a), d(o,PL^{AB}_b))}{\sigma_2}\right)^{\!2}}}
    \label{eq:per_un2}
\end{align*}
where $d(o,PL^{AB}_a)$ and $d(o,PL^{AB}_b)$ are the distances between viewer $o$ and the two perpendicular lines $PL^{AB}_a$ and $PL^{AB}_b$, and the standard deviation $\sigma_2^1$ and $\sigma_2^2$ are assumed equally as $\sigma_2$. Note that when two landmarks are very close to each other, such as the example shown in Figure~\ref{fig:edge_u2}, viewers may have great uncertainty in deciding where they are located with respect to the two perpendicular lines.
\begin{equation}
    f(o)_3 = \frac{1}{\sigma_3\sqrt{2\pi}}{\exp^{-\frac{1}{2}\left(\frac{|90^o-o_{ang}|}{\sigma_3}\right)^{\!2}}}
\end{equation}
where $o_{ang}$ is the observed angle between \emph{A, B} from a location $o$. 
    
Then, the weight of the corresponding change cost of a spatial relation can be defined inversely proportional to the probability of individual incorrect perception, as:
\begin{equation}
        w^r_k\propto \frac{1}{f(o)_k}, k=1,2,3
\end{equation}  
    

\subsection{Impact of the uncertainty of landmark location}
\label{sec:location_uncertainty}

\subsubsection{Modelling of point landmark uncertainty}
The uncertainty of landmark locations could be caused by many reasons in the data capture process~\citep{gis_uncertainty}, such as low accuracy of GPS devices and map digitising. 
Given the coordinates of two landmarks $A(x_1, y_1)$ and $B(x_2, y_2)$, if we model their uncertainty on the \emph{X} (Easting) and \emph{Y} (Northing) directions using two independent multivariate Gaussian distributions, and assume the errors on the \emph{X} and \emph{Y} directions are uncorrelated, the co-variance matrix of the four location parameters $x_{AB}(x_1, y_1, {x_2}, y_2)$ can be written as:
\begin{equation} \mathbf{\Sigma^{A,B}}=
\begin{bmatrix}
\mathbf{\Sigma^{A}} &0\\  0&\mathbf{\Sigma^{B}}
\end{bmatrix}
=
\begin{bmatrix}
\begin{array}{cc|cc}
  \sigma_{x_1}^2 & \sigma_{x_1,y_1} & 0&0\\
\sigma_{x_1,y_1} & \sigma_{y_1}^2 & 0&0\\ \hline
0&0 & \sigma_{x_2}^2 & \sigma_{x_2,y_2}\\
0&0&\sigma_{x_2,y_2} & \sigma_{y_2}^2 
\end{array}
\end{bmatrix}
=
\begin{bmatrix}
\begin{array}{cccc}
  \sigma_{x_1}^2 & 0 &0 &0\\
  0& \sigma_{y_1}^2 & 0&0\\ 
0&0 & \sigma_{x_2}^2 &0 \\
0&0& 0& \sigma_{y_2}^2 
\end{array}
\end{bmatrix}
\label{eq:landmark_un}
\end{equation}
The values of $(\sigma_{x_1},\sigma_{y_1})$ and $(\sigma_{x_2}, \sigma_{y_2})$ could be determined based on the sources of data. 
Assume the uncertain level of landmarks is much smaller than their visibility range, this uncertainty will not affect which landmarks will be observed, but rather how the spatial relations between landmarks will be perceived.
As the three types of qualitative spatial relations used in this work are relating to the three types of \emph{Dividing Lines: SL, PL, CL}, we first propagate the uncertainty in landmark locations to these lines. 
 
\subsubsection{Propagating the uncertainty of two co-visible landmarks to their connecting line}
\label{sec:sl_un}

Given the coordinates of two landmarks $A(x_1, y_1)$ and $B(x_2, y_2)$, the \textbf{\emph{Straight Line ($SL^{AB}$)}} connecting them can be defined as $\mathbf{Y=aX+b}$, where $a$, $b$ are:
    \begin{equation}
    \mathbf{a} = \frac{y_2-y_1}{x_2-x_1};\; \mathbf{b}=\frac{x_2*y_1-x_1*y_2}{x_2-x_1}  
    \label{eq:line:equation1}
    \end{equation}
When $x_1=x_2$, the line equation above can be re-written as $\mathbf{X=a'Y+b'}$ where 
    \begin{equation}
    \mathbf{a'} = \frac{x_2-x_1}{y_2-y_1};\; \mathbf{b'}=\frac{x_1y_2-x_2y_1}{y_2-y_1} 
    \label{eq:line:equation2}
    \end{equation}
Then, the uncertainty of landmark locations $\mathbf{\Sigma^{A,B}}$ (as defined in Equation~\ref{eq:landmark_un}) can be propagated to $a/a'$ and $b/b'$ by linearizing the two non-linear functions using first-order Taylor series propagation, written as: 
\begin{equation}
 \mathbf{\Sigma^{a,b;a',b'}} =
 \mathbf{J_0\Sigma^{A,B}J_0^T}
\label{eq:sl_un}    
\end{equation}
where $\mathbf{J_0}$ is the Jacobian matrix containing the first-order partial derivatives of $a,\;b,\;a'\;b'$ on $(x_1, y_1, {x_2}, y_2)$, calculated as follows:
    \begin{equation} 
    \mathbf{J_0}=
\begin{bmatrix}
  \frac{\partial a}{\partial x_1} & 
    \frac{\partial a}{\partial y_1} & 
    \frac{\partial a}{\partial x_2}&
    \frac{\partial a}{\partial y_2}\\[1ex]
  \frac{\partial b}{\partial x_1} & 
    \frac{\partial b}{\partial y_1} & 
    \frac{\partial b}{\partial x_2}&
    \frac{\partial b}{\partial y_2}\\[1ex]
    \frac{\partial a'}{\partial x_1} & 
    \frac{\partial a'}{\partial y_1} & 
    \frac{\partial a'}{\partial x_2}&
    \frac{\partial a'}{\partial y_2}\\[1ex]
  \frac{\partial b'}{\partial x_1} & 
    \frac{\partial b'}{\partial y_1} & 
    \frac{\partial b'}{\partial x_2}&
    \frac{\partial b'}{\partial y_2}
\end{bmatrix} \\
=
\begin{bmatrix}
  \frac{y_2-y_1}{(x_2-x_1)^2} & \frac{-1}{x_2-x_1} & \frac{y_1-y_2}{(x_2-x_1)^2}& \frac{1}{x_2-x_1}\\[1ex] 
  \frac{x_2(y_1-y_2)}{(x_2-x_1)^2}&\frac{x_2}{x_2-x_1}&\frac{x_1(y_2-y_1)}{(x_2-x_1)^2}&\frac{-x_1}{x_2-x_1}\\[1ex]
  \frac{-1}{y_2-y_1} & \frac{x_2-x_1}{(y_2-y_1)^2} &\frac{1}{y_2-y_1}& \frac{x_1-x_2}{(y_2-y_1)^2} \\[1ex] 
  \frac{y_2}{y_2-y_1}&\frac{y_2(x_1-x_2)}{(y_2-y_1)^2}&\frac{-y_1}{y_2-y_1}&\frac{y_1(x_2-x_1)}{(y_2-y_1)^2}
\end{bmatrix}
\label{eq:Jaco1}
\end{equation}
Then, we can get the uncertainty of a point on the line based on the uncertainty of the slope $\mathbf{a},\;\mathbf{a'}$ and $\mathbf{b},\;\mathbf{b'}$  in $\mathbf{\Sigma^{a,b;a',b'}}$. There are three situations:

\begin{enumerate}
    \item When $a\neq0$ and $a'\neq0$, given a list of $X$ or $Y$ values, we can calculate the corresponding $Y$ or $X$ using $\mathbf{Y=aX+b}$ and $\mathbf{X=a'Y+b'}$ and the uncertainties on both directions as:
       \begin{equation}
        \mathbf{\Sigma^{X,Y}}
        =\begin{bmatrix}        \mathbf{\sigma^2_{X}} & \mathbf{\sigma_{X,Y}}\\
        \mathbf{\sigma_{X,Y}}&
        \mathbf{\sigma^2_{Y}}
        \end{bmatrix}=
        \mathbf{J_0^2\Sigma^{a,b;a',b'}{J_0^2}^T}
        \label{eq:sl_un3}
        \end{equation}
    where $
    \mathbf{J_0^2}= \begin{bmatrix}  0 & 0 & Y & 1\\[1ex] 
    X & 1 & 0 & 0\end{bmatrix}$. If we expand the equation, the error on $X$ and $Y$ directions are equivalent to:
     \begin{equation}
         \begin{array}{c}
         \sigma^2_{X}=\sigma^2_{a'}Y^2+\sigma^2_{b'} + 2\sigma_{a',b'}X;\;
         \sigma^2_{Y}=\sigma^2_{a}X^2+\sigma^2_{b} + 2\sigma_{a,b}X
    \end{array}
    \label{eq:sl_4}
     \end{equation}
     
    \item When $a=0$, the line is horizontal. Therefore, $\sigma^2_{X} = 0$ and the upper-left part of $\mathbf{\Sigma^{a,b;a',b'}}$ is used to estimate the uncertainty $\sigma^2_{Y}$ of points on the $Y$ direction;
    
    \item When $a'=0$, the line is vertical. Therefore, $\sigma^2_{Y}= 0$ and the bottom-right part of $\mathbf{\Sigma^{a,b;a',b'}}$ is used to estimate the uncertainty $\sigma^2_{X}$ of points on the $X$ direction.
\end{enumerate}

\begin{figure}[htp] 
    \centering  
      \includegraphics[width = \linewidth]{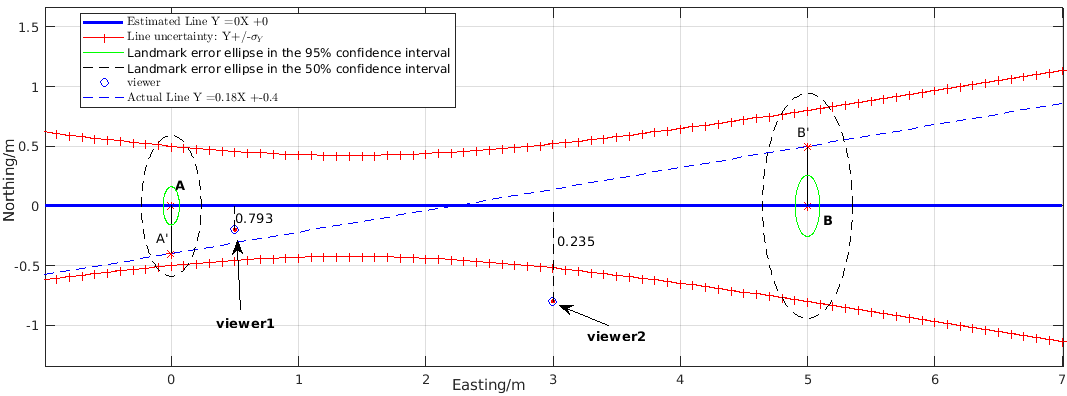} 
    \caption{Propagating the uncertainty of landmark $A(0,0)$ and $B(5,0.8)$ to the \textbf{\emph{Straight Line}} connecting the two landmarks. Landmark uncertainty is shown as error ellipses while the uncertainty of line $SL^{AB}$ is shown with upper and lower bound red curves. In the areas between the two red curves, a viewer is highly likely to observe an inconsistent ordering relation between the two landmarks. }
    \label{fig:landmark_line1}
 \end{figure}

\noindent For example, as shown in Figure~\ref{fig:landmark_line1}, given a landmark $A(0,0)$ and $B(5,0)$ with uncertainty $(\sigma_{x_1},\sigma_{y_1},\sigma_{x_2},\sigma_{y_2})$= (0.2, 0.5 0.3 0.8) shown as green and black error ellipses in the $95\%$ and $50\%$ confidence levels, the uncertainty of slope $a$ and intercept $b$ are calculated using Equation~\ref{eq:sl_un} as $\mathbf{\Sigma^{a,b}}=\begin{pmatrix}
    0.0356 &  -0.050\\   -0.050  &  0.2500
\end{pmatrix}$. The uncertainty of points on the line AB are shown using the upper and lower error bound (red curves) of the line by adding and subtracting the standard deviation of $Y$ at the $X$ of each point. 

\noindent\textbf{Probability of a viewer to observe an inconsistent ordering relation between \emph{A} and \emph{B}.} Since the location of landmarks A and B are uncertain, it is highly likely that the actual line AB could be anywhere between the upper and lower bound curves. Therefore, the observed ordering relation between A and B could be different from what we expect. For example, as shown in Figure~\ref{fig:landmark_line1}, assume there are two viewers \textbf{viewer 1}: $o_1(0.5, -0.2)$ and \textbf{viewer 2}: $o_2(3, -0.8)$ located below the solid blue line AB, we would expect both of them to observe \emph{A} followed by \emph{B}. However, as the locations of \emph{A, B} are actually different from those stored in the map, the actual dividing line is different (shown as a dashed blue line). In this case, although \textbf{viewer 2} can still observe \emph{A} followed by \emph{B}, \textbf{viewer 1} is actually on the other side of the line $SL^{A'B'}$ and would observe \emph{B} followed by A. Therefore, in the areas between the two red curves, a viewer is highly likely to observe an inconsistent ordering relation.

In fact, depending on the location of a viewer $\mathbf{o(x_o, y_o)}$ and the uncertainty of the connecting line $\mathbf{\Sigma^{a,b}}$, the likelihood for the viewer to observe a different ordering relation can be estimated by first finding the corresponding point (i.e. nearest point) of the viewer on line $SL^{AB}$ and estimating the corresponding uncertainty of this point $\mathbf{\Sigma^{X_o^1,Y_o^1}}$ using Equation~\ref{eq:sl_un3}.
\begin{align*}
    X_o^1 = \frac{ay_o+x_o-ab}{a^2+1},\; Y_o^1=aX_o+b
\end{align*}
Then, we can approximate the probability $fl(o)_1$ of a viewer being on the straight line, which is considered as their probability to observe an inconsistent ordering relation due to the uncertainty of landmark locations, as follows:
\begin{equation}
fl(o)_1 = \frac{1}{2\pi\sqrt{\mathbf{|\Sigma_{X_o^1,Y_o^1}|}}}{\exp^{\left(-\frac{1}{2}(\mathbf{x-\mu}){\mathbf{\Sigma_{X_o^1,Y_o^1}}}^{-1}(\mathbf{x-\mu})^T\right)}}
\label{eq:landmark_sl_un}
\end{equation}
where $\mathbf{x}$ is the viewer's location as a two dimensional vector $\mathbf{x}=(x_o\; y_o)$,
$\mu$ is their corresponding point on the straight line $\mu=[X_o^1\; Y_o^1]$, $\Sigma_{X_o^1,Y_o^1}$ is the covariance matrix of the uncertainty of $\mu$, and $|\Sigma_{X_o^1,Y_o^1}|$ is the determinant of the matrix. Note that when $\Sigma_{X_o^1,Y_o^1}$ is indefinite, Equation~\ref{eq:landmark_sl_un} can be simplified as by neglecting the interaction terms in the covariance matrix. 

For the above example in Figure~\ref{fig:landmark_line1}, the estimated probability of the two viewers observing an inconsistent ordering relation are, respectively, $0.793$ and $0.235$, which suggests that it is more likely for \emph{viewer 1} to observe an inconsistency than \emph{viewer 2}.
 
\begin{figure}[htp] 
    \centering  
      \includegraphics[width = \linewidth]{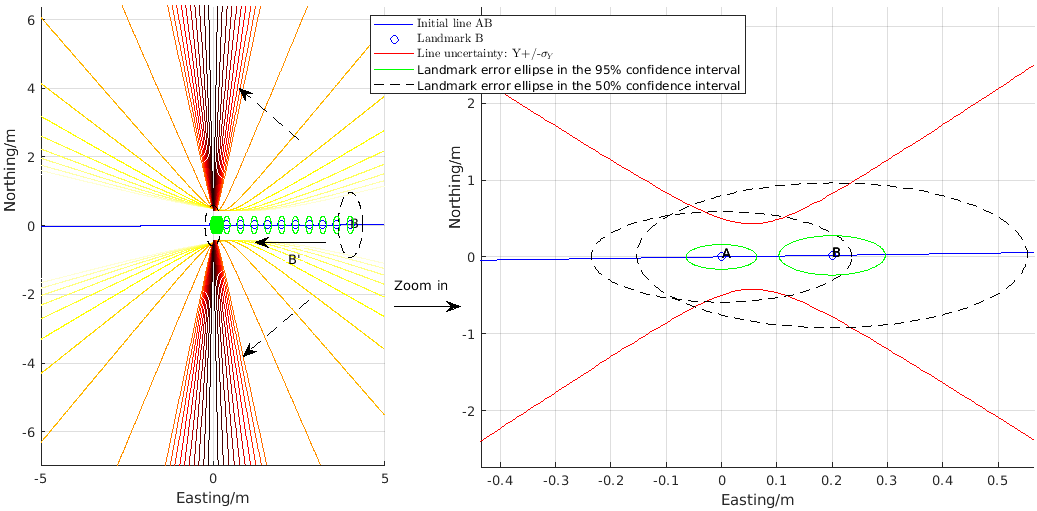} 
    \caption{(Left) When landmark \emph{B} is getting closer to landmark \emph{A}, the uncertainty of the straight line $AB$ will increase sharply. The upper and lower bound of the $Y$ values of those lines are shown as coloured curves. (Right) A zoomed view of the error ellipses and error bounds when the distance between \emph{A} and \emph{B} equals $0.18m$.}
    \label{fig:landmark_line2}
 \end{figure}

\noindent \textbf{Impact of the distance between two landmarks}.
For the same example discussed above, if the two landmarks are closer to each other (e.g. $d=0.2m$) while retaining the same level of uncertainty, the uncertainty of the line will increase rapidly. For example, as shown in Figure~\ref{fig:landmark_line2}, if \emph{A} keeps still and \emph{B} moves eastwards from $x_2=4m$ to $x_2=0.02m$, the slope of line $SL^{AB}$ will increase gradually; the upper and lower error bound will become much sharper, suggesting that from most locations in the space (between the two error bound curves), the observed ordering of the two landmarks will be uncertain. 

\subsubsection{Propagating the uncertainty of two co-visible landmarks to perpendicular lines}
\label{sec:pl_un}

Similarly, the equations of the two \textbf{\emph{Perpendicular Lines ($PL^{AB}_a, PL^{AB}_b$)}} of line $SL^{AB}$ passing through point $A$ and $B$ can be written as
\begin{align*}
Line~1:\mathbf{Y_1=a_1X+b_1}\; (or \; \mathbf{X_1=a'_1Y+b'_1});\\
Line~2:\mathbf{Y_2=a_2X+b_2}\; (or \;
\mathbf{X_2=a'_2Y+b'_2})\; 
\end{align*}
where \begin{align*}
\mathbf{a_1} = \mathbf{a_2}=\frac{x_1-x_2}{y_2-y_1};\; \mathbf{b_1}=y_1-\frac{x_1-x_2}{y_2-y_1}x_1;\;
\mathbf{b_2}=y_2-\frac{x_1-x_2}{y_2-y_1}x_2\\
\mathbf{a'_1} = \mathbf{a'_2}=\frac{y_1-y_2}{x_2-x_1};\; \mathbf{b'_1}=x_1-\frac{y_1-y_2}{x_2-x_1}y_1;\;
\mathbf{b'_2}=x_2-\frac{y_1-y_2}{x_2-x_1}y_2
\end{align*}
Note that $\mathbf{a_1}=\mathbf{a_2}$ and $\mathbf{a'_1}=\mathbf{a'_2}$ as the two perpendicular lines are parallel. Then, we can propagate the landmark uncertainty $\mathbf{\Sigma^{A,B}}$ (in Equation~\ref{eq:landmark_un}) to the parameters of the two perpendicular lines $(a_1, b_1, b_2)$ and $(a'_1, b'_1, b'_2)$, as $
        \mathbf{\Sigma^{a_1,b_1,b_2;a'_1,b'_1,b'_2}}
        =
 \mathbf{J_1\Sigma^{A,B}J_1^T}
    $, 
    where $\mathbf{J_1}$ is the Jacobian matrix calculated as below:
    \begin{equation} 
    \mathbf{J_1}=
\begin{bmatrix}
\begin{array}{c}
  \frac{\partial a_1}{\partial (x_1, y_1,x_2, y_2)}\\[1ex] 
  \frac{\partial b_1}{\partial x_1, y_1,x_2, y_2)}\\[1ex] 
  \frac{\partial b_2}{\partial x_1, y_1,x_2, y_2)}\\[1ex]\hline
  \frac{\partial a'_1}{\partial (x_1, y_1,x_2, y_2)}\\[1ex] 
  \frac{\partial b'_1}{\partial x_1, y_1,x_2, y_2)}\\[1ex] 
  \frac{\partial b'_2}{\partial x_1, y_1,x_2, y_2)}
  \end{array}
\end{bmatrix}
=
\begin{bmatrix}
\begin{array}{cccc}
  \frac{1}{y_2-y_1} & \frac{x_1-x_2}{(y_2-y_1)^2} & \frac{-1}{y_2-y_1}& \frac{x_2-x_1}{(y_2-y_1)^2} \\[1ex]
  \frac{-2x_1+x_2}{y_2-y_1} & [1-\frac{x_1(x_1-x_2)}{(y_2-y_1)^2}] & \frac{x_1}{y_2-y_1}& \frac{x_1(x_1-x_2)}{(y_2-y_1)^2} \\[1ex]
  \frac{-x_2}{y_2-y_1} & \frac{-x_2(x_1-x_2)}{(y_2-y_1)^2} & \frac{2x_2-x_1}{y_2-y_1}& [1+\frac{x_2(x_1-x_2)}{(y_2-y_1)^2]} \\[1ex]\hline
  
  \frac{y_1-y_2}{(x_2-x_1)^2}&\frac{1}{x_2-x_1} & \frac{y_2-y_1}{(x_2-x_1)^2}&\frac{-1}{x_2-x_1} \\[1ex]
  1-\frac{y_1(y_1-y_2)}{(x_2-x_1)^2} &\frac{-2y_1+y_2}{x_2-x_1}&
  \frac{y_1(y_1-y_2)}{(x_2-x_1)^2} &
  \frac{y_1}{x_2-x_1}\\[1ex]
  \frac{-y_2(y_1-y_2)}{(x_2-x_1)^2}&
  \frac{-y_2}{x_2-x_1}&
  1+\frac{y_2(y_1-y_2)}{(x_2-x_1)^2}&
  \frac{2y_2-y_1}{x_2-x_1} 
\end{array}
\end{bmatrix}
\label{eq:Jaco2_2}
\end{equation}

\begin{figure}[htb] 
    \centering  
    \subfigure[Landmark \emph{A} and \emph{B} are on a horizontal line.]{\label{fig:landmark_line3_1}
      \includegraphics[width = 0.3\linewidth]{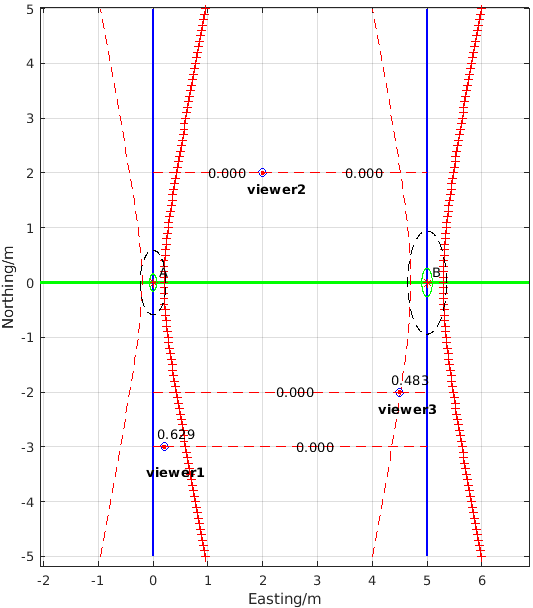} }
      \subfigure[Landmark \emph{A} and \emph{B} are on a vertical line.]{\label{fig:landmark_line3_3}
      \includegraphics[width = 0.3\linewidth]{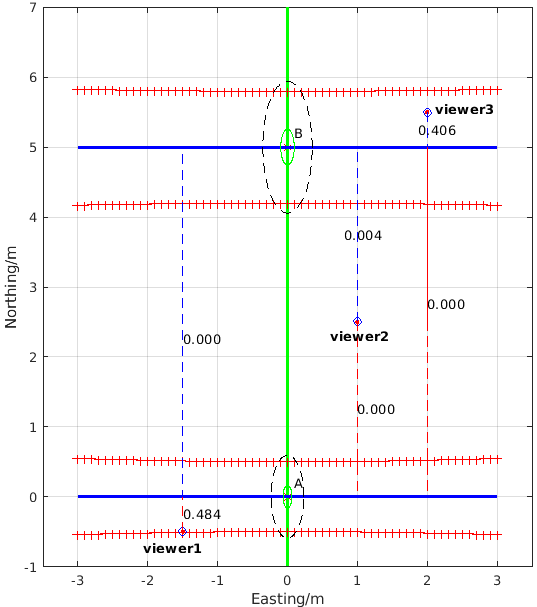}}      
    \subfigure[Landmark \emph{A} and \emph{B} are neither horizontal nor vertical.]{\label{fig:landmark_line3_2}
      \includegraphics[width = 0.31\linewidth]{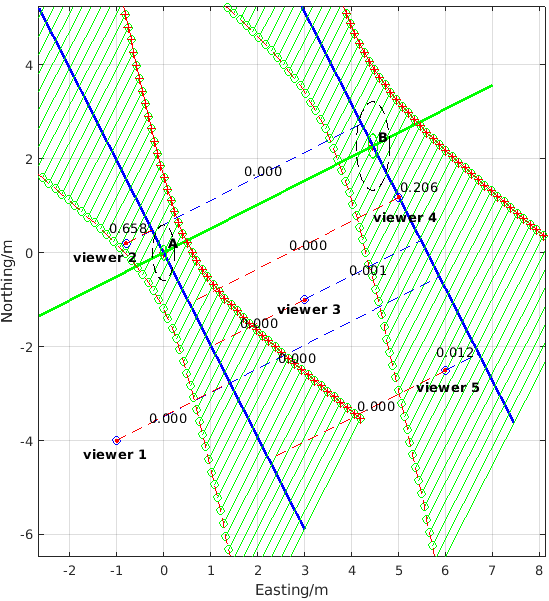} }      
    \caption{The uncertainty of the perpendicular lines of line \emph{AB} as well as the probabilities of viewers to be on such lines, which are considered as their probability to observe an inconsistent relative orientation relation from those locations.}
    \label{fig:landmark_line3}
 \end{figure}
Then, given a list of points on the line as $(X, Y)$, the errors of $Y_1$ and $Y_2$ are:
\begin{enumerate}
    \item When $a_1\neq0$ and $a'_1\neq0$, the two perpendicular lines are neither vertical nor horizontal, the uncertainty of a point on these lines can be estimated by:
        \begin{equation}
        \mathbf{\Sigma^{X_1,Y_1}}
        =\begin{bmatrix}        \mathbf{\sigma^2_{X_1}} & \mathbf{\sigma_{X_1,Y_1}}\\
        \mathbf{\sigma_{X_1,Y_1}}&
        \mathbf{\sigma^2_{Y_1}}
        \end{bmatrix}=
 \mathbf{J_2\Sigma^{a_1,b_1,b_2;a'_1,b'_1,b'_2}J_2^T}
    \label{eq:pl_un3}
    \end{equation}
    where $
    \mathbf{J_2}=
\begin{bmatrix}
 0 & 0 & 0 & Y_1 & 1&0\\[1ex] 
 X_1& 1& 0& 0& 0& 0
\end{bmatrix}$. If we expand the equation,
the error on $X$ and $Y$ directions are equivalent to:
     \begin{align*}
     \begin{array}{c|c}
     \sigma^2_{X_1}=\sigma^2_{a'_1}Y^2+\sigma^2_{b'_1} + 2\sigma_{a'_1,b'_1}X;&\sigma^2_{X_2}=\sigma^2_{a'_1}Y^2+\sigma^2_{b'_2} + 2\sigma_{a'_1,b'_2}X\\
    \sigma^2_{Y_1}=\sigma^2_{a_1}X^2+\sigma^2_{b_1} + 2\sigma_{a_1,b_1}X;&\sigma^2_{Y_2}=\sigma^2_{a_1}X^2+\sigma^2_{b_2} + 2\sigma_{a_1,b_2}X\\
    \end{array}
    \label{eq:pl_12}
    \end{align*}
    
    \item When $a_1=0$, $a'_1=Inf$, the two perpendicular lines are horizontal. Therefore, $\sigma^2_{X_1} = \sigma^2_{X_2} = 0$ and the upper-left part of $\mathbf{\Sigma^{a_1,b_1,b_2;a'_1,b'_1,b'_2}}$ is used to estimate the uncertainty of points on the lines on the $Y$ direction as:
    $\sigma^2_{Y_1}$ and $\sigma^2_{Y_2}$.
    \item When $a'_1=0$, $a_1=Inf$, the two perpendicular lines are vertical. Therefore, $\sigma^2_{Y_1} = \sigma^2_{Y_2} = 0$ and the bottom-right part of $\mathbf{\Sigma^{a_1,b_1,b_2;a'_1,b'_1,b'_2}}$ is used to estimate the uncertainty of points on the lines on the $X$ direction as $\sigma^2_{X_1}$ and $\sigma^2_{X_2}$.
\end{enumerate}
 
Three examples are shown in Figure~\ref{fig:landmark_line3} to illustrate the above three scenarios. 
Given the location of a viewer $o(x_o, y_o)$, we first find their nearest points on the two perpendicular lines $PL^{AB}_a$ and $PL^{AB}_b$ as 
\begin{align*}
    X_o^{2,i} = \frac{a_1y_o+x_o-a_1b_i}{a_1a_1+1},\; Y_o^{2,i}=a_1X_o^{2,i}+b_i
\end{align*}
where $i=1,2$
Then, the corresponding uncertainty of the two points $\mathbf{\Sigma^{X_o^{2,i},Y_o^{2,i}}}$ and the probability $fl(o)_2^i$ of a viewer being on each perpendicular line can be calculated using Equation~\ref{eq:pl_un3} and Equation~\ref{eq:landmark_sl_un}. 
For example, in the scenario shown in Figure~\ref{fig:landmark_line3} (a), the estimated probability for the three viewers being potentially on the two perpendicular lines are respectively $fl(o_1)_2= (
0.629,\;0.000)$, $fl(o_2)_2=(0.000,\;0.000)$ and $fl(o_3)_2= 0.000,\;0.483$. 
{The higher of the two probabilities is considered as the viewer's probability to observe a different relative orientation relation compared to those pre-calculated ones in the reference database.}
Therefore, it is likely for \emph{viewer 1} to observe an index $1$ rather than $3$, but it is very unlikely for \emph{viewer 2} to misjudge the relative orientation index other than $3$. 
 
\subsubsection{Propagating the uncertainty of two co-visible landmarks to their circular line}
\label{sec:cir_un}

\begin{figure}[htb!] 
    \centering  
    \subfigure[Definition of $\theta$ and $\phi$.]{\label{fig:cl_1}
      \includegraphics[width = 0.48\linewidth]{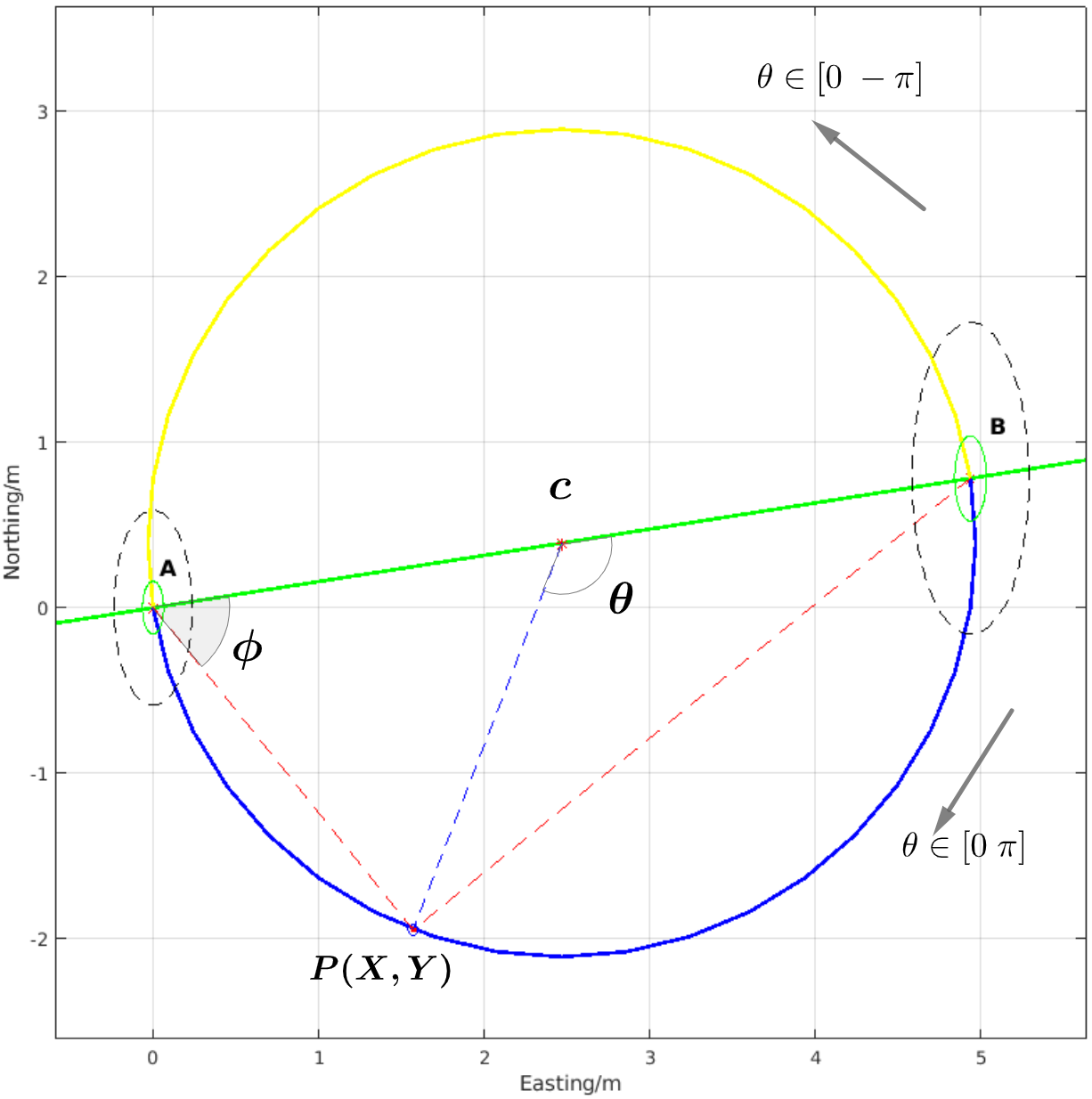}}
    \subfigure[Uncertainty of points on the circle.]{\label{fig:cl_2}
    \includegraphics[width = 0.46\linewidth]{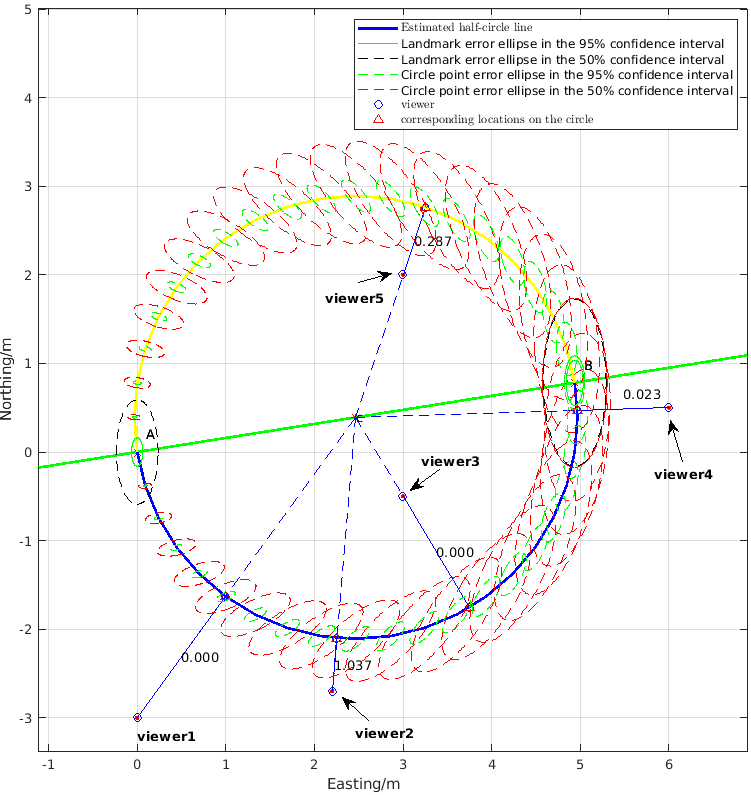} }  
    \caption{a) Definition of a circle \textcolor{black}{with a diameter \emph{AB}}; b) Uncertainty of points on the circle are defined by \textcolor{black}{the uncertainty of landmarks} \emph{A} and \emph{B} and the probabilities of viewers being on the circle.}
    \label{fig:cl}
 \end{figure}

In order to propagate the location uncertainty of landmarks \emph{A} and \emph{B} onto the corresponding
\textbf{\emph{Circular Line ($CL^AB$)}}, the equation of points \emph{P} on the half-circle passing through point $A$ and $B$ are written as:
\begin{equation}
\begin{array}{l}
\mathbf{X=(x_2-x_1)cos(\phi)^2+(y_2-y_1)sin(\phi)cos(\phi)}\\
\mathbf{Y=(y_2-y_1)cos(\phi)^2-(x_2-x_1)sin(\phi)cos(\phi)}
\end{array}
\label{eq:circle}
\end{equation}
where $\phi$ is the clockwise angle of vector $\overrightarrow{AP}$  with respect to $\overrightarrow{AB}$, as shown in Figure~\ref{fig:cl} (a). 
When $\phi\in[0,\;\frac{\pi}{2}]$, the points are on the bottom half of the circle from where \emph{A} would be observed before \emph{B};
when $\phi\in(\frac{\pi}{2},\; \pi]$, the points are on the top part of the circle from where \emph{B} would be observed before \emph{A}.

Then, given a $\phi$, the location uncertainty of a point on the circle can be estimated by propagating the uncertainty of the landmark pair $\mathbf{\Sigma^{A,B}}$ as in previous sections, written as:$
        \mathbf{\Sigma^{X,Y}}
        =\begin{bmatrix}        \mathbf{\sigma^2_{X}} & \mathbf{\sigma_{X,Y}}\\
        \mathbf{\sigma_{X,Y}}&
        \mathbf{\sigma^2_{Y}}
        \end{bmatrix}=
        \mathbf{J_3\Sigma^{A,B}J_3^T}
    \label{eq:cl_un1}
$, where $\mathbf{J_3}$ is the Jacobian matrix calculated as below:
    \begin{equation} 
    \mathbf{J_3}=
\begin{bmatrix}
\begin{array}{c}
  \frac{\partial X}{\partial (x_1, y_1,x_2, y_2)}\\[1ex] 
  \frac{\partial Y}{\partial x_1, y_1,x_2, y_2)}
  \end{array}
\end{bmatrix}
=    
\begin{bmatrix}
  -cos(\phi)^2& -sin(\phi)cos(\phi)& cos(\phi)^2& sin(\phi)cos(\phi) \\[1ex]
 sin(\phi)cos(\phi)&-cos(\phi)^2&-sin(\phi)cos(\phi)&cos(\phi)^2
\end{bmatrix}
\label{eq:Jaco3_1}
\end{equation} 

Since the location of the circle is uncertain, the actual angle observed between A and B could change between \emph{obtuse} and \emph{acute}. To estimate the probability of a viewer $\mathbf{o(x_o, y_o)}$ to observe a different qualitative angle to the stored one, we first need to find its corresponding point \emph{P} on the circle, which is the intersection of the circle with the line connecting the circle centre $c (\frac{x_1+x_2}{2}\; \frac{y_1+y_2}{2})$) and the viewer $o$, as shown in Figure~\ref{fig:cl} (b). To do this, the angle $\phi_o^3$ from the vector $\overrightarrow{AB}$ to $\overrightarrow{AP}$ is calculated as:
\begin{equation}
\begin{array}{ll} 
&v_1=\overrightarrow{cB} = [x_2\;y_2]-c;\;
v_2=\overrightarrow{co} = [x_o\;y_o]-c;\\
\Rightarrow &\theta_o^3 = -atan2(v_1^1v_2^2-v_1^2v_2^1, v_1^1v_2^1+v_1^2v_2^2)\\
\Rightarrow &\phi_o^3 = \theta_o^3/2
\label{eq:nearest_circle_point}
\end{array}
\end{equation}
Then, the coordinates of the intersection point ($\mathbf{X_o^3,Y_o^3}$), its associated uncertainty $\mathbf{\Sigma^{X_o^3,Y_o^3}}$, and the probability $fl(o)_3$ of a viewer being on the circle can be estimated by propagating the uncertainty of landmarks as in Equation~\ref{eq:landmark_sl_un}.
$fl(o)_3$ is also considered as the viewer's probability to encounter an inconsistent qualitative angle between \emph{A} and \emph{B}. 
For example, with the two landmarks $A(0,0)$ and $B(4.94,0.78)$ shown in Figure~\ref{fig:cl} (b) with uncertainty $(\sigma_{x_1},\sigma_{y_1},\sigma_{x_2},\sigma_{y_2})$= (0.2, 0.5 0.3 0.8), the estimated probability for the five viewers to potentially be on the circular lines are respectively $fl(o)_3= (0.000, 1.037, 0.000,0.023, 0.287$), suggesting that Viewer 5 is most likely to observe an obtuse angle other than the expected acute one.

\subsubsection{Propagating the uncertainty of a landmark to its Boundary Line}
\label{sec:bl_un}

As mentioned earlier in Section~\ref{sec:landmarks}, the visibility range of a landmark is first defined as a simplified circular (or fan-shaped) buffer zone with a certain radius, then clipped by the outlines of buildings using a 2D viewshed algorithm (see Appendix~\ref{sec:viewshed}). Therefore, the \textbf{\emph{Boundary Line (BL)}} of a landmark \emph{A} can either be on a circle or on building facades. An example is given below in Figure~\ref{fig:bl_1}. 
\begin{figure}[h!] 
    \centering  
      \includegraphics[width = 0.245\linewidth]{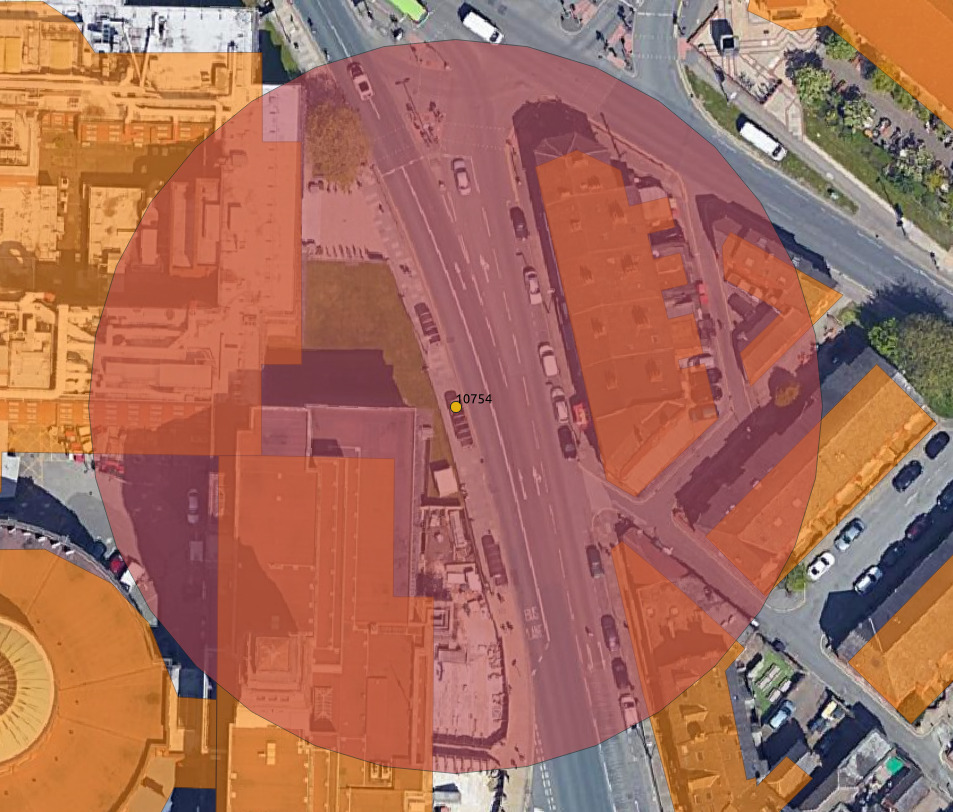} 
      \includegraphics[width = 0.245\linewidth]{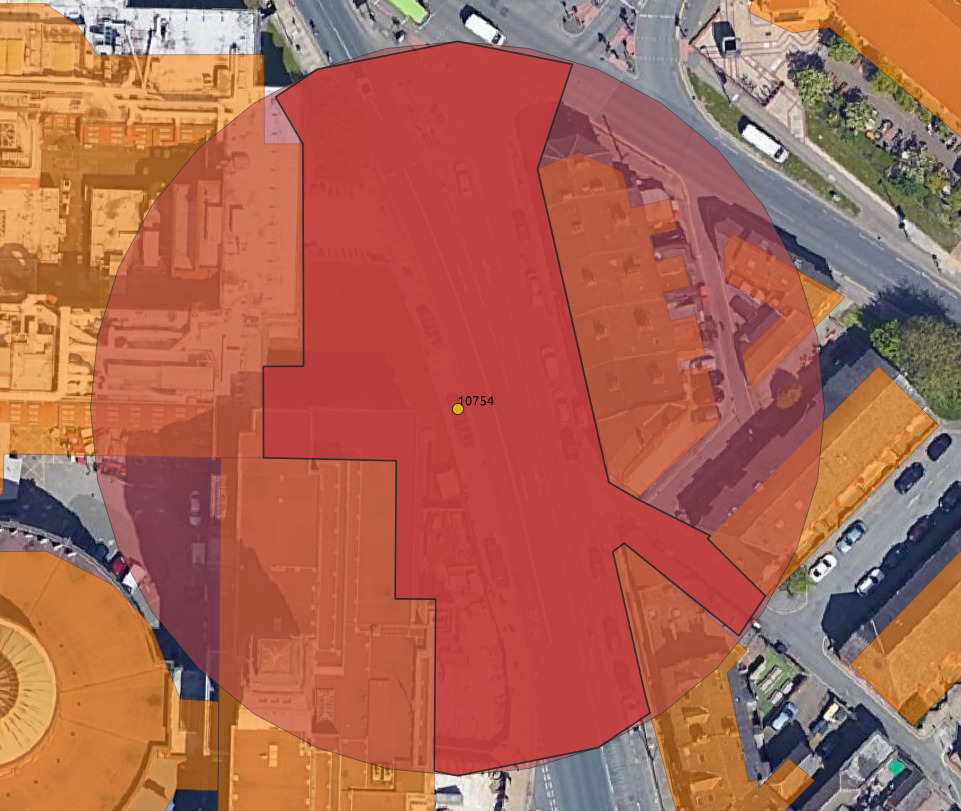}
    \includegraphics[width = 0.23\linewidth]{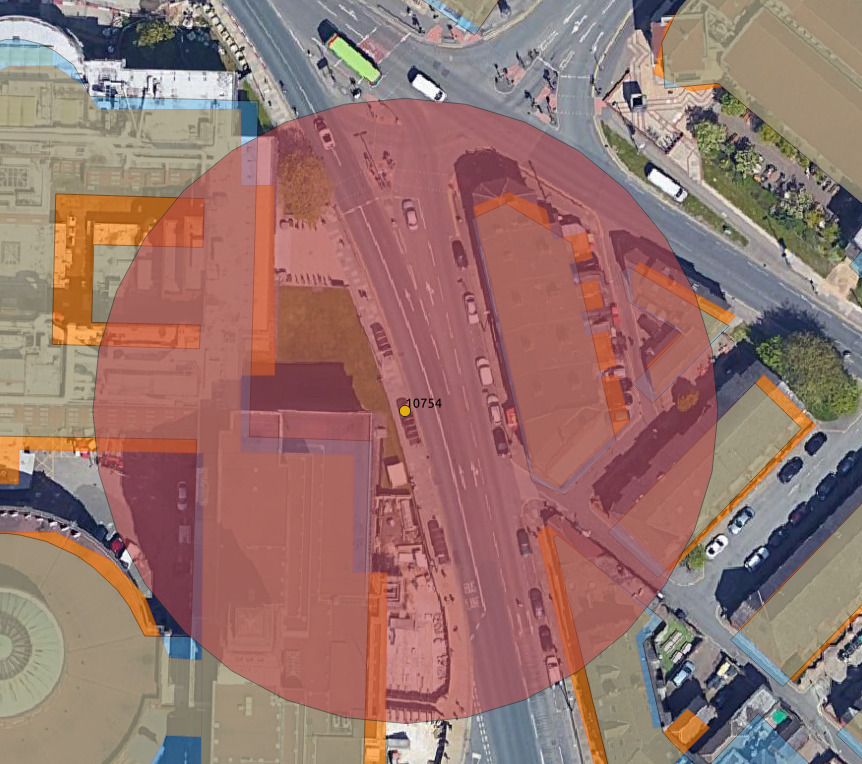}  
    \includegraphics[width = 0.23\linewidth]{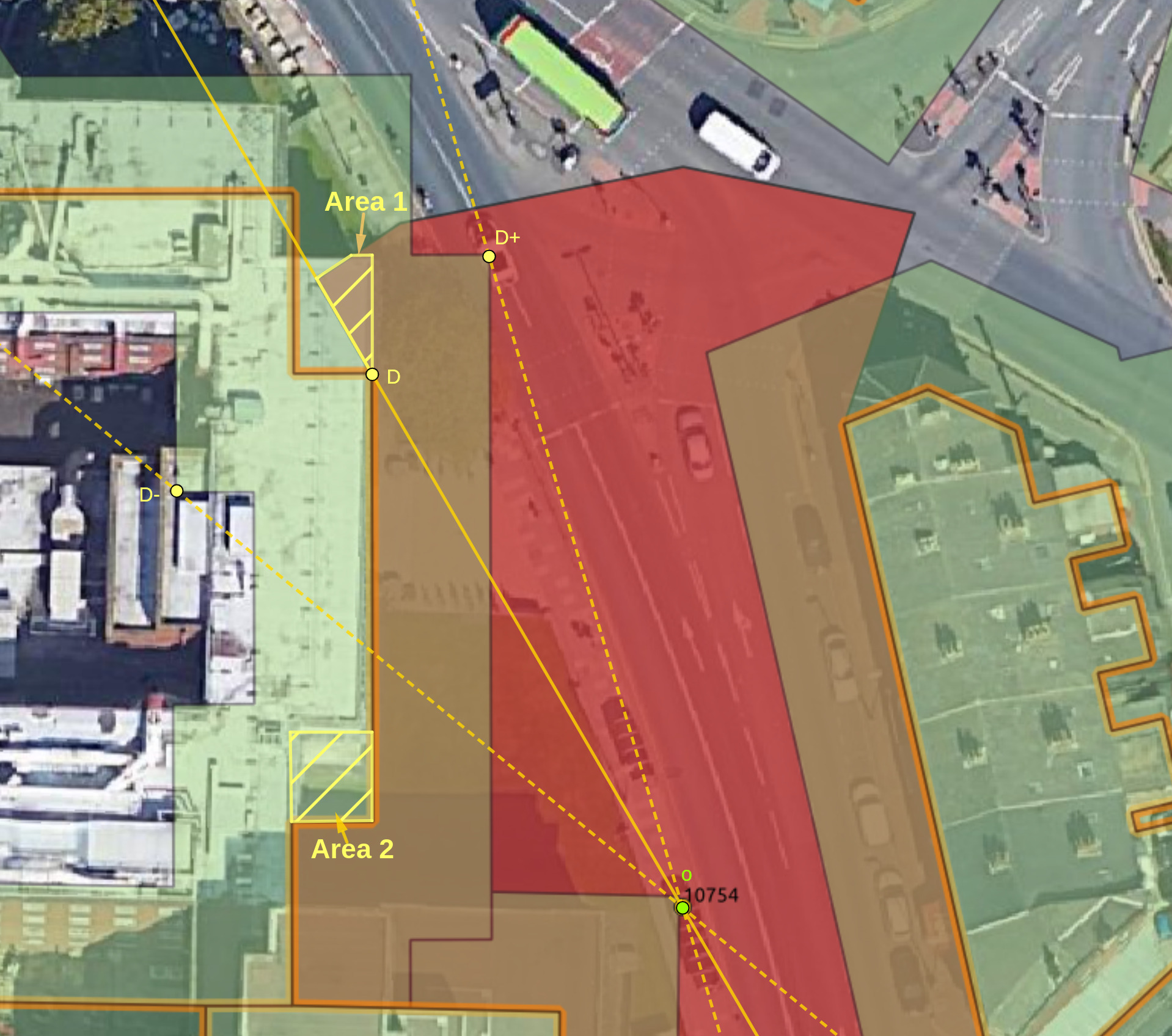}  
    \caption{In these figures, (a-b) orange polygons represent the buildings extracted from the OpenStreetMap, light red area demonstrates the visibility range of a landmark, and dark red area in the right image shows the visibility range after considering the occlusion by buildings. (c) Incoherent building outlines are observed in OpenStreetMap (orange), Ordnance Survey (blue) and Google Satellite Image. (d) The uncertainty of building outlines can be shown as a buffer zone of the building outlines. The orange lines are the outlines of buildings (OSM) and the light green zone are the 8-metre buffer zone of the building outlines.} 
    \label{fig:bl_1}
 \end{figure}
\begin{itemize}
    \item When a viewer is very close to the circular boundary (i.e. away from the landmark centre), they may not be able to observe the landmark as expected due to the location uncertainty of the landmark. Location of the points on the circular range boundary of the landmark ($x_1,y_1$) written as: $(X = x_1+Rcos(\theta);\;
        Y = y_1+Rsin(\theta))$, 
where $\theta \in [0\;2\pi]$ (when there is no occlusion). It can be seen from above equations that the uncertainty of these points are simply the same as the centre landmarks: $
        \mathbf{\Sigma^{X_4, Y_4}}=\mathbf{\Sigma^{A,B}}$.
Given the location of a viewer $o(x_o,y_o)$, if their situated place cell is on the circular boundary, the nearest point of the viewer on the boundary can be identified by:
    \begin{equation}
        \begin{array}{c}             X_o^4=\frac{R(x_o-x_1)}{\sqrt{(x_o-x_1)^2+(y_o-y_1)^2}}+x_1;\; Y_o^4=\sqrt{R^2-(X_o^4-x_1)^2}+y_1
        \end{array}
    \end{equation}
    
Then, the probability $fl(o)_4$ of a viewer being on the circle, i.e. the probability of a viewer to miss the landmark due to the landmark location uncertainty, can be approximated as in Equation~\ref{eq:landmark_sl_un}.
    
\item When the visibility boundary is on building facades, the situation becomes a bit more complex. For example, offsets were observed in the building maps from OpenStreetMap (orange areas), Ordnance survey\footnote{OS OpenMap - Local: \url{https://osdatahub.os.uk/downloads/open/OpenMapLocal}, last accessed: 2022-07-21.} (blue areas) and Google Satellite Image, as shown in Figure~\ref{fig:bl_1} (c). 
\end{itemize}
 
These offsets bring in two types of uncertainty in generating the place cells and place signatures. Specifically,
\begin{enumerate}
    \item when the outline of a building is actually further away from a landmark than expected (based on an existing map), some free space with a clear view to the landmark could be missing from the pre-defined place cells, such as \textbf{Area 2} in the Figure~\ref{fig:bl_1} (d). Therefore, if a viewer is in one of these areas, no exactly matched place signature would be found in the reference database;
    \item when the actual outline of a building is closer to the landmark than expected, this landmark is in fact invisible in some areas in the pre-generated place cells, such as \textbf{Area 1} in the Figure~\ref{fig:bl_1} (d).
\end{enumerate}
For the second type of uncertainty, we can propagate the uncertainty from building outlines to place cells based on their distances to buildings, using the similar procedure as in Section~\ref{sec:sl_un} and Section~\ref{sec:pl_un}. However, as seen in Figure~\ref{fig:bl_1} (d), offsets in different scales and directions could be observed for different buildings, making it difficult to assign a general error level. Ideally, a local alignment would be ideal to remove these offsets. To not increase the length of this article further, this part of the work will be examined in a subsequent work.

\end{document}